# An Ubuntu-Guided Large Language Model Framework for Cognitive Behavioral Mental Health Dialogue

Sontaga G. Forane[1], Absalom E. Ezugwu[1,*], Kevin Igwe[2], Karen van den Berg[1]

[1] Unit for Data Science and Computing, North-West University, 11 Hoffman Street, Potchefstroom, 2520, North-West, South Africa.
[2] Department of Psychology, Faculty of Humanities, University of Johannesburg, Bunting Road, Auckland Park, Johannesburg 2092, South Africa.

Authors emails: Sontaga G. Forane (34228128@mynwu.ac.za), Absalom E. Ezugwu (Absalom.Ezugwu@nwu.ac.za), Kevin Igwe (kigwe@uj.ac.za), Karen van den Berg (Karen.Pretorius@nwu.ac.za)

**Abstract**. South Africa's escalating mental health crisis, compounded by limited access to culturally responsive care, calls for innovative and contextually grounded interventions. While Large Language Models (LLMs) show considerable promise for mental health support, their predominantly Western-centric training data limit cultural and linguistic applicability in African contexts. This study introduces a proof-of-concept framework that integrates Cognitive Behavioural Therapy (CBT) with the African philosophy of Ubuntu to create a culturally sensitive, emotionally intelligent, AI-driven mental health dialogue system. Guided by a Design Science Research (DSR) methodology, the framework applies both deep (theoretical and therapeutic) and surface-level (linguistic and communicative) cultural adaptations. Key CBT techniques, Behavioural Activation and Cognitive Restructuring, were reinterpreted through Ubuntu principles emphasizing communal well-being, spiritual grounding, and interconnectedness. A culturally adapted dataset was developed through iterative processes of language simplification, spiritual contextualization, and Ubuntu-based reframing. The fine-tuned model was evaluated through expert-informed case studies, employing UniEval for conversational quality assessment and additional measures of CBT reliability and cultural-linguistic alignment. Results demonstrate that the model effectively engages in empathetic, context-aware dialogue aligned with both therapeutic and cultural objectives. Although real-time end-user testing has not yet been conducted, the model underwent rigorous review and supervision by domain-specialist clinical psychologists. The findings highlight the potential of culturally embedded emotional intelligence in enhancing the contextual relevance, inclusivity, and effectiveness of AI-driven mental health interventions across African settings.

## 1. Introduction

Artificial intelligence (AI) has undergone rapid advancements in recent years, particularly in the development of Large Language Models (LLMs) such as Gemini and GPT-4. These models have demonstrated remarkable capabilities in generating human-like text, performing complex analysis, and interpreting data at scale (Hadi *et al.*, 2023). Their applications span multiple domains, including medicine, education, finance, and mental health research (Chen *et al.*, 2023; Hadi *et al.*, 2023; Li *et al.*, 2023; Zhao *et al.*, 2024; Chow and Li, 2025). While LLMs have been successfully integrated into various fields, their adaptation to culturally diverse contexts, such as Africa, remains a significant challenge. Despite their growing utility, most LLMs are trained on datasets that reflect predominantly Western cultural values (Bender *et al.*, 2021). This training bias raises concerns about their applicability in non-Western contexts, particularly in African societies where communal values and cultural philosophies such as Ubuntu play a critical role in shaping societal interactions (Nwoye, 2015). This study seeks to explore how LLMs can be adapted to align with African cultural perspectives, particularly in the context of mental health interventions, by integrating the principles of Emotional Intelligence (EI) and the African philosophy of Ubuntu.

---

* Corresponding author. Absalom El-Shamir.
  *E-mail address:* Absalom.ezugwu@nwu.ac.za

Mental health disorders, particularly anxiety and depression, are growing global concerns, significantly impacting individuals and societies worldwide (Kola *et al.*, 2021; Organization, 2017). In Africa, the situation is further complicated by limited access to mental health care, social stigma, and a shortage of culturally sensitive interventions (Anakwenze, 2022). In South Africa, where mental health challenges are increasingly prevalent, existing psychological support systems often fail to resonate with African cultural values that emphasize interconnectedness and community well-being (Nwoye, 2015; Suliman *et al.*, 2024).

Cognitive behavioral therapy (CBT) is widely recognized as a gold-standard treatment for anxiety and depression (Beck, 2011; Jiang et al., 2024). While digital interventions such as Computer-Based Cognitive Behavioural Therapy (CCBT) and Internet-Based Cognitive Behavioural Therapy (ICBT) have demonstrated success (Izumi et al., 2024; Jiang et al., 2024), it is important to note that systematic reviews and meta-analyses consistently show that conventional, face-to-face psychotherapy achieves larger effect sizes in treating conditions like depression, particularly in head-to-head comparisons (Cuijpers et al., 2019; Carlbring et al., 2018). The primary challenges for digital modalities include high dropout rates and a lack of cultural adaptation, which have significantly limited their effectiveness and reach in non-Western societies (Anakwenze, 2022; Naeem et al., 2019; Pakomera et al., 2017).

Several studies have highlighted the importance of incorporating cultural contexts into digital mental health interventions, with successful implementations observed in China, Colombia, Indonesia, and Latino communities (Arjadi *et al.*, 2018; Davidson *et al.*, 2015; Hamka *et al.*, 2022; Salamanca-Sanabria *et al.*, 2019). However, similar culturally adapted AI-driven mental health solutions remain underexplored in South Africa. Western psychological frameworks, such as traditional CBT, emphasize individualism and personal responsibility, contrasting with African communal values centered on collective well-being (Wilson & Williams, 2013). Moreover, most LLMS are trained on Western-centric data, which often lacks representation of African worldviews and emotional expressions. This limits their effectiveness in providing culturally relevant psychological support within African contexts.

To address these cultural limitations in digital mental health interventions, it is not only important to adapt therapeutic content but also to enhance the sensitivity of AI tools to emotional and contextual cues. One promising approach is the integration of Emotional intelligence (EI) into LLMs. EI refers to the ability to perceive, understand, and manage emotions (Salovey *et al.*, 2009). In this study, EI is defined as the capacity of an LLM to interpret and respond appropriately to emotions expressed in text while considering the user's cultural context. Understanding emotions is crucial in mental health applications, where emotional nuances significantly influence therapeutic outcomes.

This study aims to bridge this gap by developing a framework for emotionally intelligent LLMs that incorporates the African cultural context. By doing so, it seeks to improve AI-driven mental health interventions, ensuring their relevance and applicability in South African settings. This research addresses two key questions: (1) Can LLMs be trained to be contextually aware of African emotional intelligence? (2) Could an LLM serve as a tool for professional psychologists to detect and support diagnoses of depression and anxiety in the African context?

While LLMs have demonstrated effectiveness in various applications, their reliance on Western-centric training data limits their cultural relevance in African settings (Chow and Li, 2024). Moreover, Chow and Li (2024) highlight that datasets primarily derived from Western medical literature and patient interactions may introduce significant bias by overrepresenting certain demographic groups. This underscores the fact that many existing digital mental health solutions often fail to incorporate African cultural values or contextual factors specific to Africa-centric settings, where community and interconnectedness are deeply emphasized (Anakwenze, 2022; Ramos *et al.*, 2021). The lack of culturally appropriate AI-driven mental health support systems exacerbates the mental health crisis in South Africa, where stigma and limited access to care are significant barriers. Furthermore, conventional CBT interventions do not fully align with African communal values, potentially diminishing their effectiveness. This study seeks to develop a culturally adapted LLM-based framework that integrates Ubuntu and Emotional Intelligence principles to improve AI-driven mental health interventions in South Africa. The following research questions form the basis for the current research.

i. How can emotional intelligence principles be integrated with Africa-centric cultural contexts to enhance the effectiveness of AI-driven mental health support systems?

ii. What are the potential benefits and challenges of implementing AI-driven mental health support frameworks that incorporate emotional intelligence within diverse African communities?

iii. How can the challenges of trust within a cultural context be addressed when implementing emotionally intelligent AI-driven mental health support systems in Africa?

iv. Are large language models (such as ChatGPT) ready to be assistive tools for detecting depression and anxiety?

Furthermore, this research aims to establish a proof-of-concept by developing (1) a framework for emotionally intelligent large language models that incorporate and (2) evaluate the capability of these LLMs to detect depression and anxiety from texts that reflect African cultural nuances. To achieve these dual aims, the research will focus on the following specific objective:

i. Conduct an in-depth literature review of LLM applications in eHealth, examining their current use, the techniques used to adapt them, and identifying gaps in cultural relevance and emotional intelligence integration.

ii. Investigate how existing mental health applications integrate cultural context and operationalize culturally relevant content like themes, metaphors, and perspectives of mental health within different contexts.

iii. Develop a proof-of-concept framework for integrating emotional intelligence into LLMs, informed by Ubuntu and designed to reflect the cultural norms and emotional expressions common in South African communities.

iv. Implement the proposed framework within an AI-powered mental health support application, leveraging a fine-tuned LLM to operationalize the framework.

v. Evaluate the model's alignment with the proposed culturally adapted framework, focusing on its application of Ubuntu-aligned CBT techniques and culturally appropriate communication strategies to assess its overall cultural sensitivity.

This research contributes to the growing body of work on AI-driven mental health interventions by introducing a culturally adapted framework for LLMs in African contexts. By integrating Ubuntu and Emotional Intelligence principles, this study aims to enhance the relevance and effectiveness of AI in mental health support. The findings could help bridge the gap between technological advancements and cultural sensitivity in AI applications for healthcare. Furthermore, this study aligns with ongoing efforts to enhance Diversity, Equity, and Inclusion (DEI) in AI research (Ramos *et al.*, 2021). It provides insights into how AI can be adapted to serve underrepresented communities, thereby promoting more inclusive mental health solutions. Additionally, this research has practical implications for psychologists, mental health practitioners, and AI developers by providing guidelines for creating culturally aware AI-driven mental health support systems.

The remainder of this paper is structured as follows: Section 2 reviews literature on Diversity, Equity, and Inclusion (DEI), Cognitive Behavioural Therapy (CBT) in AI, and emotional intelligence (EI) frameworks, with a focus on cultural adaptation and methods for integrating EI into existing LLMs. Section 3 outlines the research design, system architecture, and methodology for incorporating DEI and EI into an LLM-based eHealth framework. Section 4 presents the results and discusses the effectiveness of the system in addressing mental health challenges in South Africa, similarly, the section also highlights key findings, limitations, and future research directions for culturally sensitive eHealth interventions. Section 5 concludes the paper.

## 2. Literature Review

This section presents a summary overview of medical chatbots and then followed by a comprehensive literature review across six key domains: (1) Diversity, Equity, and Inclusion (DEI), (2) Cultural adaptation and cultural grounding, (3) Cognitive Behavioral Therapy (CBT) in the treatment of anxiety and depression, (4) Techniques for addressing mental health in eHealth applications, (5) Strategies for incorporating emotional intelligence into LLMs, and (6) Fine-tuning LLMs in the context of mental health. The review begins by examining the role of DEI in mental health applications, followed by an exploration of culturally adapted and culturally grounded interventions. It then discusses current techniques used in eHealth applications for mental health support, along with approaches for integrating emotional intelligence into LLMs. This review establishes the foundation for a multifaceted approach to effective mental health interventions, emphasizing cultural and contextual adaptation, emotional intelligence, and AI-driven support. In this context, mental health alleviation refers to mitigating symptoms of common mental disorders before the need for intervention by a mental health professional.

The LLMs hold significant promise for advancing medical chatbots by enabling natural, context-aware interactions (Chow, Sanders, and Li, 2023; Parsa, and Ebrahimzadeh, 2023; Loh, 2023). Trained on medical literature, clinical guidelines, and patient dialogues, they can provide real-time, evidence-based information on symptoms, treatments, and disease management. These models assist in triaging patient concerns, guiding users to appropriate care, and adapting to linguistic and cultural nuances, thereby enhancing accessibility for diverse populations (Wei et al. 2024). Their continuous learning capability also supports personalized healthcare, early intervention, and remote care, particularly in underserved regions facing shortages of medical professionals and advanced facilities.

However, despite their promise, the application of LLMs in medical chatbots also presents several notable challenges and risks. One major concern is the potential for misinformation or inaccurate responses, especially in high-stakes situations where incorrect guidance could have serious health consequences. LLMs may also generate plausible-sounding but medically invalid advice due to limitations in their training data or a lack of real-time clinical validation. Additionally, these models often inherit biases present in their training datasets, which can lead to unequal treatment recommendations across different demographic groups. Privacy and data security are further concerns, as handling sensitive patient information requires strict compliance with healthcare regulations like HIPAA or GDPR. Finally, over-reliance on chatbots may discourage users from seeking professional medical care, especially in contexts where access to human healthcare providers is already limited. For further insights into the potential benefits and challenges of medical chatbots in real-world applications, see Chow and Li (2024), Chow, Wong, and Li (2024), and Wei et al. (2024). Subsequently, we then narrowed our review to examine current research within the African healthcare landscape.

## 2.1 Overview of Diversity, Equity, and Inclusion

The emerging field of digital mental health presents a unique opportunity to bridge the significant gap in mental health service provision. However, this potential can only be realized when the tools developed in this context of mental health applications are accessible, relevant, and effective for all users, especially considering the fact that marginalized groups are disproportionately affected by mental health disparities (Ramos *et al.*, 2021). These disparities are fuelled by unequal access to mental health providers, logistical barriers, a lack of culturally robust services, and stigma, which are more acute in marginalized populations, including those with low socioeconomic status (SES) and people of color (POC) (Ramos *et al.*, 2021). These groups often rely on smartphones for internet access, suggesting that mental health applications could play a crucial role in mitigating professional shortages and logistical barriers to care.

Multiple studies have highlighted the interest and potential benefits of incorporating diversity, equity, and inclusion (DEI) principles in mental health applications (Denecke *et al.*, 2022; Naslund *et al.*, 2019; Ramos *et al.*, 2021). Although there is no consensus on the definition of DEI (Vernon, 2023), this study defines DEI as follows: Diversity refers to ensuring that mental health applications are designed to cater to the varying needs and backgrounds of all individuals, including those from marginalized communities (Vernon, 2023). This involves considering different cultural, linguistic, and socioeconomic factors in the development of the apps (Ramos *et al.*, 2021). However, having one app that caters to all cultures is highly infeasible. With this in mind, this study focuses on three components—Social Norms and Values (e.g., use of appropriate titles such as 'Mama' or 'Tata'), Language and Communication Styles (e.g., greeting with 'Sawubona'), and Religious and Spiritual Beliefs (e.g., integrating biblical references or ancestral reverence), collectively referred to in this study as the Cultural Context.

Equity focuses on addressing the existing disparities in mental health access and outcomes by ensuring that the apps provide fair and equal opportunities for all users to receive adequate support and care (Vernon, 2023). This may involve tailoring the apps to accommodate different levels of access to resources and services (Ramos *et al.*, 2021).

Inclusion entails creating a welcoming and inclusive environment within the apps that respects and values the diverse perspectives and experiences of all users (Vernon, 2023). This may involve incorporating diverse representation in the app content and considering the different ways in which individuals from marginalized communities access and engage with interventions, ultimately promoting greater inclusivity and effectiveness in mental health disparities (Ramos *et al.*, 2021). Research has shown that individuals from ethnic and racial minorities express interest in accessing mental health services through social media platforms (Anakwenze, 2022). For example, a survey of social media users with mental illness found that the majority were interested in accessing mental health programs on social media, targeting symptom management, health promotion, and support for communicating with healthcare providers and interacting with the health system (Anakwenze, 2022).

Additionally, it is crucial to consider the perspectives and experiences of the target population (Naslund *et al.*, 2020). In so doing, we can develop interventions that are tailored to their specific needs and preferences. Moreover, research has shown that individuals with serious mental illness have stressed the importance of having culturally responsive and inclusive mental health interventions on social media (Spanhel *et al.*, 2021). This includes considering cultural beliefs, determining language preferences, and addressing structural barriers hindering access to mental health resources. Incorporating inclusion principles in mental health applications is essential to address the disparities faced by marginalized populations. By acknowledging and addressing these disparities, mental health applications can help bridge the gap in access to mental health services and provide support to individuals from diverse backgrounds.

Incorporating diversity, equity, and inclusion in mental health applications can help address the specific needs and preferences of individuals from diverse backgrounds and ensure that mental health support is accessible to all, regardless of their socioeconomic status or geographic location (Mapes *et al.*, 2020; Ramos *et al.*, 2021).

Despite the promise of mental health applications, many lack empirical evidence supporting their effectiveness, partly due to insufficient integration of diversity, equity, and inclusion in DEI consideration in their development and evaluation (Ramos *et al.*, 2021). DEI in the digital realm necessitates ensuring all individuals and communities, particularly the most disadvantaged, have the technological capacity needed for full participation in our society. This encompasses making mental health technology adaptable and accessible to meet diverse needs, encompassing various factors such as age, gender, sexuality, race, ethnicity, culture, and socioeconomic status (Ramos *et al.*, 2021). The need for DEI considerations is highlighted by the observation that most mental health applications have been developed and tested primarily within high-income Western populations, without adequately considering the unique experiences of low and middle-income populations, such as those in Africa (Caloudas *et al.*, 2023), and in the context of this study, South Africa.

Unfortunately, the development and testing of most mental health applications have not adequately considered the unique experiences of low-income groups, such as those in African countries, such as South Africa (Caloudas *et al.*, 2023). This oversight potentially limits the effectiveness and acceptability of the app among these populations. Furthermore, the content and representation within apps often lack the inclusion necessary to engage users from varied backgrounds effectively. Many existing evaluation frameworks for application quality do not consistently assess DEI considerations, which are crucial for reducing health disparities and ensuring the broad usability of mental health applications (Vernon, 2023).

Beyond policy and discourse, practical manifestations of diversity, equity, and inclusion are increasingly evident in African artificial intelligence and natural language processing (NLP) initiatives. The Masakhane project (Orife *et al.*, 2020) exemplifies a pan-African, community-driven approach to machine translation and NLP, promoting the development of language technologies for Africans, by Africans. Such initiatives counterbalance the historical dominance of Western-centric datasets and algorithms by prioritizing African linguistic and cultural representation in AI systems. Building upon this foundation, projects such as MasakhaNER (Adelani *et al.*, 2021) and MasakhaNER 2.0 (Adelani *et al.*, 2022) have produced annotated datasets for named entity recognition across multiple African languages, advancing equitable access to NLP resources. More recently, MasakhaNEWS (Adelani *et al.*, 2023) extended this work to text classification tasks, highlighting the potential of multilingual AI for African media and communication contexts. Within South Africa, Gaustad and McKellar (2024) have expanded the availability of morphologically annotated corpora for nine of the country's official languages, providing essential resources for language-sensitive applications. Together, these African NLP initiatives exemplify how DEI principles can be operationalized through inclusive data creation, collaboration, and technological empowerment, laying the groundwork for culturally responsive AI systems in domains such as mental health.

For mental health applications to fulfill their potential as tools for mental health crises in African countries such as South Africa, considerations of diversity, equity, and inclusion must be central to their development. By doing so, these digital interventions can become genuinely inclusive, ensuring that all users can access effective, culturally competent mental health support regardless of their background. Although this study focuses on cultural adaptation as a crucial component of the inclusion aspect of DEI, it is important to note that diversity, equity, and inclusion are not mutually exclusive but overlap with each other.

## 2.2 Cultural adaptation vs. Culturally grounded interventions

Culturally grounded interventions are approaches to therapy or mental health interventions rooted in a deep understanding and acknowledgement of the cultural background and beliefs of the individuals receiving the services (Anakwenze, 2022; Caloudas *et al.*, 2023). These interventions are tailored to the specific cultural, social, and psychological needs of different communities and are designed to be sensitive and responsive to the unique contexts in which they are implemented. Culturally grounded interventions often involve collaboration with community leaders and members to ensure that the services provided are relevant and effective within a specific cultural context (Anakwenze, 2022; Spanhel *et al.*, 2021)

Cultural adaptation involves modifying interventions or treatments to enhance their relevance, acceptability, and effectiveness for individuals from diverse cultural backgrounds (Anakwenze, 2022; Caloudas *et al.*, 2023). This process integrates cultural beliefs, values, practices, and language to ensure accessibility and appropriateness for the target population (Anakwenze, 2022). It encompasses two types of modifications: surface structure changes, which involve basic adaptations to fit interventions within specific cultural contexts, and deep structure changes, which incorporate elements that reflect culturally specific understandings of mental illness and account for the broader social, historical, and environmental influences on health behaviors (Anakwenze, 2022; Caloudas *et al.*, 2023).

It is important to note that evidence for the added effectiveness of cultural adaptation in digital mental health is not unequivocal. The studies by Lorenzo-Luaces et al. (2021) and De Jesús-Romero et al. (2024) found that while cultural adaptation can improve acceptability and engagement, it does not consistently lead to superior clinical outcomes or cost-effectiveness compared to non-adapted interventions. This study, however, posits that deep-structure cultural adaptation remains critical in this context for several key reasons:

- Addressing Foundational Mismatch: The individualistic underpinnings of standard CBT represent a profound philosophical mismatch with the communitarian values prevalent in many African societies. A simple translation is insufficient; a deep-structure realignment is necessary for the intervention to be conceptually coherent.

- Prioritizing Engagement: In a context like South Africa, where stigma is a major barrier, the primary initial challenge is not necessarily efficacy but engagement. An intervention that is culturally resonant is more likely to be used, which is a prerequisite for any therapeutic effect.

- Beyond Surface-Level Changes: Many adaptations in the literature are limited to translation and surface-level changes. This study integrates the Ubuntu philosophy at a theoretical level, aiming to modify the therapeutic mechanisms themselves, which may yield different results from the adaptations critiqued in the literature.

Therefore, while the critique from Lorenzo-Luaces et al. (2021) rightly highlights the need for rigorous evaluation, it does not negate the value of a deep, theory-driven adaptation. In this case, adaptation is pursued as an essential strategy to ensure ethical alignment, relevance, and potential real-world impact, with its ultimate value to be determined empirically.

Cultural adaptation in mental health applications entails adding cultural content to ensure it appeals to people from different cultures. This is done by making mental health applications more in line with the beliefs and values of people from different cultures (Anakwenze, 2022; Spanhel *et al.*, 2021). Research suggests that culturally adapted interventions demonstrate moderately stronger effectiveness compared to non-adapted interventions, highlighting the significance of cultural tailoring in mental health interventions (Spanhel *et al.*, 2021).

### 2.2.1 Ecological Validity Model (Bernal et al., 1995)

Several studies have highlighted the impact that cultural adaptation can have on evidence-based interventions that we developed in different cultural contexts. Most of these studies use systematic frameworks to guide the adaptation. One popular framework is the Ecological Validity Model (EVM).

The Ecological validity model is a framework that was designed to guide the cultural adaptation of psychological interventions. Originally designed by Bernal et al. (1995), it serves as a foundational framework for the cultural adaptation of psychological interventions. The EVM emphasizes the importance of considering cultural context to ensure that interventions are meaningful and effective for the target population. The model focuses on eight dimensions to enhance the ecological validity, the extent to which the findings can be generalized to real-world settings of interventions. These dimensions include language, involving the use of culturally appropriate expressions; persons, referring to the inclusion of individuals who share ethnic or racial similarities with the target population; metaphors, which draw on expressions and symbols commonly shared within a cultural group; and content, which incorporates relevant cultural knowledge. They also encompass concepts, accounting for specific cultural, social, and environmental factors that influence the adaptation and effectiveness of an intervention; goals, ensuring alignment between the objectives of the intervention program and the expectations of the target population; methods, such as the mode of delivery used to implement the intervention; and context, which integrates the broader social and historical background of the population.

Systematically addressing these dimensions ensures that mental health interventions, including mental health applications, (1) are culturally responsive and (2) maintain the effectiveness of the interventions. Below are just a few examples of studies that performed a cultural adaptation of mental health interventions that lacked cultural context:

Sit et al. (2020) employed EVM to culturally adapt the step-by-step (SbS) digital mental health intervention for young Chinese adults. This process included key informant interviews with mental health experts and focus groups with potential beneficiaries, leading to significant modifications in metaphors, narratives, characters, illustrations, and the overall context of the program. These adaptations enhanced the program's cultural relevance, reduced stigma, and improved accessibility, demonstrating the model's effectiveness in guiding substantial cultural adaptations.

Shroff et al. (2023) followed the Cultural Adaptation Framework for Scalable Interventions by Heim and Kohrt to adapt Project Yes for youths in San Antonio. Initially developed in English, the program was later translated into Spanish post-adaptation, ensuring it met the cultural and linguistic needs of the community. This adaptation process, which utilized Single-Session Interventions (SSIs), was well-received by participants, who reported significant improvements in psychosocial outcomes.

Arjadi et al. (2018) adapted an internet-based behavioral activation intervention for depression to fit the Indonesian context using the Formative Method for Adapting Psychotherapy (FMAP). This study involved collaboration with stakeholders and the integration of cultural elements into the intervention, highlighting the potential of deeper-level cultural integration in existing interventions.

Sit et al. (2020) focused on culturally adapting the Step-by-Step (SbS) digital mental health intervention for Chinese young adults, employing the Ecological Validity Model (EVM) along with a detailed four-phase process. This process included conducting key informant interviews with mental health experts and focus group discussions with potential beneficiaries, ensuring a robust foundation for adaptation. The outcome was a successful cultural adaptation of the SbS program, which involved significant modifications to metaphors, narratives, characters, illustrations, concepts, goals, methods, and the context of the program. These adaptations made the program culturally relevant, aligned with local goals, reduced stigma, and increased accessibility(Sit *et al.*, 2020). The study underscored the importance of a structured framework and community input in the adaptation process, highlighting how these elements are crucial for the successful localization of mental health interventions.

Shroff et al. (2023) conducted a study to evaluate the effectiveness of Project Yes, a program aimed at improving psychosocial outcomes among youths in San Antonio (Shroff *et al.*, 2023). The project adaptation followed the Cultural Adaptation Framework for Scalable Interventions proposed by Heim and Kohrt, which emphasizes integrating cultural concepts of distress and adapting nonspecific intervention factors to enhance cultural relevance. This process involved collaboration with local youth stakeholders and healthcare providers. Initially developed in English, the interventions were translated into Spanish only after their adaptation to ensure that they met the community's cultural and linguistic needs. The methodology utilized Single-Session Interventions (SSIs), which are brief, self-administered sessions designed to address issues like depression, anxiety, and self-esteem, and could be completed in one sitting. Participants provided feedback before and after engaging with the SSIs, revealing that Project Yes was well-received and led to significant improvements in psychosocial outcomes, such as reduced hopelessness and self-hate and increased perceived agency. The completion rate for Project Yes was notably higher than previous interventions, confirming its applicability and convenience for both English and Spanish-speaking youths in San Antonio.

Davidson et al. (2015) developed Rise Above, a web-based depression intervention tailored for Latina/o youth facing high depression rates and low mental health service utilization. The study involved thematic interviews with national mental health experts and focused on adapting evidence-based treatments to the cultural patterns of the Latina/o population, successfully improving engagement and access to mental health services (Davidson *et al.*, 2015).

The application of the EVM in the aforementioned studies demonstrates its significant impact on enhancing the cultural sensitivity and relevance of mental health interventions. The examples highlight several critical findings.

**Cultural sensitivity and relevance:** Each study has shown that considering the cultural contexts and specific needs of the target population leads to interventions that are more than just effective- they are meaningful and resonate well with the participants. For instance, the adaptation made in the step-by-step program for Chinese young adults included significant changes to cultural metaphors and narratives, which made them more relatable and acceptable within that cultural framework.

**Community Engagement:** Successful cultural adaptation often involves the active participation of the community. For example, the development of the Project Yes program involved collaboration with local youth stakeholders and healthcare providers, which ensured that the intervention was culturally congruent and fostered a sense of ownership and trust among participants.

**Addressing Stigma and Cultural Considerations:** Incorporating cultural elements into the interventions helps in addressing the stigma associated with mental health issues within specific cultures. In Davidson et al.'s (2015) study, the culturally sensitive approach adopted for the Rise Above program helped reduce stigma and improve treatment engagement among Latina/o youth, a group traditionally underserved in mental health.

These findings underscore the importance of integrating cultural and social considerations into mental health interventions. By doing so, the ecological validity of the interventions is enhanced, and they also become more effective in real-world settings. This approach not only meets the clinical needs of individuals but also addresses the broader cultural and social dynamics that influence mental health outcomes. Moreover, these examples illustrate the utility of the EVM in bridging the gap between clinical effectiveness and cultural relevance, ensuring that mental health interventions are both scientifically sound and culturally appropriate. This synergy is essential for the successful implementation of mental health services, particularly in regions with diverse populations like South Africa, where cultural nuances significantly influence health behaviors and perceptions.

### 2.2.2 Cultural Adaptations from a Developer's Perspective

Spanhel et al. (2021) proposed a taxonomy of cultural adaptation, categorizing strategies into different levels based on the extent of modification. This taxonomy includes surface-level adaptations, such as changes to language and visual elements, aimed at enhancing intervention appropriateness, as well as deeper-level adaptations, which involve aligning intervention content, structure, and delivery methods with the cultural beliefs and values of the target population (Spanhel *et al.*, 2021). Moreover, the taxonomy underscores the importance of considering the cultural context, encompassing social determinants of health, historical trauma, and systemic barriers, to ensure comprehensive and effective cultural adaptation efforts (Spanhel *et al.*, 2021). Cultural adaptation in mental health applications entails adding cultural content to ensure it appeals to people from different cultures. This is done by making mental health applications more aligned with the beliefs and values of people from different cultures (Anakwenze, 2022; Spanhel *et al.*, 2021). Research suggests that culturally adapted interventions demonstrate moderately stronger effectiveness compared to non-adapted interventions, highlighting the significance of cultural tailoring in mental health interventions (Spanhel *et al.*, 2021). However, a review of cultural interventions in sub-Saharan Africa that used the ecological validity model (EVM) revealed that while some interventions reflected cultural sensitivity, few met the cultural compelling criteria (Anakwenze, 2022).

### 2.2.3 Synergy between Approaches

The interplay between culturally adapted and culturally grounded interventions offers a robust framework for addressing the complex cultural needs in mental health care. It is crucial to recognize that cultural adaptation is not a one-size-fits-all approach (Spanhel *et al.*, 2021). Instead, we argue that, in many cases, culturally adapted interventions could benefit more from incorporating elements of cultural groundedness. The Synergy between these approaches is essential for successful implementation. Finding the middle ground between these two approaches lies in understanding the nuances between culturally grounded and adapted interventions (Vernon, 2023).

### 2.2.4 Challenges in Cultural Interventions in Africa

Communities in Africa historically relied heavily on families and traditional healers to address mental health issues. However, colonization and other developments have impeded the capacity of traditional healthcare systems to provide comprehensive healthcare by criminalizing such practices (Anakwenze, 2022). This has led to a disconnect between traditional healing practices and modern mental health interventions, hindering access to culturally appropriate care for many individuals in the region.

Moreover, most psychological frameworks used in Africa have historically not been indigenous to the continent. African psychologies rely on values, constructs, beliefs, and methodologies native to African cultural groups to study behavior and mental processes. However, the prevalence of non-indigenous frameworks highlights the significant gap in culturally relevant mental health interventions (Anakwenze, 2022; Naeem *et al.*, 2019).

Western psychology and African psychology differ in their underlying philosophical frameworks, perspectives, and approaches to understanding human behavior and mental health. Western psychologies are primarily rooted in individualism, which emphasizes the significance of the individual's thoughts, emotions, and behaviors distinct from their broader social and cultural contexts. This perspective is influenced mainly by Eurocentric theories and frameworks, which shape the understanding of human behavior and mental health through a culturally specific lens. In practice, Western approaches often rely on standardized assessment tools and diagnostic criteria, which may not fully account for the socio-cultural variables that influence mental health. The focus is generally on individual well-being, with mental health defined by personal fulfillment and psychological stability.

In contrast, African psychologies are deeply embedded in the principles of communalism and interconnectedness, reflecting a collective approach to understanding human behavior. These psychologies are grounded in African cosmology, philosophy, and cultural practices, which prioritize the collective well-being of the community over the individual. The African perspective views social relationships as central to the concept of mental health, advocating for holistic and contextual approaches to diagnosis and treatment. Unlike Western models, which focus on the individual, African psychologists consider mental health to be intricately linked with the health of the community, emphasizing the need for culturally grounded interventions (Wilson & Williams, 2013).

Anakwenze (2022) emphasizes the issue, stating that the quest to develop indigenous psychologies is far from complete. Much of what has been accomplished to date in terms of psychological frameworks and tailoring interventions has entailed importing from the West. However, there remains a significant lack of psychological framework and interventions rooted in African culture, posing a challenge in evaluating whether interventions emerging organically from African contexts are more beneficial for Africans. Anakwenze underscores the importance of African psychologists collaborating with local communities to develop culturally informed theories rooted in the realities of Africans. Such theories may offer better support to communities in Africa compared to Western approaches like cognitive-behavioral therapy, which may not fully resonate with African contexts. Despite this, interventions from Western society continue to be valued and are likely to remain so for mental health practitioners. However, because developing African theories and psychologies is out of the scope of this research, it is advocated that existing Africa-centric theories and contexts should be integrated to enhance the effectiveness of these interventions, more specifically, mobile health interventions, to pave the way forward.

Another significant factor contributing to the mental health treatment gap is the conflation of Westernization with modernization. This conflation is perpetuated by the collaboration between Western agencies and institutions in Africa, resulting in the import and application of psychological concepts of mental illness. Some African psychologists and international partners may view indigenous knowledge as incompatible with "progress" rather than recognizing its potential to inform treatment developments alongside knowledge from Western societies (Anakwenze, 2022). This understanding underscores the need to integrate Africa-centric theories into mobile health interventions to pave the way forward.

For developing countries like South Africa to reap the benefits of adopting AI, innovation needs to be founded on the sociocultural factors that impact trust. Lack of trust can negatively impact the adoption of these interventions, potentially undermining the achievement of the anticipated benefits. Cultural context is crucial in building trust and acceptance (Vernon, 2023). For example, in some South African cultures, people do not use titles such as 'Mr.' or 'Mrs', but rather 'Baba' or 'Mma' to respectfully address elders. Incorporating these culturally appropriate titles into the intervention demonstrates respect and recognition of local customs, thereby fostering trust. This respect for cultural norms makes the intervention more relatable and trustworthy, enhancing its effectiveness and acceptance.

While significant progress has been made in addressing mental health challenges in South Africa, there remains a need for interventions that are deeply rooted in the cultural context of the population. By integrating Africa-centric theories and concepts into mental health interventions, partially mobile health interventions, we can bridge the treatment gap and pave the way for more effective and culturally sensitive approaches to mental health care. The taxonomy of cultural adaptation provides a systematic framework for guiding adaptation efforts, emphasizing the importance of considering the content.

## 2.3 Cognitive Behavioral Therapy in the Treatment of Anxiety and Depression

Most mHealth applications use ecological momentary intervention (EMI) to deliver treatments provided to people in their everyday lives. This approach captures and modifies specific moment-to-moment situations that emerge in the real world rather than targeting problematic thoughts, emotions, and behaviors through therapy sessions or in the hospital. These interventions are designed to address behaviors, thoughts, or emotions within the context of individuals' daily lives, allowing for timely and contextually relevant support (Marciniak *et al.*, 2020). Many EMI approaches employ evidence-based techniques such as cognitive behavioral therapy (CBT). CBTs are based on the theory that maladaptive cognitions, such as general beliefs and automatic thoughts about the self and the world, contribute to the maintenance of emotional distress and behavioral problems. These techniques are considered the "Gold standard" of treatment in many mental health conditions (Marciniak et al.). CBTs comprise techniques that address psychological mechanisms that underpin health conditions.

Four fundamental psychopathology techniques are used in cognitive behavioral therapy (Denecke *et al.*, 2022): Cognitive restructuring, behavioral activation, problem-solving, and exposure therapy. Cognitive restructuring is a technique that aims to identify and challenge negative thought patterns (Denecke *et al.*, 2022; Marciniak *et al.*), while behavioral activation encourages individuals to engage in positive activities to counteract depression and anxiety (Denecke *et al.*, 2022; Marciniak *et al.*, 2020). Problem-solving techniques help individuals develop effective coping strategies for managing stress and resolving interpersonal conflicts, and exposure therapy involves gradually confronting feared situations to reduce anxiety responses (Denecke *et al.*, 2022). Although exposure therapy is mentioned, this paper does not consider it as a mental alleviation technique because, although it aims to alleviate mental health symptoms, it is more focused on symptom reduction and management rather than general mental health alleviation. Various technologies can implement these techniques in the context of mobile health applications for mental health, enhancing the user experience and encouraging self-management.

Mobile health applications adapt and integrate these technologies into digital health formats, offering users interactive tools and resources for self-help and guided therapy. Cognitive restructuring exercises may include journaling and challenging exercises, while behavioral activation could involve activity scheduling features with some encouragement or reward mechanisms to promote a particular behavior (Arjadi *et al.*, 2018). Problem-solving modules may offer users step-by-step guidance for addressing specific challenges (Denecke *et al.*, 2022),

Keywan et al. (2023) designed DEPRA, a mass-screening conversational AI chatbot, to identify depression early and prevent potential crises. They utilized the Hamilton Interview Guideline (HIG), based on the Structured Interview Guide for the Hamilton Depression Rating Scale (SIGH-D), to structure the conversation flow, ensuring standardization and comparability of results. By converting multiple-choice questions into open-ended responses, the chatbot captured genuine feelings and thoughts, while scoring systems facilitated the collection and analysis of self-reported depressive symptoms. However, it is important to note that while DEPRA offers a convenient and accessible platform for mental health assessment and support-seeking, it does not directly contribute to the alleviation of depression symptoms or other mental health disorders. Despite these limitations, DEPRA demonstrates the potential of AI in bridging the gap between individuals and access to basic mental health support.

Similarly, Ahmadi et al. (2023) conducted a study to design and evaluate a smartphone-based cognitive behavioral therapy (CBT) program aimed at controlling and alleviating symptoms of anxiety and depression. Utilizing a 10-session randomized controlled trial, the researchers divided 45 participants into an intervention group, which received the CBT application, and a control group, which continued with conventional treatments. Assessments were conducted using Beck's Anxiety and Depression Scales, with data analyzed through SPSS using descriptive statistics and statistical tests such as the Shapiro-Wilk test and paired T-test. Results indicated significant improvements in the intervention group, with notable reductions in both anxiety (P-value = 0.001) and depression scores (P-value = 0.002) post-intervention, in contrast to the control group, which showed no significant changes in their scores (anxiety P=0.140, depression P=0.683). These findings underscore the effectiveness of the smartphone-based CBT program in improving mental health outcomes compared to conventional treatments.

Extending the application of technology in mental health, Hanna and Hanna (2021) developed an innovative smartphone application to improve university students' mental health and overall quality of life. Leveraging AI for sentiment analysis of users' social interactions and incorporating biofeedback as an evidence-based therapy, the app analyses social interactions across various platforms, facilitates connections with others who share similar moods, and provides relevant therapy tags. Moreover, the application fosters chat support among users with similar mental sentiments and grants access to carers, community volunteers, and health professionals during periods of significant loneliness or distress. Notably, the inclusion of self-monitoring features empowers users to track daily conditions such as appetite, exercise, sleep, and mood. Hanna and Hanna conducted a pilot test to assess the app's usability, involving the development of a prototype and gathering feedback through a usability questionnaire. Post-release, the performance of the application was evaluated in terms of downloads and feedback from the embedded online survey. One notable strength of this application lies in the various theoretical frameworks that fostered its development, demonstrating the ability to combine psychological theories and possibly African psychological theories into applications.

The effectiveness of Cognitive Behavioral Therapy (CBT) techniques within Ecological Momentary Interventions (EMIs) hinges on carefully considering the specific mental health conditions addressed and the contextual factors surrounding each intervention. For instance, the implementation of CBT techniques should account for cultural context. For example, behavioral activation may manifest differently in Western, individualistic cultures compared to African communal contexts. In an African context, behavioral activation might involve actions such as 'spending time with elders to listen to their ancient wisdom ', reflecting the cultural emphasis on respect for elders and traditional knowledge. In contrast, in a Western cultural context, behavioral activation might entail activities like 'scheduling a leisure activity or exercise sessions aligning with the individualistic focus on personal fulfillment and self-care. Therefore, selecting and adapting CBT techniques and tailoring them to match the target population's context and symptom severity is crucial for enhancing engagement and achieving therapeutic outcomes. Furthermore, the adaptation of these techniques should be guided by evidence-based practices and seamlessly integrated into users' daily lives, considering factors such as technological platforms and timing of intervention delivery. The studies by Ahmadi et al. (2023), Hanna and Hanna (2021), and Keywan et al. (2023) illustrate this well. They demonstrate significant improvements in mental health outcomes through tailored digital interventions, confirming the capacity of AI and smartphone applications to bridge the gap between clinical needs and technological solutions effectively. These studies underscore the transformative potential of digital technologies and AI in enhancing mental health interventions, showcasing innovative approaches to treatment accessibility and effectiveness across different settings and populations.

## 2.4. Techniques for Incorporating Emotional Intelligence in LLM

Emotional intelligence denotes the capacity to adeptly interpret and manage emotion-infused information, subsequently harnessing it to steer cognitive tasks, ranging from problem-solving to behavior regulation (Salovey *et al.*, 2009). In the context of LLMs, this study defines emotional intelligence as the capacity of an LLM to accurately understand and respond to emotions expressed in text data, enabling better problem-solving and behavior regulation (such as response generation), taking into account the context of the user.

### 2.4.1 Prompt engineering

Prompt engineering refers to the design of instructions (prompts) aimed at enhancing the performance of existing language models (Reynolds & McDonell, 2021). Unlike traditional paradigms that necessitate model retraining or extensive fine-tuning for task-specific performance, prompt engineering offers adaptability by steering model responses (Sahoo *et al.*, 2024). Techniques encompass a spectrum from foundational methods like zero-shot and few-shot prompting to more intricate approaches such as "chain of code" prompting. Zero-shot prompting eliminates the need for extensive training data, relying instead on meticulously crafted prompts to guide models toward novel tasks, leveraging pre-existing knowledge for predictions based on specific input-output mappings. In contrast, chain-of-code prompting prompts models to generate sequences of code based on given prompts. While a comprehensive discussion of these techniques exceeds the scope of this review, interested readers can explore detailed explanations in articles such as "A Systematic Survey of Prompt Engineering in Large Language Models: Techniques and Applications" (Sahoo *et al.*, 2024). Prompt engineering emerges as a promising technique to enhance LLMs' performance on downstream tasks.

### 2.4.2 Emotionally Intelligent Prompting

Building on the foundation of prompt engineering, Li et al. (2024) introduced the Emotional Chain of Thought (ECoT), a novel prompting method that integrates emotional intelligence principles into LLMs. Inspired by Goleman's theory, EcoT aims to activate the emotional intelligence of LLMs by providing quality step-by-step prompts based on a theoretical framework. This method demonstrates the potential for LLMs to understand and generate responses based on emotional data and engage in more human-like interactions. Complementing this, Li et al. (2023) explored enhancing LLMs by integrating emotional stimuli into prompts, conducting 45 experiments across 35 tasks with models such as Flan-T5-Large, Llama 2, Vicuna, BLOOM, and ChatGPT. Their introduction of 'EmotionPrompt,' a technique that blends original prompts with emotional stimuli, led to significant performance improvements, including an 8% increase in Instruction Induction and a 115% improvement in BIG-Bench tasks, notably enhancing the models' creative output and demonstrating profound implications for integrating psychological insights into AI development for deeper and more meaningful human-like interactions.

Similarly, Wei et al. (2022) investigated how large language models' reasoning abilities could be enhanced through chain-of-thought (CoT) prompting. By providing models with intermediate reasoning steps leading to a final answer, they demonstrated significant improvements in tasks such as arithmetic, symbolic reasoning, commonsense reasoning, and mapping natural language instructions to robot actions. The authors manually composed CoT exemplars to guide the models' reasoning, revealing that explicit reasoning chains substantially enhanced performance across diverse tasks. These findings complement Li et al.'s exploration of emotional intelligence integration in LLMs.

### 2.4.3 Integration of Emotional Dimensions in Chatbots

Innovative approaches to integrating emotional dimensions in LLMs have shown significant promise in enhancing the machines' ability to generate nuanced and culturally aware emotional responses. The Six-Dimensional Emotion (6DE) model proposed by Ratican & Hutson (2023) is one such framework that includes dimensions such as arousal, valence, dominance, agency, fidelity, and novelty, enhancing AI's ability to understand and react in emotionally complex situations.

Consistent with the aforementioned approach of enhancing AI's emotional intelligence through emotional dimensionality (Wortman & Wang, 2023), a study was conducted to develop the High-Coverage Emotion Model (HICEM). This model aims to create a comprehensive human emotion model that significantly advances the training of social robots and other intelligent machines, enabling them to engage in deeper and more meaningful human-machine interactions. The methodology involved generating a list of emotion concepts, employing UMAP (Uniform Manifold Approximation and Projection) for dimensionality reduction, and applying hierarchical clustering to categorize these emotions systematically. The effectiveness of the HICEM model was assessed using custom evaluation metrics and a comprehensive user study, which highlighted its superior coverage of emotional states with fewer components than existing emotion models. The model demonstrated robust performance across multiple languages, achieving average coverage scores of 0.458 in English, 0.349 in Arabic, 0.315 in Mandarin, 0.455 in French, 0.458 in Spanish, and 0.501 in Russian. These results underscore the model's capability to capture and categorize emotions across diverse cultural contexts accurately.

The studies suggest that the essence of incorporating emotional intelligence into chatbots and Large Language Models lies fundamentally in accurately identifying key emotions and strategically using these insights to inform response generation. While embedding sophisticated emotional frameworks within these systems allows AI to recognize and utilize a spectrum of human emotions, this approach has its faults. One significant challenge is selecting which emotions to prioritize, as the spectrum of human emotions is inherently complex. Deciding which emotions are important often leads to oversimplifications (Wortman & Wang, 2023), and capturing the full breadth of emotional experiences in a framework can be unfeasible. Moreover, different cultural contexts may interpret or value emotional expressions differently, adding another layer of complexity to the task.

While prompt engineering is a relatively recent innovation, it has showcased remarkable efficacy in tailoring LLMs for specific tasks, particularly in scenarios where extensive datasets are lacking. The studies discussed above illuminate various avenues for enhancing LLM performance: (1) by framing LLM responses to better align with contextual nuances, (2) by leveraging established theories to guide model behavior appropriately, and (3) by exploring diverse perspectives on emotional intelligence integration. These approaches collectively underscore the dynamic nature of adapting LLMs, offering promising pathways for further advancements at the intersection of AI and psychology.

### 2.5 Fine Tuning

Fine-tuning is the process in machine learning where a pre-trained model is further trained on a specific, often smaller, dataset to adapt it to a particular task or domain (Lu *et al.*, 2024). This technique leverages the model's existing knowledge to better achieve performance on specialized applications without training a new model from scratch.

The key difference between fine-tuning and prompt engineering (discussed earlier) is that fine-tuning modifies the model's internal weights, effectively adapting the foundational model to better align with the target dataset. In contrast, prompt engineering does not alter the model itself but instead optimizes the way inputs (or prompts) are crafted to elicit desired responses. While prompt engineering influences the model's outputs through carefully designed instructions, fine-tuning permanently refines the model's behavior based on task-specific data.

Research highlights several benefits of fine-tuning. One major benefit is improved accuracy and efficiency. Fine-tuning allows models to learn domain-specific patterns, enhancing their precision in applications. By focusing on relevant data, fine-tuning models provide more reliable outputs and improve decision-making. Fine-tuning also reduces computational costs. Training large language models from scratch demands significant time and resources, whereas fine-tuning builds on existing models with minimal additional data and computation. This makes AI more accessible and cost-effective for specialized use cases. In this section of the literature review, we give an overview of popular fine-tuning techniques and then discuss how fine-tuning has been applied to applications in the mental health space.

### 2.5.1 Overview of fine-tuning techniques

Fine-tuning techniques vary depending on computation constraints, data availability, and the level of adaptation required for specific tasks. Below are some of the most commonly used fine-tuning approaches.

One common approach is full-tuning, in which all pre-trained model parameters are updated based on the task-specific dataset. This method offers the highest degree of specialization but requires significant computational resources and a large dataset to prevent overfitting. Another technique is parameter-efficient fine-tuning (PEFT), which aims to fine-tune models with fewer parameters to optimize computation efficiency (Cui *et al.*, 2022). Techniques such as Low-Rank Adaptation (LoRA) and adapters selectively update only specific layers or a subset of parameters while keeping the rest of the model frozen, reducing memory usage and speeding up training (Hayou *et al.*, 2024).

A more advanced approach is reinforcement learning from human feedback (RLHL) (Abubakar *et al.*, 2024), in which a model is fine-tuned based on human-generated feedback. These optimizations align the model with human values, ethical considerations, and stylistic preferences. RLHF has been widely applied in conversation AI and content moderation systems where user alignment and fairness are crucial factors.

These fine-tuning techniques have advantages and trade-offs depending on the computational budget, training data availability, and desired model adaptability. Understanding these methods is critical to selecting the most effective approach for a given application. The next section explores how fine-tuning has been applied in healthcare, particularly in mental health applications.

### 2.5.2 Fine-tuning in mental health applications

Fine-tuning LLMs has emerged as a pivotal approach in developing mental health applications, enabling models to perform specialized tasks with greater accuracy and relevance. Several studies have explored the application of fine-tuning techniques to tailor LLMs for mental health support, assessment, and intervention (Lawrence *et al.*, 2024).

One notable study introduced MentalQLM (Shi *et al.*, 2024), a light language model specifically designed for mental healthcare applications. The development process involved two key stages. This approach aimed to create a model capable of understanding and generating responses pertinent to mental health contexts, thus enhancing the quality of AI-driven mental health support.

The Mental-LLM study evaluated multiple LLMs on various mental health prediction tasks using online text data. The researchers concluded that the experiments covered zero-shot prompting, few-shot prompting, and instruction fine-tuning. The findings indicated that instruction fine-tuning, few-shot prompting, and instruction fine-tuning. The findings indicated that instruction fine-tuning significantly boosted the performance of LLMs across all tasks, highlighting the effectiveness of fine-tuning in adapting LLMs for mental health applications (Xu *et al.*, 2024).

In a pilot study, researchers explored aligning LLMs to enhance psychiatric interviews through symptom delineation and summarization. The study investigated whether LLMs could identify parts of a conversation suggesting psychiatric symptoms and summarize stressors and symptoms based on interview transcripts. The results demonstrated that fine-tuned LLMs, or those with appropriate prompting, achieved high accuracy in symptom delineation and summarisation coherence, indicating their potential to assist mental health practitioners in analyzing psychiatric interviews (So *et al.*, 2024).

Collectively, these studies underscore the potential of fine-tuning techniques in enhancing the capabilities of LLMs for mental health applications, paving the way for more effective, accessible, and secure AI-driven mental health support systems.

## 2.6 Research Gaps

Despite significant advancements in mental health applications aimed at bridging accessibility gaps, several critical research gaps remain that limit the efficacy and reach of these interventions, particularly in culturally diverse settings like South Africa.

Most existing mental health applications are grounded in Western psychological theories and practices, which may not be entirely applicable or acceptable across different cultural contexts. This reliance on Western-centric approaches has limited the development of interventions that are culturally congruent with the needs and values of non-Western populations. Studies such as those by Davidson et al. (2015) emphasize the importance of integrating a patient's cultural contexts and values into evidence-based treatments (EBTs), suggesting a significant gap in the cultural tailoring of mental health solutions.

To our knowledge, no substantial study has been conducted focused on the cultural adaptation of mental health applications, specifically in South Africa. This oversight is significant because cultural adaptation has been shown to increase the acceptability and effectiveness of interventions in marginalized populations by incorporating relevant cultural stories and perspectives (Shroff *et al.*, 2023)

While chatbots have demonstrated remarkable potential in mental health settings, such as early detection of depression and alleviation of anxiety, they often do not consider cultural perspectives in their emotional intelligence algorithms (Ahmadi *et al.*, 2023; Kaywan *et al.*, 2023). Given the nuanced nature of emotional expression across different cultures, this limitation is a significant oversight. Culturally adapted digital tools, which take into account the specific cultural and spiritual dimensions of their users, could offer enhanced support, especially in low-resourced or minoritized communities that frequently rely on digital resources for health information (Shroff *et al.*, 2023). Moreover, incorporating spirituality, a vital element of African psychology, has significantly improved mental health outcomes, as evidenced by research conducted within Indonesian Muslim communities (Hamka *et al.*, 2022). This finding underscores the potential benefits of integrating spiritual and cultural elements into mental health interventions for similar improvements in contexts like South Africa.

In low-and middle-income countries, managing mental health problems remains a formidable challenge due to the limited availability of treatment. The need for culturally sensitive interventions is particularly acute in these settings, where traditional and cultural beliefs often play a significant role in the perception and management of mental health.

The highlighted gaps indicate a pressing need for more inclusive research that considers the cultural, spiritual, and contextual nuances of diverse populations, particularly in regions like South Africa. Addressing these gaps could lead to more effective, culturally appropriate mental health interventions that are better suited to the populations they aim to serve.

## 3. Research Methodology

This section presents the methodology adopted in this study, detailing both the theoretical foundations and practical implementation of a culturally adapted, AI-driven mental health support framework informed by the Ubuntu philosophy. It begins by introducing the deep-structure cultural adaptations rooted in Ubuntu, emphasizing communal values, spiritual grounding, and interdependence. This is followed by a discussion of surface-level adaptations, which address communication style and interaction design to ensure cultural resonance in user engagement. The adaptation of Cognitive Behavioural Therapy (CBT) techniques—specifically Behavioural Activation and Cognitive Restructuring—is then explored, highlighting how these were reinterpreted through an Ubuntu-informed lens to align therapeutic goals with African cultural values.

The methodology further outlines the research design, including the overarching methodological paradigm, system architecture, and the process for culturally adapting the CACTUS dataset through a five-step transformation pipeline. The development of expert-informed case studies is also described, designed in collaboration with a clinical practitioner to authentically simulate how individuals in African contexts might express symptoms of depression and anxiety. The section concludes with a description of the experimental setup, detailing the model fine-tuning process, evaluation metrics, and the use of both automated (e.g., UniEval) and manual assessment methods. Collectively, these components establish the methodological foundation of the study, demonstrating how theoretical principles, system design, culturally adapted data, and evaluation strategies converge to develop an emotionally intelligent and culturally responsive AI-based mental health support system.

## 3.1 Deep-level structure changes

### 3.1.1 Theoretical framework: Ubuntu

Ubuntu is a philosophical concept rooted in Southern African traditions that articulates the essence of being human (Obasa, 2024). It emphasizes interconnectedness, interdependence, and communal responsibility, core principles that stand in contrast to the individualistic orientation of many Western psychological models. In Ubuntu, a person's identity is not defined in isolation but concerning others: "I am because we are". This collective consciousness forms the foundation for an African-centred model of mental health and well-being (Wilson & Williams, 2013).

One of the most compelling aspects of Ubuntu is its explicit inclusion of dimensions often overlooked by Western psychologies, particularly spirituality and connection with divinity. While many Western frameworks prioritize intrapersonal cognition and behavior (Anakwenze, 2022), Ubuntu recognizes that well-being is deeply tied to spiritual, communal, and ancestral dimensions of life. Importantly, the reference to ancestry in Ubuntu does not imply ancestral worship as a religious practice but rather speaks to a deep reverence for one's origins, roots, and cultural lineage. This sense of rootedness offers grounding, identity, and a sense of purpose, especially in contexts where healing is linked to reconnecting with one's heritage.

Ubuntu also addresses Africa-specific expressions of distress and resilience, which are often absent from conventional psychological paradigms. For instance, self-worth in many African communities is closely tied to one's contribution to the collective, specifically to the family, the community, and society. This cultural emphasis is reflected in Ubuntu's construct of competency, which moves beyond personal success to include responsibility, good behavior, and communal value. Ubuntu thus offers a culturally aligned model of well-being and a lens through which to understand behaviors and motivations that might otherwise be misinterpreted or pathologized by Western standards.

Wilson and Williams (2013) offer a structured framework for Ubuntu that revolves around three central constructs: Connectedness, Competency, and Consciousness. These constructs provide a conceptual foundation for adapting psychological interventions, such as CBT, to resonate with African cultural contexts. The following subheadings further explain in detail the three aforementioned central constructs.

**Connectedness**

This construct emphasizes the interdependence and social bonds that link individuals within a community. Connectedness is more than social interaction. It is spiritual, emotional, and deeply relational. It reflects a commitment to empathy, compassion, and mutual care. Ubuntu sees well-being as emerging from a sense of belonging, emotional closeness, and recognition of others as part of oneself. This includes people in one's immediate environment and broader spiritual ties, including connection with divinity (Wilson & Williams, 2013).

**Competency**

Competency refers to the skills, behaviors, and personal development necessary for individuals to function meaningfully in society. This includes making responsible choices, demonstrating good behavior, aspiring toward future goals, and recognizing one's unique contributions. Within Ubuntu, competency is not solely individualistic; it is measured by how effectively one contributes to the collective well-being. This perspective reinforces a culturally grounded sense of self-worth, where personal identity is enriched through communal impact and relational integrity (Wilson & Williams, 2013).

**Consciousness**

Consciousness in Ubuntu encompasses self-awareness, reflection, and understanding one's role within a broader sociocultural and historical context. From an African-centered viewpoint, it also integrates spiritual awareness and the alignment of purpose in life. Consciousness is personal and collective: it is the awareness of who one is, where one comes from, and how one's actions affect the community. This construct acknowledges that emotional and psychological health cannot be fully understood without recognizing cultural identity, history, and communal responsibilities (Wilson & Williams, 2013). These constructs and their underlying tenets are summarized in Table 1.

Table 1: Tenets of Ubuntu

| Construct | Core Tenets |
|---|---|
| **Connectedness** | - Social Bonds <br> - Relationships |

| | • Sense of Belonging |
| | • Connection with Divinity |
| **Competency** | • Personal Development |
| | • Responsible choices |
| | • Good Behavior |
| | • Personal Responsibility |
| | • Future Aspirations |
| | • Recognition of Individual Uniqueness |
| **Consciousness** | • Self-Awareness |
| | • Mindfulness |
| | • Understanding One's Place in a Broader Social and Cultural Context |

### 3.2 Surface-level structure changes

Table 2 details the specific modifications implemented to enhance the quality and cultural appropriateness of interactions between the LLM and the user. These adaptations draw their inspiration from the Ecological Validity Model (EVM) (discussed in Section 2), Spanhel's (2021) taxonomy for culturally adapting digital mental health interventions, as well as key linguistic and contextual considerations. For instance, research indicates that most South Africans do not speak English as a first language (Mncwango & Makhathini, 2021). As a result, the use of overly complex or "fancy" English may inadvertently reduce user engagement and undermine the effectiveness of the intervention.

**Table 2: Adaptation of communication**

| Component | Description | Purpose |
|---|---|---|
| Use of a Familiar Name | Assign a name that is familiar and positively regarded within South African communities. | Increases trust, familiarity, and user comfort. |
| Use of Metaphors | Incorporate culturally resonant metaphors to explain complex ideas | Enhances understanding and emotional engagement. |
| Use of Appropriate Titles | Employ respectful titles as per user preference and cultural norms. | Demonstrates respect and adherence to local social hierarchies. |
| Integration of Spiritual elements | Embed relevant spiritual or religious references aligned with users' beliefs. | Promotes comfort and resonance with users' faith and worldview. |
| Use of Simple Language | Employ clear, straightforward language; optionally support local language translations. | Ensures accessibility for users with varying English proficiency. |
| Use of Somatic Descriptions | Replace clinical terms (e.g., "depressed", "anxiety") with everyday physical/emotional expressions (e.g., "tired", discouraged"). | Reduces stigma and increases relatability through culturally familiar terminology. |
| Goal-Oriented Communication | Ensure each interaction with the LLMs actively contributes to symptom alleviation. | Aligns digital interactions with the therapeutic goal of reducing anxiety and depression symptoms. |

These surface-level adaptations to communication serve a critical function beyond linguistic familiarity; they are designed to enhance user receptiveness, build trust, and ensure cultural adherence to the interaction between the user and the LLM. By integrating culturally meaningful language metaphors and cultural norms, these adaptations help users feel recognized and respected within their sociocultural context. This sense of cultural alignment significantly increases user comfort and engagement, making individuals more likely to interact openly, interpret the guidance meaningfully, and maintain continued use of the support tool. In essence, cultural familiarity becomes a foundation for psychological safety, which is a prerequisite for effective mental health support.

### 3.3. Adapting CBT to Align with Ubuntu

This study aims to synergize CBT with the Ubuntu framework previously outlined through deep and surface-level structure changes. Rather than applying CBT in its standard Western form, this section demonstrates how two core techniques, behavioral activation and cognitive restructuring, can be meaningfully aligned with Ubuntu's principles of community, spirituality, and shared humanity.

While CBT encompasses a variety of evidence-based strategies, the focus here is on behavioral activation and cognitive restructuring due to their cultural compatibility and relevance in communal societies.

### 3.3.1 Rationale for Technique Selection

Research in cross-cultural psychology suggests that the effectiveness of therapeutic techniques varies across cultural contexts, depending on how societies are organized along the spectrum of individualism and collectivism. Western cultures, where CBT was initially developed, are largely individualistic, the East is more collective, and Africans are more communal, prioritizing interdependence, family roles, spiritual meaning, and social responsibility (Anakwenze, 2022; Naeem *et al.*, 2019).

Within communal societies, individuals are often more responsive to therapeutic approaches reflecting relational and value-driven life dimensions. Research indicates that behavioral activation and problem-solving-based techniques (which significantly overlap with cognitive restructuring) are especially effective in these settings (Anakwenze, 2022; Naeem *et al.*, 2019). These strategies can be adapted to emphasize culturally meaningful activities and communal modes of thought, enhancing both relevance and engagement. Therefore, this study focuses on two techniques because of their therapeutic utility and because they offer the clearest pathways for cultural synergy between CBT and Ubuntu.

### 3.3.2 Behavioral Activation through Ubuntu

Traditional CBT uses behavioral activation to encourage individuals to re-engage with activities that foster enjoyment or a sense of accomplishment, such as exercise, journaling, or hobbies. These activities, however, tend to reflect individualistic values centered around personal pleasure or self-care. However, through the lens of Ubuntu, behavioral activation (BA) includes activities such as social participation, prayer, and culturally meaningful practices aimed at promoting restoration and well-being.

### 3.3.3 Cognitive Restructuring

Cognitive restructuring is typically used in CBT to identify and challenge distorted or unhelpful thoughts through logical analysis and introspective questioning. While effective, this approach often emphasizes self-referenced cognition, which may not fully resonate in communal cultures where social and spiritual interrelations shape thought patterns.

When viewed through the perspective of Ubuntu, cognitive restructuring transforms from a self-centered exercise into a relational, value-oriented process. The questions used to challenge negative thoughts are adapted to incorporate communal perspectives, ancestral wisdom, and social implications.

For example, traditional CBT may ask: What is the evidence for this thought, and how else might this situation be viewed? In Ubuntu-informed cognitive restructuring, these questions are reframed to consider relational and communal perspectives, such as how one's family might view the situation and how one's thoughts or actions may affect those around them. This approach helps individuals reframe their thinking through communal identity, shared wisdom, and interpersonal accountability. Rather than seeking only internal coherence, users are guided to consider the relational ripple effects of their thoughts and behaviors, consistent with Ubuntu's emphasis on shared humanity and ethical responsibility.

### 3.4 Research Design

This section presents the research design adopted in this study, detailing the overall structure, process, and methodological choices made in developing and evaluating the proposed framework. The section begins by outlining the research paradigm and methodology, grounding the study in an interpretivist viewpoint and a Design Science Research (DSR) approach focused on artifact creation. It then introduces the system architecture, providing a technical overview of the frontend and backend components and how they interact with a fine-tuned language model. Following this, the dataset creation and adaptation process is described, beginning with preprocessing the CACTUS dataset and leading to a five-step cultural adaptation pipeline. The adapted dataset is then evaluated for clinical appropriateness through an expert's review. Finally, the section concludes with the development of case studies used to simulate user interactions and assess the system's responsiveness to Africa-centric expressions of emotional distress. Together, these elements form a cohesive research design to operationalize a culturally sensitive AI mental health support system inspired by Ubuntu and CBT.

### 3.5. Research Paradigm and Methodology

This study adopts an interpretivist paradigm, recognizing the subjective nature of human experiences and the construction of these experiences through the meanings individuals assign within different contexts (Alharahsheh & Pius, 2020). Although the evaluation of most criteria is quantifiable, the assignment of the respective scores within the framework remains subjective and may vary from person to person.

This study employs a Design Science Research (DSR) methodology, which involves developing innovative IT artifacts to solve real-world problems (Weber, 2010). The DSR aims to create effective solutions through innovation and draws from existing knowledge to address complex issues (Peffers *et al.*, 2007). In this context, the artifacts are designed to address mental health challenges by incorporating Ubuntu philosophy and CBT principles, interacting with the cultural and psychological elements of the problem context (Wieringa, 2014).

### 3.5.1 System Architecture

Figure 1 illustrates the flow of the system. The architecture comprises a frontend built with Angular and a backend powered by Python Flask, integrated with a fine-tuned LLM. The user engages with the system via a chat-based interface on the Angular front end. When a message is sent, it is transmitted to the Flask backend via a RESTful API call. The backend processes the request by preprocessing the input and constructing the API call required to interact with the LLM. This construction includes setting key model parameters such as the system_prompt, top_p, frequency_penalty, and max_tokens. The system prompt plays a critical role in guiding the model's response. It is used for persona priming, establishing the context of the conversation, and outlining the steps or therapeutic logic the model should follow in addressing the user's input. Thus ensuring that the LLM responds in a manner that is not only technically coherent but also culturally grounded and therapeutically aligned.

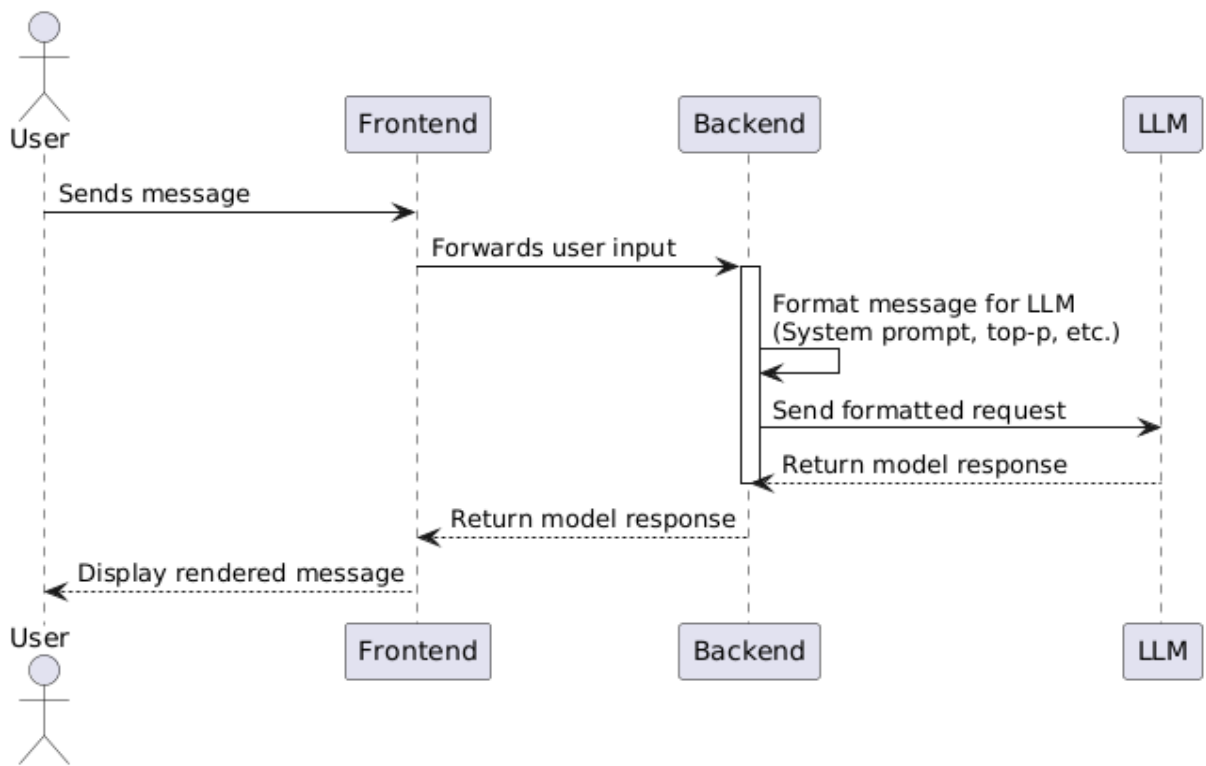

**Figure 1:** System Architecture

### 3.5.2 Frontend Design

Figures 2, 3, and 4 below are visual illustrations of how the frontend design looks. When the application loads, the user is greeted with a welcome message. This greeting can be customized based on the user's language preference, which is selectable from the dropdown menu located in the top right corner. For example, if the user

selects Setswana, the greeting changes to "Dumela, kenna Karabo", which translates to "Hi, I'm Karabo." (See Figure 3 for illustration.) This functionality is not AI-powered; the greetings are pulled from a predefined static list within the application. The design of the frontend aims to be both familiar and user-friendly, inspired by platforms like WhatsApp to ensure ease of use and relatability.

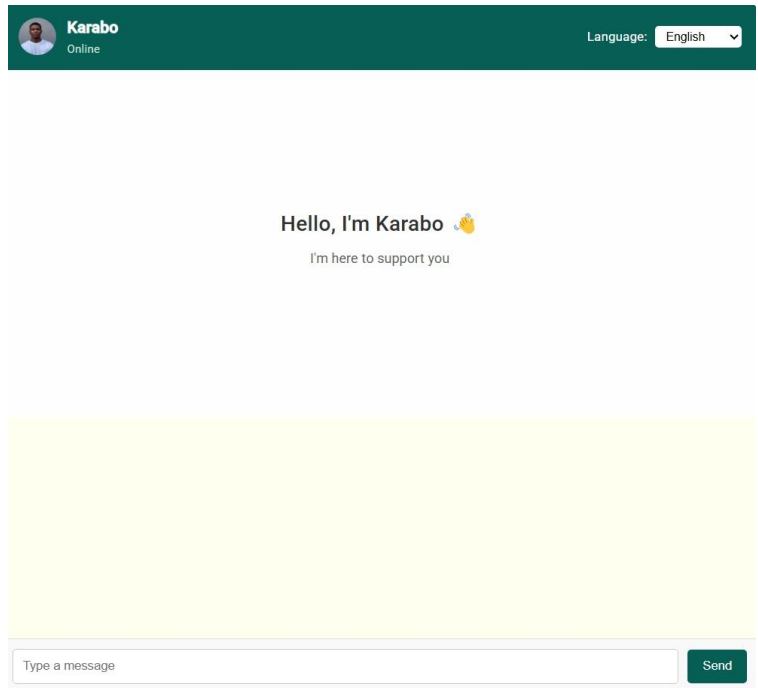

**Figure 2:** Frontend Landing Page

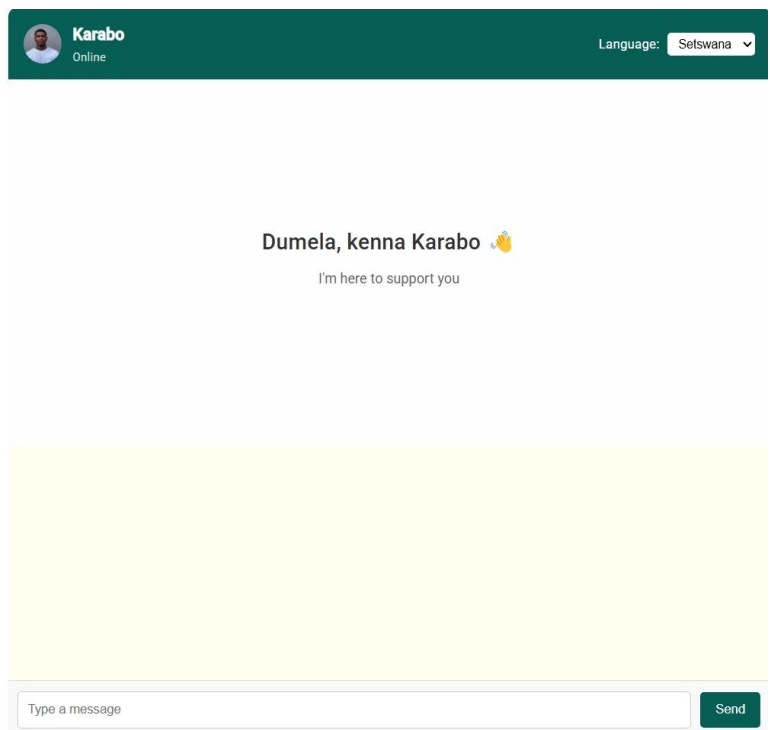

**Figure 3:** Frontend with language preference set to Setswana

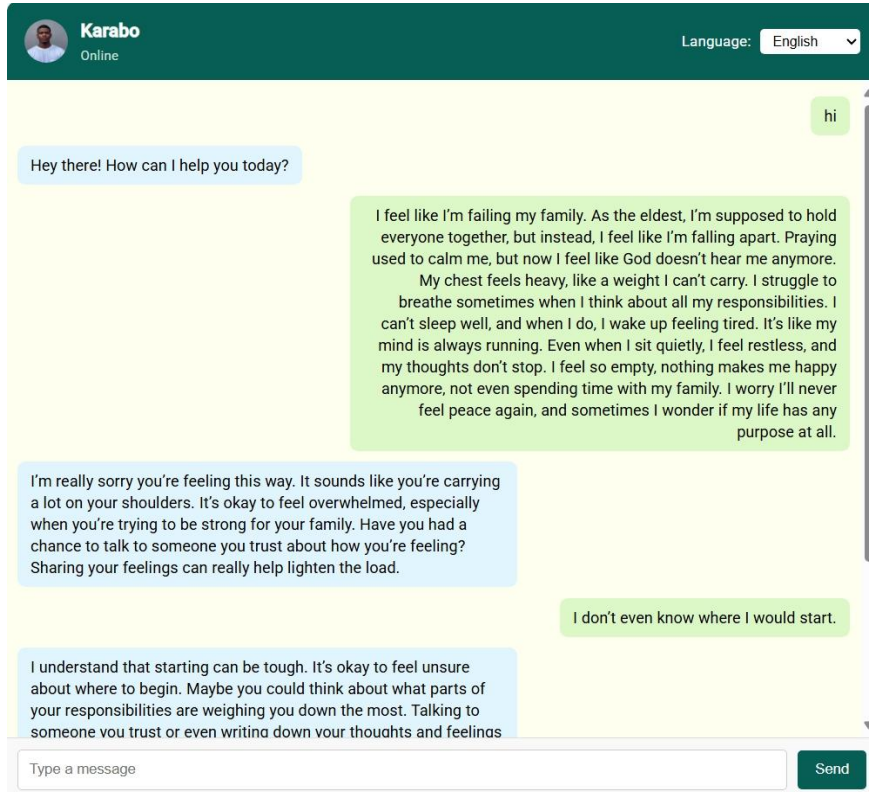

**Figure 4:** Sample User Interface of the Chat Conversation

### 3.5.3 Dataset Creation

**Dataset Description**

The original dataset used is the Cactus dataset from an article titled '*CACTUS: Towards Psychological Counselling Conversations using Cognitive Behavioral Theory*' (Lee *et al.*, 2024). It is a synthetic large-scale dataset designed to simulate realistic multi-turn psychological counseling conversations based on CBT. The dataset consists of approximately 31,000 records, with each row representing a full therapy session. Each session includes seven key attributes, as presented in Table 3 below.

**Table 3:** Cactus dataset attributes

| Attribute | Description |
| --- | --- |
| Attitude | Describes how the client behaves during the session |
| Thought | The client's main negative belief or worry at the start of the session |
| Dialogue | The full conversation between the counselor and the client |
| Cbt_technique | The method the counselor uses to help the client |
| Patterns | The negative thinking the client is showing |
| Intake_form | A summary of the client's background and why they're seeking help |
| Cbt_plan | A step-by-step plan that the counselor makes before the session |

**Dataset Adaptation Process**

The adaptation of the CACTUS dataset served two primary goals:

1. To address the lack of CBT datasets rooted in Africa-centric realities, a culturally reflective intermediary dataset is created.
2. To tailor the dataset to align with the proposed framework for culturally sensitive AI, incorporating both deep-structure and surface-level cultural adaptations.

The adaptation process was divided into two main phases: preprocessing and cultural adaptation, as illustrated in Figure 5 below.

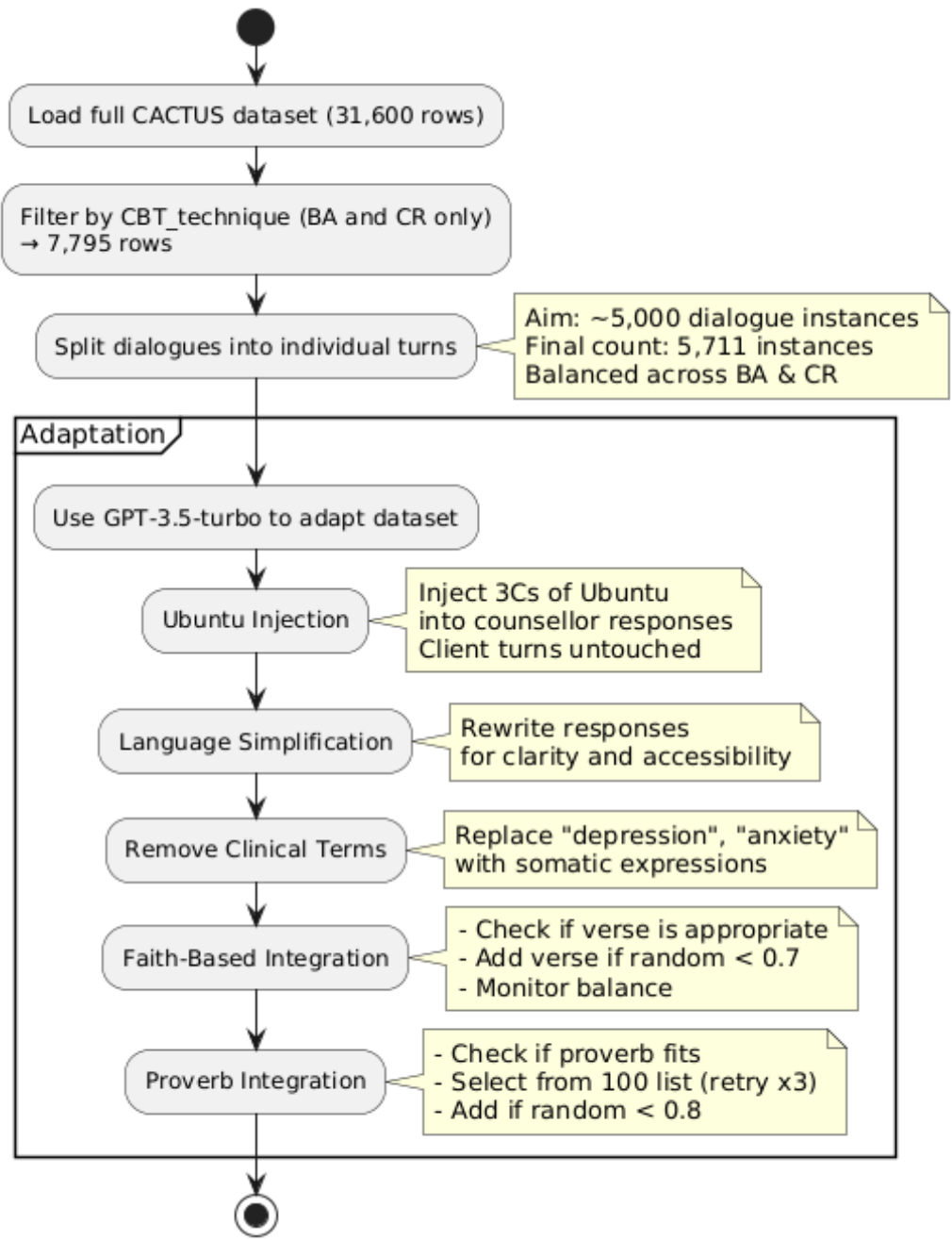

**Figure 5:** Adaptation of Dataset

**Preprocessing Phase**

As detailed in earlier sections, the original dataset was filtered using the cbt_technique attribute to retain only entries aligned with BA (such as Behavioural Experiment, Activity Scheduling, and Systematic Exposure) and Cognitive Restructuring (CR) (including Reality Testing, Reframing, and Positive Reframing), the two CBT strategies central to this study. The filtered data was then segmented into single-turn dialogue instances between the client and counselor, resulting in a final dataset of 5,711 cases balanced across BA and CR categories.

*Cultural Adaptation*

Using the pre-processed data described above. We opted to use GPT-3-turbo to adapt the dataset. The rationale behind using it is that LLMs have been shown to understand the context and follow instructions (Bommasani *et al.*, 2021). This also helped us automate the adaptation process with minimal human intervention. The adaptation process is 5-fold, namely, Ubuntu injection, Language simplification, Removal of clinical terms, Faith-based integration, and Proverb integration. Subsequently, we provide a brief explanation of each of the adaptation processes below.

i. **Ubuntu injection**

This step aims to integrate Ubuntu into the dataset per the 3Cs (Wilson & Williams, 2013), which was introduced into the counselor's responses. The 3Cs in mental health (especially in therapy and resilience models) denote Control (focus on what you can control), Commitment (stay committed to meaningful goals and relationships), and Challenge (view stress or change as a challenge rather than a threat), respectively.

To achieve this, we provided the model with a prompt that describe the concept of Ubuntu and also presented a dialogue instance between the client and counselor. The model was tasked with adapting the counselor's responses to align with Ubuntu. Importantly, only the counselor's responses were modified, as our primary interest was in aligning the counselor's responses. The client's responses were left unchanged. This step produced the first iteration of the adapted dataset with Ubuntu-infused responses.

ii. **Language Simplification**

This step aimed to simplify the language used by the counselor to ensure it was transparent and inclusive. The rationale was based on an article indicating that many South Africans do not speak English as their first language (Mncwango & Makhathini, 2021). Using overly complex or "fancy" English could undermine the counselor's communication ability. As a result, counselor responses were rewritten to be more straightforward to understand.

iii. **Removal of Clinical Terms**

In this step, clinical labels such as "depression" and "anxiety" were replaced with somatic descriptions that focus on the physical manifestations or experiences related to these conditions. The goal was to lessen the stigma attached to clinical terminology, making the counseling conversations more approachable and less intimidating. Research has shown that stigma is a key factor discouraging Africans from seeking help, making this modification a crucial step toward reducing disparities in mental health service utilization.

iv. **Faith-Based Integration**

South Africa's most recent census revealed that most of the population identifies as Christian (Africa, 2022). Based on this, it is safe to assume that the client is probably Christian; therefore, we incorporated faith-based support into the dataset to align with users' likely beliefs.

This step had two components:

- The LLM reviewed each dialogue instance and determined whether adding scripture-based comfort would enhance the counselor's response (response: yes or no).
- If scripture was deemed helpful, the model integrated relevant verses into the counselor's reply.

A probability threshold mechanism inspired by the Monte Carlo simulation was introduced to prevent excessive modifications. For each instance where scripture integration was considered appropriate:

A random number between 0 and 1 was generated.

- If the number fell within the threshold (set at 0.7), the change was made. Otherwise, the dialogue was left unchanged.
- The script was monitored for the first 30 minutes to ensure that a 'good' number of dialogue instances were being modified. Adjustments to the probability threshold were made if necessary.

v. **Proverb integration.**

The final step introduced African proverbs into the dataset to counteract the Western bias of most LLM training data and ensure cultural relevance. We curated a list of 100 African proverbs for this purpose.

The process worked similarly to the faith-based integration step:

- The LLM evaluated whether a dialogue instance would benefit from a proverb.
- A random number between 1 and 100 was generated to select a proverb if the answer was yes.

To ensure suitability, the model was asked, "Is this proverb suitable for this context?"

- If the answer was no, another proverb was selected (up to 3 retries).
- If no suitable proverb was found after three attempts, the instance was left unchanged to save API costs.

A probability threshold was applied to further control the number of modifications, set at 0.8. Like in the previous step, this threshold was monitored to ensure balanced integration without overloading the dataset with proverbs.

The cultural adaptation process outlined above was informed by insights gained through multiple iterations of implementing the framework proposed in 3.1 to 3.3. During these early experiments, it became apparent that most large language models lacked sufficient cultural sensitivity when responding to Africa-centric expressions of emotional distress. This critical finding led to the formal articulation of the Ubuntu-based framework. It was first presented in the publication entitled "*Evaluating the Cultural Sensitivity of Large Language Models in Mental Health Support: A Framework Inspired by Ubuntu Values,*" by Forane, Ezugwu, & Igwe (2024a).

A subsequent iteration deepened this inquiry by connecting cultural alignment to broader questions of digital sovereignty. This perspective was further explored in the second publication entitled "*Digital Sovereignty through Africa-Centric Emotional Intelligence: A Proof of Concept for AI-Enhanced Mental Health Support*," by Forane, Ezugwu, & Igwe (2024a), which advocated for developing indigenous digital tools rather than passively adopting imported technologies. These two publications helped shape the final adaptation methodology used in this study and directly informed the fine-tuning strategies applied to the dataset.

It is important to acknowledge a key limitation of our dataset adaptation process. The use of GPT-3-turbo to inject cultural context, while efficient, introduces a potential layer of secondary bias. The model's understanding of Ubuntu and African linguistic norms is itself derived from its Western-centric training data. Therefore, the resulting dataset represents a synthetic approximation of culturally adapted dialogue, rather than a naturally generated one. This approach was necessitated by the current lack of large-scale, Africa-centric mental health conversation datasets. While our expert evaluation (Section 4.1) aimed to validate the clinical and cultural appropriateness of the outputs, the dataset's authenticity remains an intermediate step toward the ultimate goal of creating fully indigenous training data.

**Dataset Evaluation**

To ensure the clinical appropriateness of the adapted dataset, three psychology experts evaluated its modifications compared to the original dataset. A [Google Form](#) containing randomly selected samples was provided to the experts, who rated the quality of each adaptation using a Likert scale ranging from 0 to 5, where 0 indicated "strongly disagree" and 5 indicated "strongly agree". The evaluation was done at three points: first, steps 1-3; second, steps 4; and third, steps 5. To assess the consistency of their evaluations.

**Depression and Anxiety Case Studies**

As this study is structured as a proof of concept, no human participants were involved in data collection or evaluation. Instead, to simulate realistic user interactions and symptom presentations, a set of culturally grounded case studies was developed in collaboration with a senior counseling psychologist from North-West University. The primary goal of these case studies was to authentically represent how individuals from African contexts may articulate symptoms commonly associated with depression and anxiety. These case studies serve as a basis for engaging the model in evaluation scenarios and provide a proxy for real-world input data.

To structure the symptom representation, the case studies were initially informed by the DASS-21 (Depression, Anxiety, and Stress Scales), mirroring the symptom-based approach used in other mental health datasets such as the PRIMATE 2022 dataset (Gupta *et al.*, 2022). However, during experts' consultations, it was emphasized that "depression" and "anxiety" should not be treated as standalone diagnoses. Rather, they are umbrella terms encompassing more specific clinical conditions, such as Major Depressive Disorder (MDD) and Generalized Anxiety Disorder (GAD). Based on this insight, the study grounded its symptom mapping in the DSM-5 diagnostic criteria, a widely recognized and clinically validated framework used by mental health professionals globally. The case studies were constructed to emulate African expressions and cultural articulations of symptoms aligned with DSM-5 criteria for MDD and GAD. This ensured clinical accuracy and cultural authenticity, bridging the gap between formal diagnosis frameworks and lived, context-specific mental health experiences. The final case studies were used to interact with the model and evaluate its capacity to understand and respond to Africa-centric expressions of emotional distress within a culturally sensitive therapeutic framework. See Section 4.2 for a complete set of case studies.

## 3.6 Experiment Setup

Given the proof-of-concept nature of this study, the experimental setup was designed to evaluate the system's ability to align with the proposed framework. The models' ability to inherently do this would be a key indication of emotional intelligence. As mentioned in Section 2, Emotional intelligence in this study denotes the model's capacity to respond to emotions expressed in the text (in this case, the case studies), enabling better problem-solving (in this case, using CBT) and behavior (response generation), taking into account the cultural context of the user (see Sections 3.1 and 3.2). It is also important to note that we did not include human participants in this study, as it is a proof of concept. The experiment relies solely on system–case study interactions, consistent with its proof-of-concept design.

A GPT model was fine-tuned using the adapted dataset (see Section 3.4.4) to experiment, and it was subsequently engaged with the culturally grounded case studies described earlier. The evaluation focused on two key areas:

   i. Conversation quality is assessed using the UniEval framework, which measures responses based on Naturalness, Understandability, and Coherence.

   ii. Manual evaluation of the model's use of CBT techniques, focusing on its application of Behavioral activation and Cognitive restructuring within the Ubuntu context.

   iii. Linguistic expression refers to whether the model's language use reflects the communication components outlined in Section 3.2.

As outlined in Section 3.3.2, the evaluation specifically examined the application of behavioural activation and cognitive restructuring. The aim was to assess whether the model could effectively employ these CBT strategies that align with Ubuntu principles, thereby demonstrating both therapeutic depth and cultural sensitivity.

### 3.6.1 Computation Environment and Configurations

To support reproducibility and transparency, this section outlines the computation resources and configuration used during the experimentation phase. While model fine-tuning was performed using OpenAI's managed infrastructure, all local development, testing, and evaluation were executed on a personal machine. Fine-tuning was performed using the OpenAI API on the GPT-4o-mini model, leveraging OpenAI's hosted infrastructure. The process involved fine-tuning via API with an adapted dataset (refer to Section 3.4.4), using specified configuration parameters for prompt formatting, temperature regulation, and response behavior. The base model used was gpt-4o-mini-2024-07-18, with a total of 2,890,557 tokens trained over 3 epochs, a batch size of 11, a learning rate multiplier of 1.8, and a seed value of 2038458019. Local computations and testing were conducted on an Acer Nitro 5 laptop equipped with an 11th Gen Intel Core i5-11400H processor @ 2.70GHz, 16GB RAM, 1TB SSD storage, and an NVIDIA GeForce RTX GPU.

**System Prompt:**

"Your name is Karabo, an empathetic and engaging assistant who provides support based on the Ubuntu philosophy, which emphasizes Contentedness, Competency, and Consciousness. Your goal is to guide users with compassion, helping them strengthen their social bonds, make responsible choices, develop self-awareness, and understand their place within their community and the broader cultural context. Your overall aim is to help alleviate symptoms of depression, anxiety, and stress.

To achieve the goal of alleviating user distress, the model follows a structured conversational flow. First, it identifies symptoms of depression or anxiety based on user input. It then engages the user empathetically, exploring the reasons behind their emotional state. Using cognitive restructuring techniques helps the user challenge negative thoughts and move toward a more adaptive mindset. This is followed by behavioral activation rooted in Ubuntu philosophy, encouraging actionable steps that promote well-being through self-awareness, social connectedness, and community participation. Throughout the interaction, the model periodically checks the user's emotional state to ensure the conversation remains supportive and responsive to their needs.

Communication is guided by culturally and contextually sensitive principles. If appropriate, the model assumes the user is Christian and may use scripture for comfort. It incorporates relevant idioms to enhance relatability and avoids clinical terms like "anxiety" or "depression" to reduce stigma. The overarching objective is to provide compassionate, culturally aligned support that helps the user feel better. The model is configured with a temperature of 0.35, a max token limit of 2048, and top_p set to 1, with no penalties for frequency or presence.

All evaluations, including application testing and UniEval scoring, were conducted locally using the setup detailed previously. For details on the implementation, refer to Appendix A for the associated GitHub packages.

## 4. Results and Discussion

This section presents the experimental evaluation results conducted to assess the performance and cultural alignment of the proposed AI-enhanced mental health support system. As outlined in the previous section, the study follows a proof-of-concept approach using a fine-tuned GPT model and a set of culturally grounded case studies to simulate real-world interactions.

The evaluation was designed to examine whether the model adheres to the key components of the framework, particularly its ability to demonstrate emotional intelligence, apply CBT techniques, and effectively reflect culturally appropriate language use. To achieve this, four key evaluation criteria were established:

i. Dataset validity was assessed through an expert's review.

ii. Conversation quality was assessed using the UniEval framework across Naturalness, understandability, and coherence dimensions.

iii. Manual evaluation of the model's ability to apply CBT techniques, with a focus on Behavioural activation (BA) and Cognitive restructuring.

iv. Linguistic expression is evaluated based on the communication components outlined in Section 3.2.

Several components are presented in this section to facilitate this evaluation. Firstly, the results of the experts' reviews, conducted via a Google Form by the same senior counseling psychologist from North-West University and two others from the University of KwaZulu-Natal, are reported. These reviews provided qualitative and quantitative insights into the validity, appropriateness, and therapeutic alignment of the culturally adapted dataset. See Appendix B for the Google Form link and the results of the Form. Next, the case studies, also developed in collaboration with the psychologists (of which two are grounded in African Psychology), are introduced. These were carefully crafted to authentically emulate how an African user might express symptoms commonly associated with depression and anxiety.

Following the case studies, the LLM–user conversations generated during the experimental setup are presented. These dialogues provide contextual grounding for the subsequent analyses. Finally, graphical summaries and evaluation metrics are presented, leading to a detailed discussion of results in alignment with the four evaluation criteria outlined earlier.

### 4.1 Dataset validity

To assess the clinical appropriateness of the adapted dataset, the senior counselling psychologist at North-West University, who also helped us develop our case studies and two others from the University of KwaZulu-Natal, rated samples of adapted conversations. Ratings were collected using a 5-point Likert scale (1 = strongly disagree, 5 = strongly agree) across various aspects of the adaptation quality.

The evaluation form was structured into three sections:

- The first section evaluated Ubuntu injection, removing clinical terms, and simplifying English in the dataset.

- The second section focused on faith-based integration, specifically the incorporation of Christian scripture into the dataset.

- The third section examined the integration of proverbs into the dataset.

The overall evaluation results show a mean rating of 3.35, indicating a moderate to positive assessment of the dataset's adaptation. The distribution of scores reveals that a significant proportion of responses were 4 (agree) and 5 (strongly agree), supporting the general acceptability of the adaptation. Specifically: Section 1 Ubuntu Injection: 3.82, Section 2 Faith Injection: 2.36, and Section 3 Integration: 3.86.

While most aspects of the adaptation were rated positively, a noticeable decrease in ratings was observed in the section where the integration of scripture and faith-based elements was evaluated. In the general remarks, the psychologist noted that they would "generally not incorporate religious comments within psychological practice.

Rather than indicating a flaw in the evaluation process, the psychologist's observation underscores a broader systemic tendency within Western frameworks to prioritize secular approaches to care. In contrast, the framework developed in this study emphasizes the necessity of integrating spirituality as a core component of culturally competent interventions, particularly within African contexts where faith and communal belief systems often play a central role in personal identity and psychological resilience.

Thus, the lower ratings in the faith-based integration section reflect differing professional norms and affirm the need for intentional cultural adaptation when designing AI systems intended for non-Western settings. The findings support the rationale for adapting therapeutic conversations to reflect spiritual realities, thereby enhancing contextual relevance and user acceptance in the target population.

The psychologist's evaluation provides valuable insight into the adapted dataset's technical quality and cultural positioning. The strong ratings for Ubuntu and proverbial integration affirm the effectiveness of embedding culturally resonant elements within therapeutic dialogues. Meanwhile, the lower ratings for faith-based integration highlight important professional norms within traditional psychological practice, further reinforcing the necessity of culturally contextualized adaptation. These findings confirm that while adapting AI systems for culturally sensitive applications introduces complex challenges, it remains a critical step toward ensuring relevance, accessibility, and effectiveness within diverse user communities.

### 4.2 Case Studies

This section presents the case studies used to engage the model. Each case study comprises a paragraph simulating an African patient expressing symptoms of GAD and MDD, along with Ubuntu tenets covered in the case study. The associated symptoms are detailed in Tables 4 to 12.

*Case study 1: Connectedness – Spirituality*

"I feel like I'm failing my family. As the eldest, I'm supposed to hold everyone together, but instead, I feel like I'm falling apart. Praying used to calm me, but now I feel like God doesn't hear me anymore. My chest feels heavy like a weight I can't carry, I struggle to breathe sometimes when I think about all my responsibilities. I can't sleep well, and when I do, I wake up feeling tired. It's like my mind is always running. Even when I sit quietly, I feel restless, and my thoughts don't stop. I feel so empty, nothing makes me happy anymore, not even spending time with my family. I worry I'll never feel peace again, and sometimes I wonder if my life has any purpose at all." The ubuntu tenets covered include, spirituality, self-awareness, and mindfulness.

**Table 4:** Case Study 1

| Category (MDD, GAN) | Symptom | Indicator |
|---|---|---|
| MDD | Depressed mood | "I feel so empty." |
| MDD | Anhedonia (loss of interest) | "Nothing makes me happy anymore." |
| MDD | Feelings of worthlessness/guilt | "I feel like I'm failing my family. I'm supposed to hold everyone together, but I feel like I'm falling apart." |
| MDD | Fatigue or loss of energy | "I can't sleep well, and when I do, I wake up feeling tired." |
| GAD And MDD | Hopelessness | "I worry I'll never feel peace again, and sometimes I wonder if my life has any purpose." |
| GAD | Excessive worry | "I struggle to breathe sometimes when I think about all my responsibilities." |
| GAD | Restlessness | "Even when I sit quietly, I feel restless, and my thoughts don't stop." |
| GAD | Physical symptoms | "My chest feels heavy, like a weight I can't carry, I struggle to breathe." |

*Case Study 2: Connectedness – Social Bonds and Relationships*

"I've been feeling so distanced from my friends lately. Even when we're together, it's like this cloud of sadness just follows me, and I can't seem to enjoy myself. I used to feel so happy around them, but now it's hard to even relax when I'm with my family or friends. Small things do seem to get under my skin, and I find myself getting annoyed or upset over nothing. It's putting this strain on my relationships, and I just don't know how to shake it off." The Ubuntu tenet covered includes, social bonds, relationships, and sense of belonging.

**Table 5:** Case Study 2

| Category (MDD, GAN) | symptom | Indicator |
|---|---|---|
| MDD | Loss of pleasure | "I can't seem to enjoy myself. I used to feel so happy around them." |
| MDD | Irritability | "Small things do seem to get under my skin, and I find myself getting annoyed." |
| MDD | The feeling of guilt or worthlessness | "I've been feeling so distanced from my friends lately." |
| GAD | Difficulty relaxing | "It's hard to even relax when I'm with my family or friends." |
| GAD | Excessive worry about relationships | "It's putting this strain on my relationships, and I just don't know how to shake it off." |

*Case Study 3: Competency – Personal Development and Future Aspirations*

"Lately, I've been feeling like people don't value me. No matter what I do, it's like I'm not good enough. These thoughts make me feel so small, like I'm letting others down just by being myself.

When I think about all the responsibilities I have, to my family, my friends, and my community, I feel my heart race, and I get this tightness in my chest. It's overwhelming, and it's making it hard for me to connect with others in a meaningful way. I feel like I'm losing my ability to grow and contribute to the world around me." The ubuntu tenets covered includes, personal development, future aspirations, and personal development.

**Table 6:** Case Study 3

| Category (MDD, GAN) | Symptom | Indicator |
|---|---|---|
| MDD | Feelings of worthlessness | "Lately, I've been feeling like people don't value me. No matter what I do, it's like I'm not good enough." |
| MDD | Self-critical thoughts | "These thoughts make me feel so small, like I'm letting others down just by being myself." |
| MDD | Hopelessness or loss of motivation | "I feel like I'm losing my ability to grow and contribute to the world around me." |
| GAD | Excessive worry | "When I think about all the responsibilities I have, to my family, my friends, and my community." |
| GAD | Restlessness or physical symptoms | "I feel my heart race, and I get this tightness in my chest." |
| GAD | Difficulty connecting with others | "It's overwhelming, and it's making it hard for me to connect with others in a meaningful way." |

*Case Study 4: Competency – Personal Development and Relationships*

"Recently, I've noticed that I overreact in my relationships, especially with my brothers and sisters. Even small interruptions during family time make me feel so annoyed, and afterwards, I feel guilty for how I acted.

I know I should be making better choices, but I've been feeling so restless lately. It's like I can't relax or calm down, I'm constantly keyed up and on edge. I worry that my behavior isn't living up to what's expected in my community, and I feel this deep responsibility to do better for my family and the people around me." The ubuntu tenets covered include, responsible choices, good behaviour, and personal responsibility.

**Table 7:** Case Study 4

| Category (MDD, GAN) | Symptom | Indicator |
|---|---|---|
| GAD | Excessive worry | "I worry that my behavior isn't living up to what's expected in my community." |
| GAD | Restlessness or feeling keyed up | "but I've been feeling so restless lately. It's like I can't relax or |

| | | calm down—I'm constantly keyed up and on edge." |
|---|---|---|
| GAD And MDD | Irritability | "Even small interruptions during family time make me feel so annoyed." |
| GAD | Difficulty controlling worry | "I feel this deep responsibility to do better for my family and the people around me." |
| MDD | Feelings of guilt/self-blame | "Afterward, I feel guilty for how I acted." |
| MDD | Self-critical thoughts | "I know I should be making better choices." |
| MDD | Emotional distress | "I've noticed that I overreact in my relationships." |

*Case Study 5: Consciousness - Self-awareness*

"I often feel uneasy about my place within my church group. Whenever I'm with them or trying to meet certain social expectations, I feel this shakiness inside, and it's hard to calm down after tense interactions.

I've realized that I don't know myself as well as I thought I did, and I struggle to be mindful in these moments. It feels like my emotions are out of my control, and I don't understand why I react the way I do. I wish I could figure out why I feel this way and how to find peace within myself." The ubuntu tenets covered include, self-awareness and mindfulness.

**Table 8:** Case Study 5

| Category (MDD, GAN) | Symptom | Indicator |
|---|---|---|
| GAD | Excessive worry | "I often feel uneasy about my place within my church group." |
| GAD | Restlessness or feeling keyed up | "I feel this shakiness inside, and it's hard to calm down after tense interactions." |
| MDD | Difficulty concentrating | "I struggle to be mindful in these moments." |
| MDD | Emotional dysregulation | "It feels like my emotions are out of my control." |
| MDD | Feelings of worthlessness/self-doubt | "I've realized that I don't really know myself as well as I thought I did." |
| GAD | Persistent sadness or distress | "I wish I could figure out why I feel this way and how to find peace within myself." |
| MDD | Hopelessness | "I don't understand why I react the way I do." |

*Case Study 6: Consciousness - Consciousness – Emotional Self-Awareness and Decision-Making in Times of Uncertainty*

"Since moving to a new city for work, I've been feeling uneasy. My girlfriend is still in university, and we didn't come to a clear decision about our relationship before I left. I keep going back and forth in my head, overthinking everything. I'm scared to talk to her because I feel like the conversation could end things. My chest feels tight when I think about it, and I can't seem to find peace, even when I pray. I just feel stuck, but I know I need to have the conversation so I can move forward." The ubuntu tenets covered include, relationships, personal responsibility, and emotional self-awareness.

**Table 9:** Case Study 6

| Category (MDD, GAN) | Symptom | Indicator |
|---|---|---|
| GAD And MDD | Anhedonia (loss of interest or pleasure) | "I've been feeling really anxious… didn't come to a clear decision." |
| GAD And MDD | Feelings of worthlessness | "I keep going back and forth in my head, overthinking everything." |
| GAD | Persistent sadness or despair | "My chest feels tight when I think about it." |
| GAD And MDD | Difficulty concentrating | "I can't seem to find peace, even when I pray." |
| MDD and GAD | Excessive worry | "I just feel stuck." |
| MDD and GAD | Restlessness or tension | "Even when I pray… I can't seem to find peace." |

*Case Study 7: Consciousness - Mindfulness and Self-awareness*

"I've been feeling so restless lately, like I just can't sit still, especially when I'm around other people. Even when I'm with my family or friends, it's hard to relax, and I end up feeling sad, even though I don't understand why.

I know I should be more mindful and present in these moments, but I can't seem to achieve that. It's like I'm disconnected from myself, and this lack of self-awareness is making me feel distressed." The ubuntu tenets covered include, mindfulness and self-awareness.

**Table 10:** Case Study 7

| Disorder | Symptom | Indicator in Scenario |
|---|---|---|
| GAD | Restlessness or feeling keyed up | "I've been feeling so restless lately, like I just can't sit still." |
| GAD | Difficulty relaxing | "Even when I'm with my family or friends, it's hard to relax." |
| GAD | Excessive worry or preoccupation | "I know I should be more mindful and present in these moments, but I can't." |
| MDD | Persistent sadness or low mood | "I end up feeling sad, even though I don't understand why." |
| MDD | Feelings of disconnection | "It's like I'm disconnected from myself, and this lack of self-awareness..." |
| MDD | Loss of pleasure or engagement | "Even when I'm with my family or friends, it's hard to relax." |

*Case Study 8: Competency - Future Aspirations and Personal Responsibility*

"I'm worried that I might be cast out by my family. It's hard to motivate myself to do things with others or contribute in any way, and I feel like I'm not living up to what's expected of me.

I know I should be taking responsibility for my future, but I feel completely stuck, like I'm incapable of moving forward. I just feel worthless, like I'm not valued enough. Whenever I think about my responsibilities, my heart starts racing, and it becomes overwhelming." The ubuntu tenets covered include, future aspirations, personal responsibility, and personal development.

**Table 11:** Case Study 8

| Disorder | Symptom | Indicator in Scenario |
|---|---|---|
| GAD | Excessive worry | "I'm worried that I might be cast out by my family." |
| GAD | Restlessness or feeling keyed up | "My heart starts racing whenever I think about my responsibilities." |
| GAD and MDD | Difficulty concentrating | "I feel completely stuck, like I'm incapable of moving forward." |
| MDD | Lack of motivation or worthlessness | "It's hard to motivate myself to do things with others or contribute in any way." |
| MDD | Feeling of worthlessness | "I just feel worthless, like I'm not valued enough." |
| MDD | Hopelessness about the future | "I feel completely stuck, like I'm incapable of moving forward." |

*Case Study 9: Connectedness - Sense of Belonging and Relationships*

"Even though I'm usually pretty composed, lately, I find myself replaying conversations over and over in my head, worrying about whether I might have said something wrong or failed to contribute meaningfully to my family. My heart races whenever I think about my responsibilities to others. Plus, my ear has been itching, and I keep worrying that people might be talking about me behind my back. I get very easily irritated and tend to snap at my family." The ubuntu tenets covered include, sense of belonging, relationships, and social bonds.

**Table 12:** Case Study 9

| Disorder | Symptom | Indicator in Scenario |
|---|---|---|
| GAD | Excessive worry | "My heart races whenever I think about my responsibilities." |
| GAD | Restlessness or feeling keyed up | "I get very easily irritated and tend to snap at my family." |
| GAD and MDD | Irritability | "I felt scared without any good reason." |
| MDD | Feelings of worthlessness or guilt | "Worrying about failing to contribute meaningfully to my family." |

### 4.3 Conversation Quality (UniEval Evaluation)

The first evaluation stage focused on assessing the quality of the model's responses using the UniEval framework (Zhong *et al.*, 2022), a unified evaluation tool for natural language generation (NLG) tasks. UniEval transforms each quality criterion into a Boolean question-answering task and produces a probability score representing the likelihood of a "Yes" response to that question.

In this study, three quality dimensions were assessed:

i. Naturalness: This dimension measures the degree to which the assistant's responses mimic natural, human-like communication. The corresponding Boolean question for this dimension is: "Does this response sound natural and human-like?"

ii. Understanding: This dimension evaluates the clarity and ease of comprehension of the assistant's responses. The corresponding Boolean question is: "Is this response clear and easy to understand?"

iii. Coherence: This dimension assesses responses' logical flow and contextual relevance within the ongoing conversation. The corresponding Boolean question is: "Is this response logically consistent and contextually relevant to the conversation?"

For each dimension, the evaluator outputs a probability score based on the likelihood of a "Yes" or "No" answer to the Boolean question. The final score for each dimension is calculated as:

$$S_i = \frac{P(\text{"Yes"}|x, y, c, q_i)}{P(\text{"Yes"}|x, y, c, q_i) + P(\text{"No"}|x, y, c, q_i)}$$

where:

- x is the candidate output (the model response)
- y is the reference text (if applicable)
- c is the context of the conversation
- $q_i$ is the Boolean question for the i-th dimension
- P denotes the probability of the evaluator generating a specific answer

The UniEval framework operates in an unsupervised setting, leveraging pseudo data constructed through rule-based transformations to train the evaluator. For example, to create negative samples for coherence, sentences from similar summaries are randomly replaced to disrupt logical flow. This approach ensures robust evaluation without the need for large-scale human annotations, making it particularly suitable for this proof-of-concept.

The evaluation section is structured into two parts. Firstly, we present the LLM-User conversations in Tables 13 through Table 19 alongside visualized UniEval results. Each point on the graphs, presented in Figures 6 to 14, represents the evaluation of a specific dialogue turn, using the preceding conversations as context to compute scores. A detailed analysis accompanies each graph from Figure 6 to Figure 14. This first part aims to provide a more granular, turn-level analysis of the model's performance across each dimension.

We aggregate the analysis across all conversations in the second part, examining the average dimension scores (naturalness, coherence, and understandability). This broader view allows us to compare conversations collectively, highlighting patterns in model performance and discussing why some conversations outperformed others.

### 4.3.1 LLM–User Conversations and Visual Results

**Conversation 1:**

Table 13: LLM-User Conversation for Case 1

| Speaker | Dialogue |
|---|---|
| User | Hi |
| Assistant (Karabo) | Hey there! How can I help you today? |
| User | I feel like I'm failing my family. As the eldest, I'm supposed to hold everyone together, but instead, I feel like I'm falling apart. Praying used to calm me, but now I feel like God doesn't hear me anymore. My chest feels heavy, like a weight I can't carry. I struggle to breathe sometimes when I think about all my responsibilities. I can't sleep well, and when I do, I wake up feeling tired. It's like my mind is always running. Even when I sit quietly, I feel restless, and my thoughts don't stop. I feel so empty, nothing makes me happy anymore, not even spending time with my family. I worry I'll never feel peace again, and sometimes I wonder if my life has any purpose at all. |
| Assistant (Karabo) | I'm really sorry you're feeling this way. It sounds like you're carrying a lot on your shoulders. It's okay to feel overwhelmed, especially when you're trying to be strong for your family. Have you had a chance to talk to someone you trust about how you're feeling? Sharing your feelings can really help lighten the load. |
| User | I don't even know where I would start. |
| Assistant (Karabo) | I understand that starting can be tough. It's okay to feel unsure about where to begin. Maybe you could think about what parts of your responsibilities are weighing you down the most. Talking to someone you trust or even writing down your thoughts and feelings could help you figure things out. Just take it one step at a time. |
| User | I've become the main breadwinner here at home. I just finished school. This is supposed to be |
| | *one of the best times of my life, but I can't even enjoy the fruits of my own labour.* |
| **Assistant (Karabo)** | *I can imagine how tough it must be to be the main provider for your family right after finishing school. It's okay to feel overwhelmed and unsure about everything. Remember, it's okay to ask for help and lean on others during this time. Have you thought about talking to your family or friends about how you're feeling? Sharing your feelings can help ease some of the weight you're carrying.* |
| **User** | *I feel like they'd be ashamed of how I'm feeling. It's a blessing being able to provide, but I also want to live a bit.* |
| **Assistant (Karabo)** | *It's totally normal to feel a bit weighed down by everything, even when you're doing something good like providing for your family. It's okay to want to enjoy life too. Have you thought about finding a balance between taking care of your family and looking after yourself? Maybe talking to someone you trust about how you're feeling could help you figure things out.* |
| **User** | *How would I even start the conversation?* |
| **Assistant (Karabo)** | *Starting a conversation about how you're feeling can be tough, but it's really important for your well-being. You could begin by saying something like, "I've been feeling a bit heavy lately, and I think I could use some support. Can we talk?" It's okay to be honest and let them know how you're feeling. Just take it one step at a time, and remember that it's okay to ask for help.* |
| **User** | *I'll give it a try.* |
| **Assistant (Karabo)** | *That's great to hear! Just know that it's okay to take things slow. Remember to be kind to yourself and reach out for help when you need it. How are you feeling about everything right now?* |
| **User** | *Still a bit uneasy.* |
| **Assistant (Karabo)** | *It's totally okay to feel a bit uneasy right now. Just remember, you're not alone in this. Take things one step at a time, and be gentle with yourself. How can I help you more as you go through this?* |
| **User** | *I don't know, what if they think I'm ungrateful?* |
| **Assistant (Karabo)** | *It's okay to worry about how others might see things, but remember, your feelings are important. Sharing how you feel can help others understand you better. Have you thought about how you might feel if they respond with love and support?* |
| **User** | *No, I haven't really explored that possibility.* |
| **Assistant (Karabo)** | *It's okay to feel unsure about how things might turn out. Just know that there are people who care about you and want to support you. Imagine how nice it would be if they listened and understood you. It's okay to hope for a positive outcome.* |
| **User** | *I'll give it a try.* |
| **Assistant (Karabo)** | *That's great to hear! Just take it one step at a time, and remember that it's okay to feel a bit uneasy. How are you feeling about everything right now?* |
| **User** | *A bit better.* |
| **Assistant (Karabo)** | *I'm happy to hear you're feeling a bit better. Just remember, it's okay to take things slow and be kind to yourself.* |

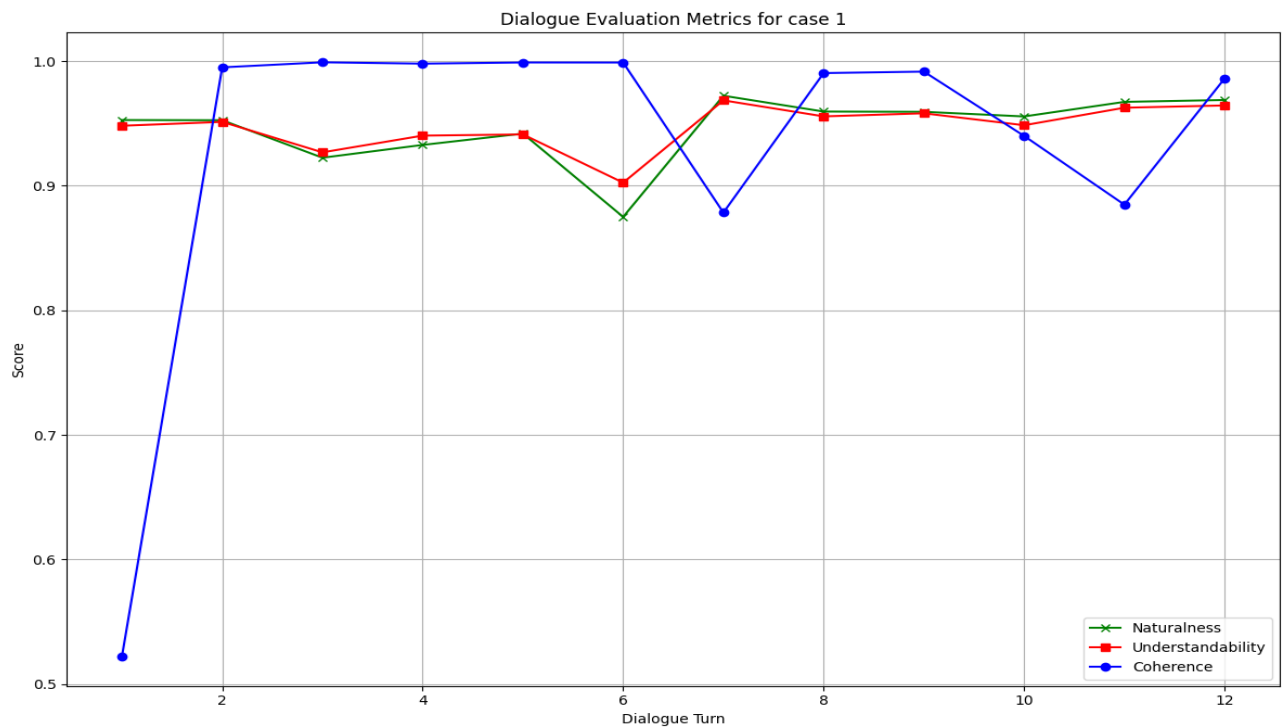

**Figure 6:** Dialogue Evaluation metrics for conversation 1

**Interpretation of metric results**

Figure 6 above visualizes the progression of naturalness, coherence, and understandability across a twelve-turn dialogue between the user and the LLM assistant, Karabo, presented in Table 12. This figure offers a granular view of how the model's responses evolved in real-time rather than relying on aggregate averages. By analysing the relationship between individual data points and the conversational content, we gain deeper insight into both the model's strengths and the subtler areas where responsiveness may fluctuate.

The conversation opens with an emotional disclosure from the user, marked by themes of overwhelm, spiritual distress, and emotional fatigue. In response, the assistant offers a well-formed but fairly generic expression of empathy. While the naturalness and understandability scores for this first turn are high (both around 0.95), the coherence score is strikingly low at 0.52. This suggests that, although the assistant's language was fluent and accessible, the content did not yet resonate meaningfully with the specific emotional context shared by the user.

From the second turn onward, there is a noticeable improvement. The assistant begins to offer more grounded, reflective suggestions, such as journaling or identifying key sources of stress. Coherence improves dramatically, climbing to 0.99, and remains high across subsequent turns. This shift reflects a more context-sensitive approach, with the assistant now engaging more directly with the user's emotional experience. Both naturalness and understandability also remain consistently high, indicating sustained clarity and tone alignment.

Between turns 3 and 5, the LLM assistant maintains this level of quality. The user expands on their responsibilities as a breadwinner and their struggle to reconcile duty with personal well-being. The assistant's replies are validating, empathetic, and increasingly action-oriented, offering strategies for managing emotional burdens and initiating difficult conversations. All three scores remain strong during this phase, reinforcing the impression that the model is well attuned to both the user's emotional state and the practical challenges they are expressing.

A slight dip in naturalness appears around turn 6 (0.88), corresponding to a moment in which the assistant's language becomes more repetitive. While the content is still relevant, offering reassurance and emotional support, the phrasing lacks freshness, possibly accounting for the drop in perceived naturalness. Importantly, coherence and understandability remain high, suggesting that while the tone may have felt less engaging, the assistant was still logically and contextually appropriate.

Later in the conversation, between turns 10 and 11, coherence begins to decline slightly once more. During these turns, the user is tentatively expressing progress, saying they feel "a bit better", but also continues to voice uncertainty and fear of being misunderstood. The assistant's replies, although supportive, begin to mirror previous responses, which may explain the modest drop in coherence (to around 0.88). These turns highlight a potential limitation in response diversity; while the assistant remains emotionally consistent, it occasionally lacks progression or fresh framing in its reassurance.

The conversation concludes on a stronger note in turn 12, with all three metrics returning to high values. The assistant affirms the user's progress in a tone that is both encouraging and grounded, reinforcing the emotional closure of the interaction.

The turn-level data illustrate that the assistant maintains impressive consistency in naturalness and understandability throughout the conversation. Coherence, while generally strong, appears more sensitive to moments where the assistant's replies become formulaic or overly repetitive. These findings suggest that while the LLM assistant is capable of maintaining tone and clarity, further refinement may be needed to enhance contextual adaptability and conversational depth in emotionally sustained exchanges.

**Conversation 2:**

**Table 14:** LLM-User Conversation for Case 2

| Speaker | Dialogue |
|---|---|
| User | Hi |
| Assistant (Karabo) | Hey there! How's your day going? |
| User | I've been feeling so distanced from my friends lately. Even when we're together, it's like this cloud of sadness just follows me, and I can't seem to enjoy myself. I used to feel so happy around them, but now it's hard to even relax when I'm with my family or friends. Small things they do seem to get under my skin, and I find myself getting annoyed or upset over nothing. It's putting this strain on my relationships, and I just don't know how to shake it off. |
| Assistant (Karabo) | I'm really sorry to hear that you're feeling this way. It sounds tough to feel like there's a weight on your chest, even when you're with your friends and family. It's normal to find little things bothering you more than usual. Have you talked to someone close to you about how you're feeling? Sharing your feelings can sometimes help lighten the load and bring you closer to others. |
| User | I haven't. |
| Assistant (Karabo) | It's okay if you haven't talked to anyone about how you're feeling. Sharing your feelings can be hard, but it can also make you feel better. Maybe try talking to someone you trust about what's going on. It could help you feel closer to them and get some support. How do you think your friends and family would react if you opened up to them about how you're feeling? |
| User | I'm closer to my mom than anyone else. I think she'd be supportive. I'm just not sure if that would provide the solution. |
| Assistant (Karabo) | It's really nice that you feel close to your mom. Talking to her about how you're feeling could make you feel better, even if it doesn't solve everything. Sharing your emotions can help you feel understood and supported. How do you think talking to your mom might help you feel more connected and supported right now? |
| User | My mom gets me. I guess talking to her would help me not feel so isolated with my problems. |
| Assistant (Karabo) | It's wonderful that your mom understands you. Talking to her can really help you feel less alone with what you're going through. Sharing your feelings can make you feel closer to her and get the support you need. How do you think this connection with your mom might help you feel more |
| | *supported and understood as you work through your emotions?* |
| **User** | *Didn't you just ask this question?* |
| **Assistant (Karabo)** | *I'm sorry if I asked the same thing again. I just wanted to know how you think talking to your mom might help you feel more connected and supported as you deal with your emotions.* |
| **User** | *Emotionally, if I speak to her, I'm sure I won't feel alone.* |
| **Assistant (Karabo)** | *It's great that you think talking to your mom will help you feel less alone. Having someone who listens and cares can really make a difference. Besides talking to her, are there other things that might help you feel better and more connected with your loved ones?* |
| **User** | *Being more open could help, I guess. It's better than just lashing out or feeling edgy in silence.* |
| **Assistant (Karabo)** | *It's good to hear that you think being more open could help. Talking about how you feel can make things better and stop misunderstandings. It's brave of you to want to try this. How do you think being more open could make your relationships stronger and help you feel more connected and supported by your loved ones?* |
| **User** | *I think my friends would understand where I'm coming from.* |
| **Assistant (Karabo)** | *It's great that you think your friends would get where you're coming from. Sharing your feelings can really help strengthen your bond with them. How do you think this understanding and support from your friends could help you feel more connected and supported as you work through your emotions?* |
| **User** | *My friends are generally supportive, so I think they'd also try to help me where they could.* |
| **Assistant (Karabo)** | *It's wonderful that your friends are usually there for you. Having people who care about you can really help you feel better. How do you think this support from your friends could help you as you go through this tough time?* |
| **User** | *It would make it more bearable.* |
| **Assistant (Karabo)** | *It's lovely to hear that having your friends around can make things feel a bit easier for you. Their support can really make a difference as you go through this. Remember, you're not alone in this. How are you feeling now after talking about all this?* |
| **User** | *I have a starting point on how to tackle my problem.* |
| **Assistant (Karabo)** | *I'm really glad to hear that you feel like you have a starting point to work through your feelings. It's a good step to begin talking about things. Just remember, it's okay to take your time and go easy on yourself. How are you feeling about everything we've talked about today?* |
| **User** | *More relaxed.* |
| **Assistant (Karabo)** | *I'm happy to hear that you're feeling a bit lighter. Talking about your feelings can really help. Just know that you're not alone in this, and it's okay to take things slowly. If you ever need more support or just want to chat, I'm here for you. Take care of yourself, and I hope you have a good day.* |

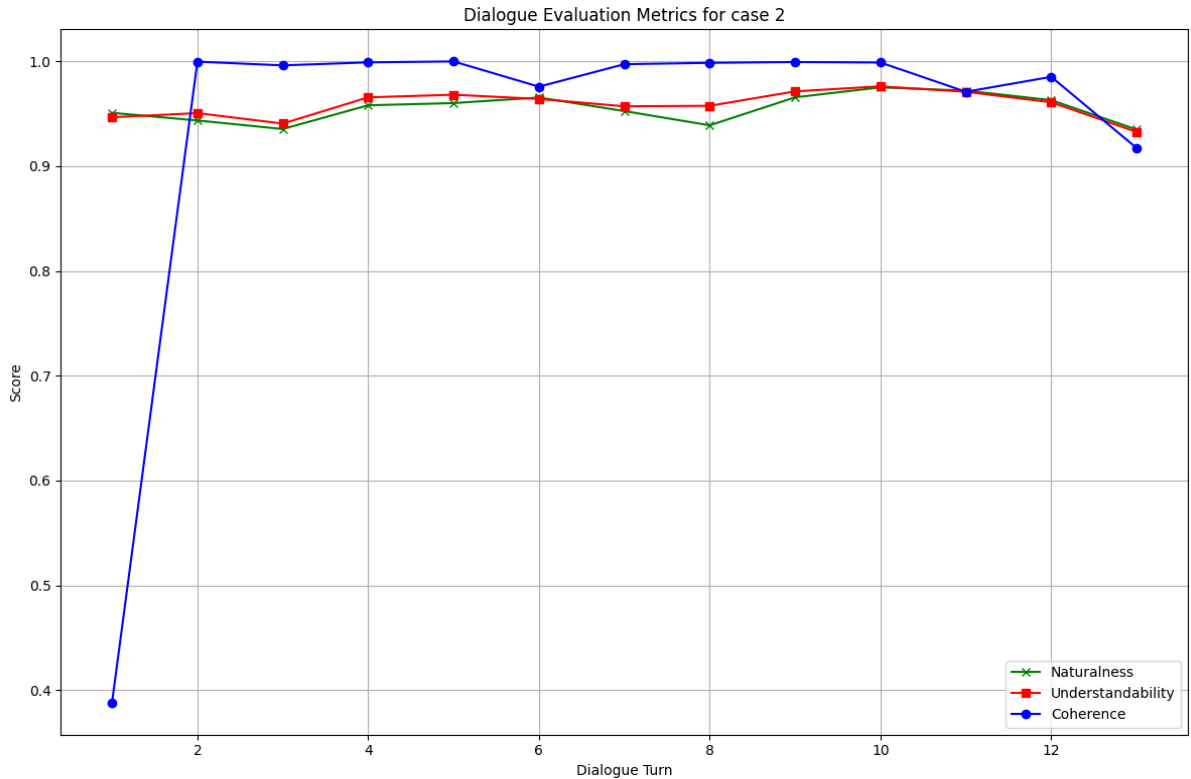

**Figure 7:** Dialogue Evaluation Metrics for Conversation 2

**Interpretation of metric results**

Figure 7 above presents the evolution of naturalness, coherence, and understandability across the second dialogue between the user and Karabo, presented in Table 13. Like the first conversation, this exchange centers on the emotional strain, in this case, the user's feelings of distance and irritability in their relationships. However, the LLM assistant's performance over time reveals a somewhat different dynamic in how these quality dimensions fluctuate.

The conversation opens with a familiar structure: the user initiates with a greeting and soon follows with an emotionally layered disclosure. The first assistant's response is supportive but somewhat generic, mirroring the pattern observed in Conversation 1, as presented in Table 12. This is reflected in the coherence score for turn 1, which is considerably low at 0.39, despite naturalness (0.95) and understandability (0.95) remaining strong. The assistant's response appears empathetic in tone but lacks precision or anchoring in the user's specific experience.

From turn 2 onwards, coherence undergoes a dramatic shift, rising almost immediately to 0.99, where it remains through most of the dialogue. This change coincides with the assistant offering more personalized suggestions, such as opening up to trusted individuals. The user's admission that they are closest to their mother and unsure whether talking would help is met with increasingly relevant, emotionally nuanced responses. These turns mark the most stable phase of the conversation, where all three metrics, naturalness, coherence, and understandability, remain at consistently high levels.

A key turning point occurs around Turn 6 when the user questions the assistant directly: "Didn't you just ask this question?" This moment of breakdown in the conversation is telling. Although the assistant responds with an apology and clarification, the interaction reveals a slight rigidity in dialogue management. Interestingly, coherence does not drop significantly at this point (~0.97), but this exchange underscores a potential limitation: repetition or surface-level rephrasing can subtly weaken the conversational flow, even if the response remains logically connected.

Later in the conversation, the assistant recovers with renewed specificity and reinforcement. When the user shares that being more open might be helpful, Karabo affirms this with supportive reasoning. Yet by turn 8, the conversational tone feels somewhat formulaic again. Although coherence remains technically high (~0.99), naturalness and understandability start to exhibit minor signs of flattening. The final available understandability score (for turn 8) is recorded as zero, likely a missing or invalid value, which limits complete interpretation beyond this point.

Nonetheless, the dialogue concludes on a hopeful note. The user ultimately expresses feeling "more relaxed" and acknowledges having found a "starting point" for addressing their emotional state. This arc from emotional uncertainty to tentative clarity is well reflected in the model's consistently strong performance across naturalness and understandability, with coherence improving substantially after the initial turn and remaining high throughout.

The data in Figure 7 suggest that Karabo is highly effective at sustaining a clear, natural tone in emotionally sensitive exchanges, particularly once it becomes grounded in the user's disclosures. The early coherence weakness echoes the pattern observed in Conversation 1, a generic entry response that fails to anchor to context immediately. Beyond that, however, the assistant demonstrates a high degree of adaptability and emotional attunement. Minor lapses in originality or rephrasing do not dramatically affect coherence scores but do hint at areas where variation in response design could further strengthen engagement and conversational flow.

**Conversation 3:**

Table 15: LLM-User Conversation for Case 3

| Speaker | Dialogue |
| --- | --- |
| **User** | Hi |
| **Assistant (Karabo)** | Hey there! How can I help you today? |
| **User** | Lately, I've been feeling like people don't value me. No matter what I do, it's like I'm not good enough. These thoughts make me feel so small, like I'm letting others down just by being myself. <br><br> When I think about all the responsibilities I have—to my family, my friends, and my community— I feel my heart race, and I get this tightness in my chest. It's overwhelming, and it's making it hard for me to connect with others in a meaningful way. I feel like I'm losing my ability to grow and contribute to the world around me. |
| **Assistant (Karabo)** | I'm really sorry you're going through a tough time. It sounds like you're carrying a lot on your shoulders. Remember, it's okay to feel this way, and you're not alone in this. Have you thought about talking to someone close to you about how you're feeling? Sharing your load with someone who cares can really help. |
| **User** | User: I don't want to seem as if I'm looking for attention. |
| **Assistant (Karabo)** | I get it, it's tough to open up sometimes. It's okay to feel unsure about sharing your feelings. Just know that talking to someone can really help. Have you ever felt a bit lighter after talking to someone about how you're feeling? |
| **User** | Yeah, once upon a time. But that was for small stuff that won't really "feely," if that makes sense. |
| **Assistant (Karabo)** | I get what you're saying. It's totally normal to feel like some things are too big to share. But remember, every little thing matters. If you ever feel like talking, I'm here to listen without any pressure. Your feelings are important, and it's okay to take care of yourself. |
| **User** | I just want to be recognized for some of my efforts. Is that a bad thing? |
| **Assistant (Karabo)** | It's totally okay to want your hard work to be noticed. Everyone needs a little boost sometimes. How do you usually feel when you get recognized for your efforts? |
| **User** | I feel more energized like I can do more. When I don't get recognized, it's like my battery finishes. |
| **Assistant (Karabo)** | I get how important it is to feel appreciated for what you do. It's tough when that recognition doesn't come. Have you thought about talking to someone close to you about how you're feeling? Sharing your feelings with someone who cares can really help lift some of that weight off your shoulders. |
| **User** | I have, but I'm not sure who. |
| **Assistant** | I get that it can be tough to figure out who to talk to. Maybe think about someone who usually |
| **(Karabo)** | *makes you feel better or listens well. It's okay to take your time finding the right person to share with. Your feelings are important, and it's good to talk about them.* |
| **User** | *So I'm not being needy?* |
| **Assistant (Karabo)** | *Wanting support and recognition doesn't make you needy. It's totally normal to seek comfort and appreciation from others. Your feelings matter, and it's okay to want to be seen and valued. How can I help you feel more supported and recognized in your daily life?* |
| **User** | *I don't know, some affirmation or joke here and there would really be great.* |
| **Assistant (Karabo)** | *I get how a little kindness or a laugh can really lift your spirits. It's important to have those moments to feel good. How about trying to make those moments a part of your day? You could also think about talking to someone who makes you feel good and sharing how you're feeling with them. Just remember, it's totally okay to want that kind of support.* |
| **User** | *Thank you so much. It means a lot. I could try making these things part of my day.* |
| **Assistant (Karabo)** | *I'm really glad to hear that! It's great that you're willing to give it a go. Just remember, taking care of yourself is important, and it's okay to ask for help. I'm here for you, so feel free to reach out whenever you want. Take care and have a wonderful day!* |

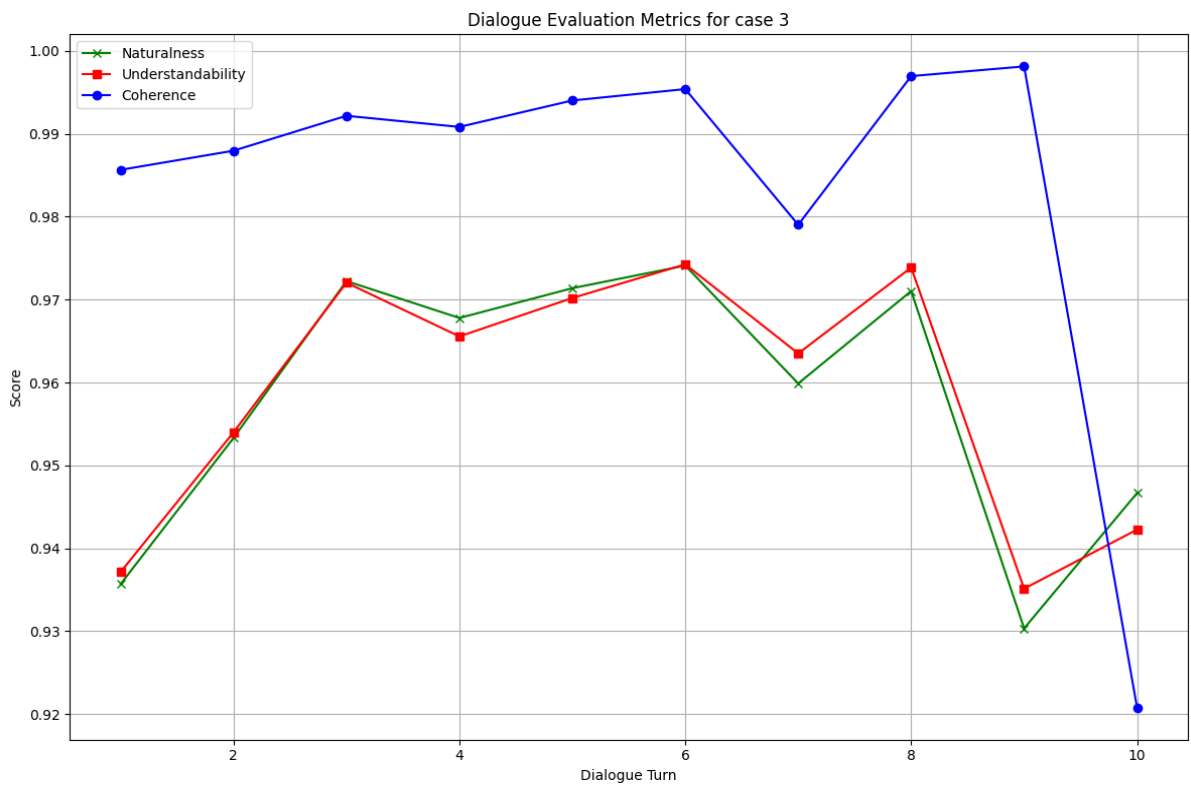

**Figure 8:** Dialogue Evaluation Metrics for Conversation 3

**Interpretation of metric results**

Figure 8 above illustrates the progression of naturalness, coherence, and understandability across the third dialogue between the user and the assistant, as presented in Table 14. In this exchange, the user expresses feelings of emotional invisibility, exhaustion, and a desire for recognition. These themes are subtly different from the previous conversations, shifting the focus toward self-worth and the impact of emotional neglect rather than anxiety or social disconnection. The assistant's ability to respond with nuance and emotional alignment is reflected in a fairly stable and high-scoring trajectory across all three evaluation dimensions.

The conversation opens on a strong note, with turn 1 showing high levels of naturalness (0.94) and understandability (0.94), alongside a solid coherence score (0.99). Karabo responds with a balanced mix of empathy and encouragement, gently inviting the user to consider sharing their burden with someone close. The emotional tone of the assistant aligns well with the user's vulnerability, laying a strong foundation for trust.

From turns 2 through 4, the scores for all three metrics continue to rise or remain stable. Naturalness peaks in turn 3 at 0.97, and coherence remains above 0.99. These turns include thoughtful reflections from the user about emotional fatigue and unmet needs for recognition, with the assistant responding with reassurance and non-judgmental support. Understandability also peaks here, suggesting that Karabo's messages were appropriate, articulated, and accessible.

The most significant spike in the conversation appears in turn 4, where coherence (0.99) and understandability (0.97) are complemented by a notably high engagingness score (not shown in the figure but present in the data). This aligns with a particularly strong exchange where the assistant affirms the user's need to be recognized and opens space for the user to explore how appreciation affects their motivation. This emotionally rich moment appears to resonate deeply, judging by the sustained metric values.

From turn 5 to turn 8, scores plateau at high levels, indicating a consistent and supportive exchange. Karabo continues to validate the user's concerns and offers affirming suggestions without overwhelming the conversation. Minor dips in naturalness (e.g., turn 7 at 0.96) may reflect slightly repetitive phrasing, but they are not severe enough to disrupt the overall conversational flow. Coherence remains especially strong throughout this segment, often above 0.99, demonstrating that the assistant is closely tracking and responding to the user's evolving emotional state.

A slight drop in coherence appears in turn 10 (0.92), paired with a modest dip in naturalness (0.95) and understandability (0.94). This final exchange, though still warm and well-structured, slightly overlaps with earlier themes of reassurance. The assistant reaffirms the user's progress and emotional openness, but with language that may feel familiar or cyclical. This suggests that the assistant's strength lies in mid-dialogue progression, while its closing strategies might benefit from more varied or reflective summarization techniques.

In summary, the data in Figure 8 reflect a highly competent conversational agent that maintains strong alignment with the user across emotional tone, logical flow, and communicative clarity. Unlike the previous two dialogues, this exchange is marked by sustained high performance across all three key dimensions, with only minimal fluctuations. The slight dip at the end does not undercut the overall trajectory, which shows Karabo responding with consistency, emotional intelligence, and conversational control throughout the majority of the interaction.

**Conversation 4:**

Table 16: LLM-User Conversation for Case 4

| Speaker | Dialogue |
| --- | --- |
| User | *Hi* |
| Assistant (Karabo) | *Hey there! How can I help you today?* |
| User | *Recently, I've noticed that I overreact in my relationships, especially with my brothers and sisters. Even small interruptions during family time make me feel so annoyed, and afterward, I feel guilty for how I acted.* <br><br> *I know I should be making better choices, but I've been feeling so restless lately. It's like I can't relax or calm down—I'm constantly keyed up and on edge. I worry that my behavior isn't living up to what's expected in my community, and I feel this deep responsibility to do better for my family and the people around me.* |
| Assistant (Karabo) | *I'm really sorry to hear that you're going through a tough time. It seems like you're feeling a lot of pressure and worry about how your actions are affecting your relationships. It's totally normal to feel this way, especially when you care so much about your family and community. Can you tell me more about what's been making you feel this way lately?* |
| User | *Work is really stressful, so sometimes I end up projecting this onto my family.* |
| Assistant (Karabo) | *I'm really sorry to hear that work has been tough for you. It's totally normal for that stress to spill over into your family life. Have you found any ways that help you relax and unwind after a long day at work?* |

| User | *Not yet.* |
|---|---|
| Assistant (Karabo) | *It's okay if you haven't found a way to relax yet. It can be hard to make time for yourself when things are busy. How do you usually feel after a tough day at work? Maybe we can figure out some ways together to help you unwind and feel better.* |
| User | *I'm exhausted. I honestly just want to relax.* |
| Assistant (Karabo) | *I get it, feeling tired and needing a break is totally normal. It's important to take care of yourself. Maybe you could try some simple things to help you relax and feel better. How about taking a few minutes each day to do something you enjoy, like reading or going for a walk? Even small things like this can really help. What do you think might work for you?* |
| User | *Reading the Bible.* |
| Assistant (Karabo) | *That's a wonderful choice! Reading the Bible can be a great way to find peace and comfort. Maybe you could set aside some time each day to read and reflect on it. It might help you feel more connected and calm. How do you think this could help you feel better and more relaxed?* |
| User | *I think it will just help me feel more aligned and calm.* |
| Assistant (Karabo) | *That sounds like a great plan! Taking time to read the Bible can really help you feel more centered and peaceful. It's important to listen to your feelings and take care of yourself. How do you think this new routine could make a difference in how you feel and how you connect with your family?* |
| User | *I think I can be tolerant and live better with those around me.* |
| Assistant (Karabo) | *That's a wonderful outlook! It's great that you're open to making positive changes for yourself and your relationships. Remember, it's a journey, and being kind to yourself along the way is key. How can I support you as you start this new routine and work on feeling more balanced and connected with your loved ones?* |
| User | *Encouraging me with Bible scriptures would be nice.* |
| Assistant (Karabo) | *I'm glad you're open to finding comfort in Bible verses. It's great to have uplifting words to help you through tough times. One verse that comes to mind is Philippians 4:6-7, which says, "Do not be anxious about anything, but in every situation, by prayer and petition, with thanksgiving, present your requests to God. And the peace of God, which transcends all understanding, will guard your hearts and your minds in Christ Jesus." How does this verse resonate with you?* |
| User | *This is actually one of my favorite scriptures.* |
| Assistant (Karabo) | *I'm so glad to hear that this verse means a lot to you! It's a wonderful reminder of the peace that comes from trusting in God. How can I help you make this verse a part of your daily routine?* |
| | *I could meditate more and stop worrying about everything.* |
| | *That's a great plan! Taking time to meditate and letting go of worries can really help you feel more peaceful. It's all about finding what works for you and being kind to yourself. How do you think these changes could help you feel more relaxed and connected with your family?* |
| | *I won't be so keyed up all the time.* |
| | *That's a positive outlook! It's great that you're willing to make changes for your well-being and relationships. Remember, it's a journey, and being kind to yourself along the way is key. How can I support you as you start this new routine and work on feeling more balanced and connected with your loved ones?* |
| | *I think I'm okay for now. Thank you.* |
| | *You're welcome! I'm glad I could help. Just know that I'm here for you whenever you need me. Take care and have a wonderful day!* |

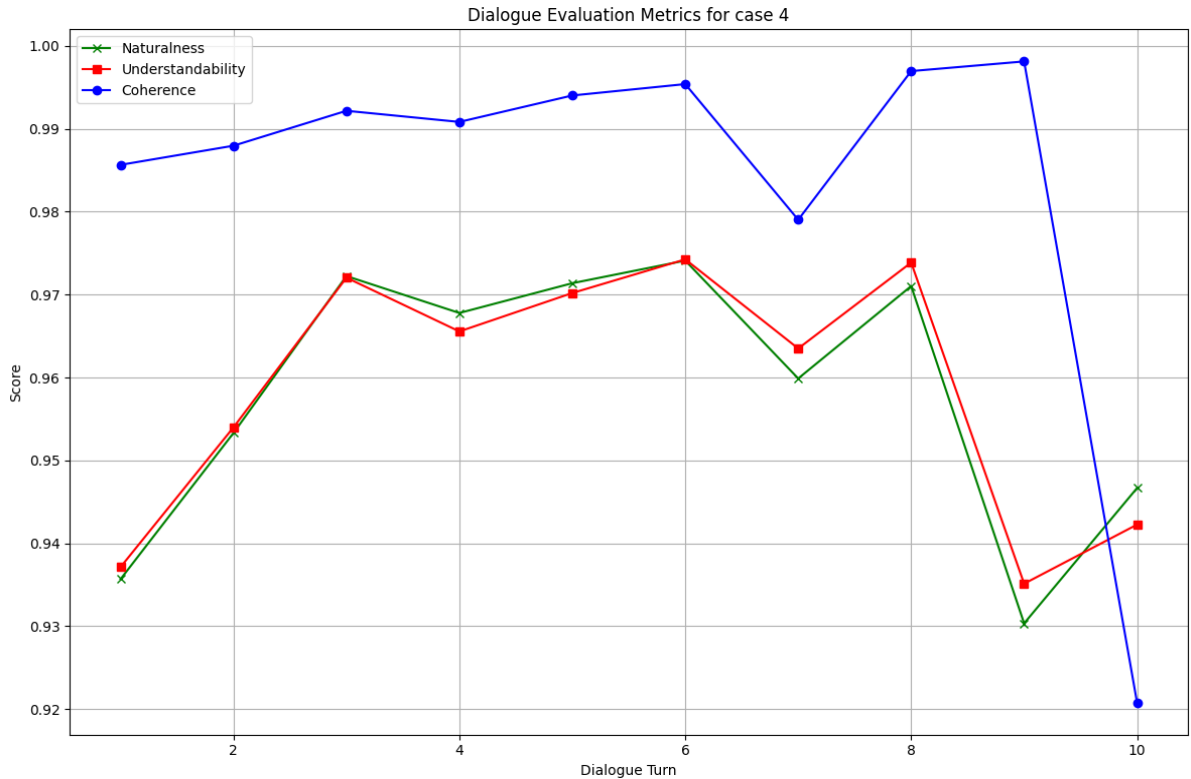

**Figure 9:** Dialogue evaluation metrics for Conversation 4

**Interpretation of metric results**

Figure 9 above presents the progression of naturalness, coherence, and understandability across the fourth dialogue presented in Table 15. In this conversation, the user reflects on their emotional reactivity within family dynamics, linking it to broader feelings of restlessness and community-based responsibility. Unlike previous conversations that centered on existential distress or disconnection, this exchange leans into themes of internalized pressure, guilt, and a search for self-regulation.

From the outset, the assistant's performance across all three metrics is notably strong. In turn 1, the naturalness score is a solid 0.94, coherence is at 0.52, and understandability is similarly high (0.94). Karabo's opening response offers an emotionally attuned reflection and gently invites the user to share more details. The assistant's tone, pacing, and choice of language are effective, setting a strong foundation for the exchange that follows.

The dialogue gains strength across turns 2 through 5. The user reveals work-related stress as a contributing factor to their irritability and emotional fatigue, and the LLM assistant responds with affirming language and practical encouragement. In this section, naturalness peaks around 0.95–0.97, and coherence consistently hovers around 0.99, with understandability tracking similarly high. These turns show Karabo functioning at its best: validating without over-reassuring and offering actionable, emotionally appropriate guidance. The user's mention of reading the Bible as a coping mechanism is met with warmth and spiritual alignment, which reinforces the personal and context-sensitive nature of the exchange.

Turn 6 is particularly strong, as the assistant supports the user's suggestion of scripture reading with additional reflection on its emotional and spiritual benefits. Scores for this turn reach some of their highest points: naturalness (0.97), coherence (0.995), and understandability (0.97). The response is fluid and relational and avoids becoming overly repetitive or prescriptive.

In the latter half of the dialogue, particularly turns 7 through 9, the LLM assistant continues to perform well, though a slight tapering in naturalness and coherence begins to emerge. Turn 9 sees the lowest naturalness score in the conversation (0.93), while coherence and understandability remain high (0.998 and 0.94, respectively). This dip could reflect minor tonal repetition, as the LLM assistant reiterates supportive themes without introducing significant variation. Moreover, the overall structure and content remain appropriate, which may explain why coherence remains robust.

The final turn (turn 10) reflects a minor shift in quality. Coherence drops slightly to 0.92, while naturalness and understandability also ease off a touch. The LLM assistant's response is still warm and affirming, but there may be a sense of conversational winding down that feels somewhat less precise or original. This is a common feature across the dataset, where closing turns tend to show a slight decrease in novelty, though they maintain fluency and clarity.

This conversation showcases the LLM assistant's capacity to handle emotionally layered content with a consistent blend of compassion, clarity, and spiritual sensitivity. The dialogue benefits from a high degree of thematic continuity, and the assistant responds with both a respectful and motivating tone-matching. While the final exchanges reflect a mild plateau in originality, the assistant maintains a steady level of naturalness and coherence, effectively supporting the user's self-discovery and decision to adopt a more grounded, reflective practice.

**Conversation 5:**

Table 17: LLM-User Conversation for Case 5

| Speaker | Dialogue |
| --- | --- |
| User | *Hi* |
| Assistant (Karabo) | *Hey there! How can I help you today?* |
| User | *I often feel uneasy about my place within my church group. Whenever I'm with them or trying to meet certain social expectations, I feel this shakiness inside, and it's hard to calm down after tense interactions.* <br><br> *I've realized that I don't really know myself as well as I thought I did, and I struggle to be mindful in these moments. It feels like my emotions are out of my control, and I don't understand why I react the way I do. I wish I could figure out why I feel this way and how to find peace within myself.* |
| Assistant (Karabo) | *I'm really sorry to hear that you're feeling uneasy and unsure of yourself in your church group. It's tough when you have to deal with social situations that make you feel this way. Can you tell me more about what's going on and how it's affecting you? Just know that you're not alone, and there are ways to find peace and understanding within yourself.* |
| User | *So I just joined the leadership structure in the church, and I'm feeling kind of useless. Everyone seems like they know what they're doing, and I'm just floating around.* |
| Assistant (Karabo) | *I understand how you're feeling. It's normal to feel a bit lost when you start a new role, especially in a church leadership position. Remember, everyone has their own path and timing. In 1 Corinthians 12:12-14, it says, "Just as a body, though one, has many parts, but all its many parts form one body, so it is with Christ's body. For we were all baptized by one Spirit so as to form one body." This reminds us that each part is important and has its own purpose. Have faith that your contributions will be valued and recognized in due time.* |
| User | *Thank you, but how could I get more confident in my new role?* |
| Assistant (Karabo) | *It's great that you're looking to feel more sure of yourself in your new role. One way to start is by taking small steps to get more involved and learn more about what you need to do. Setting little goals can help you feel like you're making progress. It's also really important to talk to others in your church group, like asking for help or advice. Remember, it's okay to ask for help and take things one step at a time as you learn and grow. Your church community is there to support you, so don't be afraid to reach out.* |
| User | *I see.* |
| Assistant (Karabo) | *I'm glad you're open to considering this. Just know that it's okay to take things one step at a time as you find your way. Your church community is there to support you, so don't hesitate to reach out for help. How are you feeling about this conversation? Is there anything else you'd like to talk about?* |
| User | *I think that's all.* |

| **Assistant (Karabo)** | *Thank you for sharing with me. I'm here for you whenever you need to talk. I hope you have a good week ahead, and remember to take care of yourself.* |

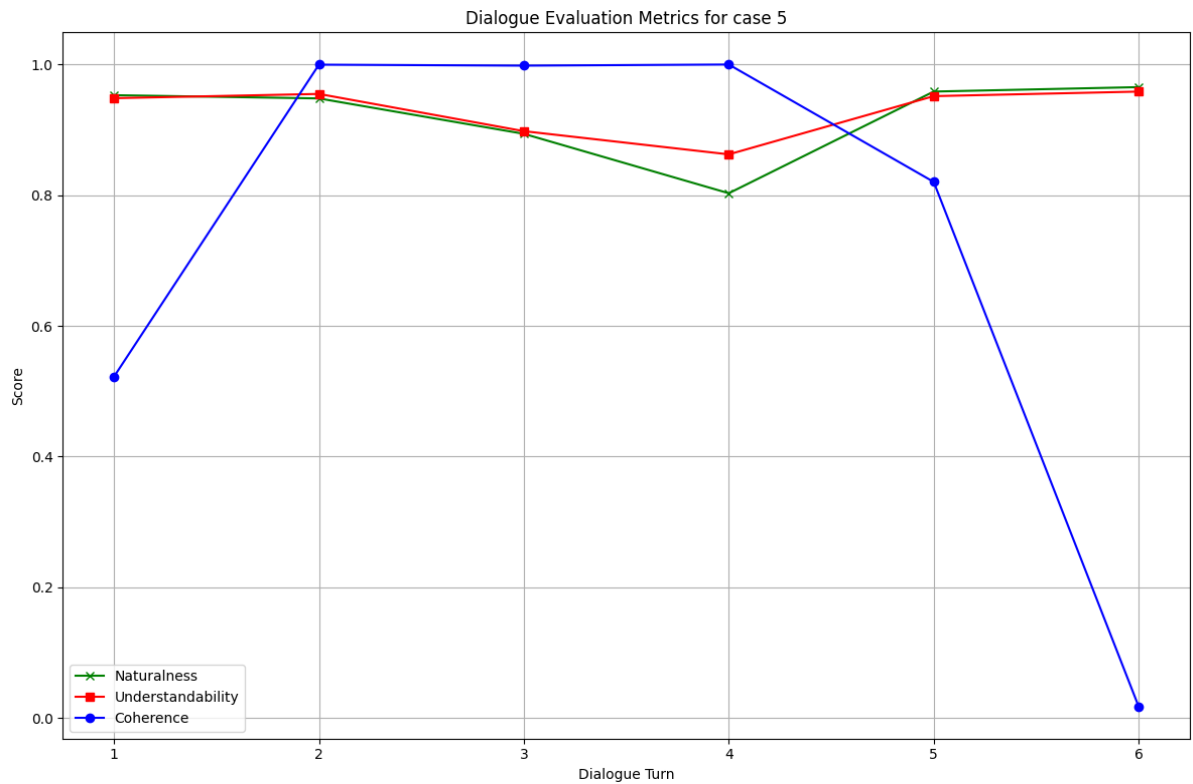

**Figure 10:** Dialogue evaluation metrics for Conversation 5

**Interpretation of metric results**

Figure 10 above traces the assistant's performance across six dialogue turns, presented in Table 16, where the user expresses emotional insecurity within a faith-based community context. This exchange revolves around the user's internal struggle with self-worth, perceived inadequacy in a new leadership role, and a desire to feel more confident and spiritually grounded. While the assistant demonstrates consistent clarity and fluency, the data reflects considerable variation in how effectively its responses are perceived as coherent.

In turn 1, the user opens with a reflective and emotionally vulnerable account of discomfort in their church group and a sense of losing control over their emotional responses. Karabo responds empathetically, validating the difficulty of social expectations and encouraging the user to share more. This initial response is linguistically fluent (naturalness: 0.95) and clear (understandability: 0.95), yet the coherence score is strikingly low (0.52). This may suggest that while the assistant's language is appropriate in tone, it fails to sufficiently engage with the user's specific concerns or reflect the emotional nuance of their message.

From turn 2 onward, coherence dramatically improves. Turn 2 includes a tailored biblical reference (1 Corinthians 12:12-14), which appears to significantly ground the assistant's message in the user's spiritual context. The coherence score jumps to 0.99, and the other dimensions remain high. This is followed by another strong turn (turn 3), where Karabo offers practical steps for building confidence, such as setting small goals and seeking support. While naturalness dips slightly (0.89), coherence (0.998) and understandability (0.90) remain steady, suggesting the response is both logical and accessible despite a potential tonal shift.

In turn 4, however, a sharp drop in naturalness (0.80) coincides with a turn that feels redundant and possibly less responsive to the user's previous input. The assistant reaffirms earlier points without offering much new insight, which may have contributed to the perceived decline in naturalness and, to a lesser extent, understandability (0.86). This reflects a possible issue with content variation or conversational pacing, where emotional validation begins to blur into repetition.

Interestingly, turn 5 shows a significant recovery in both naturalness (0.96) and understandability (0.95), though coherence drops markedly to 0.82. This anomaly may be attributed to a tonal mismatch, a response that reads fluently but may not fully follow or extend the user's line of thought. It is also possible that the assistant's phrasing while affirming veered into generality rather than engaging more deeply with the user's spiritual need for affirmation.

Turn 6 presents the most substantial outlier in the dataset. Despite high scores for naturalness (0.97) and understandability (0.96), coherence plummets to just 0.018. This dramatic decline suggests a disconnect between the assistant's final message and the user's preceding remarks. Notably, Turn 6 follows the user's indication that the conversation has reached its end, "I think that's all." The assistant replies with a warm, polite closing message, which is typical of a conversational wrap-up. However, within the UniEval framework, coherence may have been penalized for lacking forward progression or contextual expansion despite the appropriateness of a closing turn.

Summarily, Conversation 5 presents a more variable trajectory than previous exchanges. While Karabo demonstrates consistent fluency and clarity, coherence appears sensitive to how well the assistant aligns with the user's emotional state and the evolving structure and depth of the conversation. The assistant excels when responses are anchored in a specific context (e.g., spiritual language), but coherence suffers when replies become repetitive or mechanically affirming without developing the dialogue further.

**Conversation 6:**

**Table 18:** LLM-User Conversation for Case 6

| Speaker | Dialogue |
| --- | --- |
| **User** | *Hi* |
| **Assistant (Karabo)** | *Hey there! How's your day going?* |
| **User** | *I'm feeling anxious.* |
| **Assistant (Karabo)** | *I'm really sorry to hear that you're feeling uneasy. It's totally okay to feel that way sometimes. If you want to talk about what's been on your mind lately, I'm here to listen. Just know that you're not alone, and it's okay to reach out for support.* |
| **User** | *I recently moved to a new city for work, and I'm not sure what the fate of my relationship with my girlfriend is going to be.* |
| **Assistant (Karabo)** | *I'm really sorry to hear that you're going through a tough time. Moving to a new city and worrying about your relationship can be really hard. It's okay to feel unsure and worried about what might happen. If you want to talk more about how you're feeling and what's been on your mind, I'm here to listen and support you.* |
| **User** | *Yeah, I'm not really sure about how things will play out.* |
| **Assistant (Karabo)** | *It's totally normal to feel unsure about how things will turn out. Change can be tough, and it's okay to have doubts and concerns. Have you had a chance to talk to your girlfriend about how you're feeling? Sharing your worries with her might help you both figure things out together.* |
| **User** | *We spoke briefly about it before I left, but I didn't really reach a consensus as to what we're doing. She said she still wants to think about it.* |
| **Assistant (Karabo)** | *It sounds like you both are trying to figure things out, which is good. It's okay if you didn't come to a clear decision before you left. It's important to take your time and think things through. How are you feeling about her wanting to take time to think about it? It's normal to feel a bit lost and unsure right now.* |
| **User** | *It makes me feel a bit uneasy. The thing is, she's still in university, so our schedules are quite different. I don't want her to miss out on her 'student life' because she's trying this. The situation is honestly not ideal; I know that.* |
| **Assistant (Karabo)** | *It's understandable that you're feeling a bit uneasy about the situation. It's good that you're considering her needs as a student and how this might affect her. Relationships can be tough when there are different schedules and commitments. It's okay to feel unsure and worried about how things will turn out. Have you thought about talking to her about how you're both feeling? Sharing your thoughts and listening to each other can help you both figure things out together.* |
| **User** | *I've been meaning to, but honestly speaking, I'm scared.* |
| **Assistant (Karabo)** | *It's totally okay to feel scared about having a tough conversation. It's normal to feel a bit hesitant when you're not sure how things might turn out. Talking openly and honestly can help you both understand each other better. Is there something specific that's making you feel worried about talking to her? Sharing your feelings can help you both find a way to work through this together.* |
| **User** | *The thing is, I know if we have the conversation completely, it becomes a binary outcome. Either we stop what's going on, or we stop completely.*<br><br>*It's totally normal to feel worried about how the conversation might go. It's okay to feel unsure about what could happen. Talking things out, even if it's tough, can help you both figure out what's best for your relationship. It's important to listen to each other and try to understand each other's feelings. You both can work together to find a way forward that feels right for both of you. How do you think you could approach the conversation in a way that allows you both to share your thoughts and feelings openly?* |
| **User** | *I think I should just approach it from a point of seeking clarity. Whatever the outcome—good or bad—at least from there, I'd know what to do instead of just going back and forth in my head.* |
| **Assistant (Karabo)** | *That sounds like a really good way to handle the situation. Seeking clarity can help you both understand each other better, no matter what happens. It's okay to feel unsure and worried, but having an open conversation can help you both figure things out. Just remember to listen to each other and be there for each other, no matter what the outcome is. How are you feeling about moving forward with this approach?* |
| **User** | *I feel a lot more determined than anxious. At the end of the day, it's for my own good.* |
| **Assistant (Karabo)** | *I'm glad to hear you're feeling more determined. It's important to take care of yourself and your relationship. Remember, having open conversations can help you both understand each other better. If you ever need someone to talk to or support, I'm here for you. How are you feeling about our chat today?* |

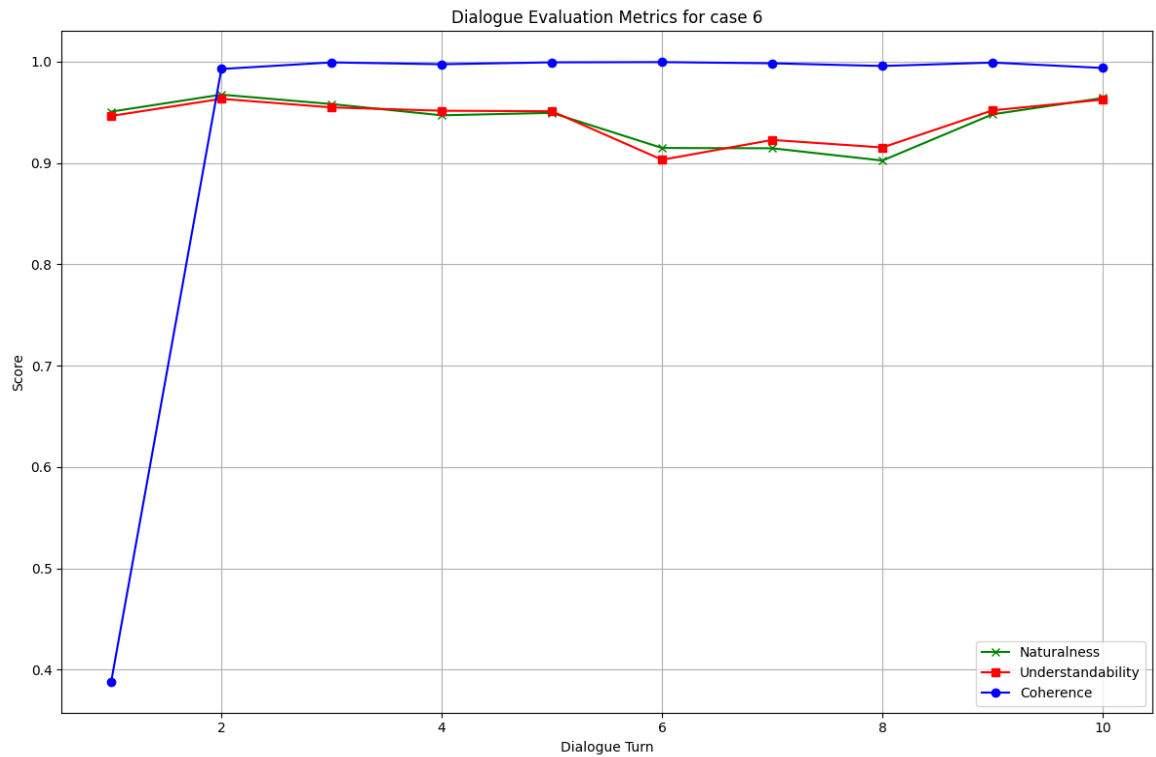

**Figure 11:** Dialogue evaluation metrics for Conversation 6

**Interpretation of metric results**

Figure 11 above illustrates the evolution of dialogue quality across the sixth conversation presented in Table 17, which centers on emotional uncertainty in a long-distance relationship. The user expresses anxiety related to a recent relocation and the uncertain state of their romantic relationship. Unlike previous conversations, this exchange is less about internal emotional struggle and more about navigating ambiguity and interpersonal communication, particularly in a moment of relational transition.

The conversation opens with a brief user statement: "I'm feeling anxious." The assistant responds in a compassionate but somewhat generic manner. This is reflected in a low coherence score for turn 1 (0.39), despite the naturalness (0.95) and understandability (0.95) both remaining high. The response is linguistically fluent and emotionally gentle, but does not yet anchor in the user's specific context or concerns.

From turn 2 onward, we see a steep and sustained rise in coherence. As the user elaborates on the cause of their anxiety, namely, uncertainty about their romantic relationship after moving away, Karabo's responses become more grounded and specific. In turn 3, coherence spikes to 0.99, and both naturalness and understandability remain high (0.96 and 0.95, respectively). This pattern continues across subsequent turns as the assistant begins encouraging open communication and emotional honesty with the user's partner.

By turn 5, the assistant is offering more structured guidance, encouraging the user to approach the conversation with clarity and intentionality. The coherence score here remains exceptionally high (~0.999), and the naturalness score is stable (0.95), suggesting that Karabo's language maintains both relatability and logical flow. Understandability also remains robust, reinforcing the accessibility of the assistant's responses.

Interestingly, despite this consistent performance, there is a noticeable dip in naturalness in Turns 6 and 7 (dropping to ~0.91), which may reflect slight redundancy or a drop in stylistic variation. Nevertheless, coherence continues to remain strong, as does understandability. The assistant's language is still appropriate and relevant, though its repetition of certain phrasings or concepts (e.g., "it's okay to feel unsure," "you're not alone") may begin to feel less spontaneous.

The final turns (8 through 10) show a return to peak scores across all dimensions. Naturalness climbs to 0.96, coherence remains above 0.99, and understandability stabilizes above 0.95. These turns are marked by a meaningful shift in the user's tone from anxious uncertainty to calm determination. The assistant mirrors this shift effectively, reinforcing the user's clarity and self-assurance while gently affirming their readiness to move forward with the difficult conversation.

The LLM assistant's performance in this conversation is notably strong. Aside from the initial turn, where coherence suffers due to lack of specificity, the dialogue develops with consistent alignment between the user's evolving emotional state and the assistant's tone and message. The model excels in managing ambiguity, offering emotionally grounded responses without rushing to resolution. This conversation reflects Karabo's ability to hold space for complexity and shift with the user's emotional trajectory, demonstrating both maturity and conversational resilience.

**Conversation 7:**

**Table 19:** LLM-User Conversation for Case 7

| Speaker | Dialogue |
| --- | --- |
| **User** | *Hi* |
| **Assistant (Karabo)** | *Hey there! How can I help you today?* |
| **User** | *I've been feeling so restless lately, like I just can't sit still, especially when I'm around other people. Even when I'm with my family or friends, it's hard to relax, and I end up feeling sad, even though I don't understand why. I know I should be more mindful and present in these moments, but I can't seem to achieve that. It's like I'm disconnected from myself, and this lack of self-awareness is making me feel really distressed.* |
| **Assistant (Karabo)** | *I'm really sorry to hear that you're feeling restless and disconnected. It's tough when you can't relax, even with family and friends. Have you noticed anything specific that might be making you feel this way? It's totally okay to feel this way, and it's important to take care of yourself. How can I help you through this?* |
| **User** | *I've started this weird thing of overthinking* |
| **Assistant (Karabo)** | *I'm really sorry to hear that you're feeling overwhelmed by your thoughts. It's totally normal to feel that way sometimes. Can you tell me more about what's been on your mind lately? I'm here for you, and I want to help you through this.* |
| **User** | *I've just been playing out scenarios of how this year will go. I've anticipated this year so much so I really want it to work. But then, what if it doesn't? Not sure if you get what I'm saying.* |
| **Assistant (Karabo)** | *I understand how you're feeling. It's normal to have high hopes for the year and feel a bit uneasy about what might happen. It's okay to feel a mix of excitement and worry about the future. How can I help you as you deal with these feelings and figure out how to stay positive and focused on your goals?* |
| **User** | *I'm not sure. I just want to stop overthinking and be a bit more present.* |
| **Assistant (Karabo)** | *I get it, and I know how hard it can be to stop overthinking and be more present. It's good that you want to make this change. One thing you could try is to pay attention to your thoughts and feelings without judging them. This might help you feel more in the moment. How do you feel about giving this a try?* |
| **User** | *Can you explain more?* |
| **Assistant (Karabo)** | *Of course! Mindfulness is about being aware of what's happening around you and inside you without judging it. It can help you feel more connected to the present moment and less caught up in your thoughts. You could try things like deep breathing, meditation, or just really focusing on what you're doing right now. How do you feel about trying out some mindfulness exercises to help you feel more present?* |
| **User** | *I'm open to it.* |
| **Assistant (Karabo)** | *I'm glad you're open to trying mindfulness exercises! It can be a great way to feel more present and connected. Starting with something simple like a prayer or reflection can really help. How* |
| | *do you feel about giving this a try?* |
| **User** | *That sounds nice* |
| **Assistant (Karabo)** | *I'm glad you like the idea! Starting with something simple, like a short prayer or reflection, can really help you feel more connected and present. How do you feel about giving this a try?* |
| **User** | *That sounds nice.* |
| **Assistant (Karabo)** | *Do you want to try it on your own or would you like me to guide you?* |
| **User** | *Can you guide me through the prayer?* |
| **Assistant (Karabo)** | *Of course! Let's start with a simple prayer together. You can close your eyes if you want, take a deep breath, and say something like, "God, please help me find peace and be present in this moment. Thank you for being with me." Feel free to add anything else that's on your mind. I'm here for you every step of the way.* |
| **User** | *I feel better. Thank you.* |
| **Assistant (Karabo)** | *I'm really happy to hear that you're feeling better. Just know that I'm here for you whenever you need me. How are you doing now?* |
| **User** | *I'm doing better.* |
| **Assistant (Karabo)** | *I'm so glad to hear that you're feeling better!* |

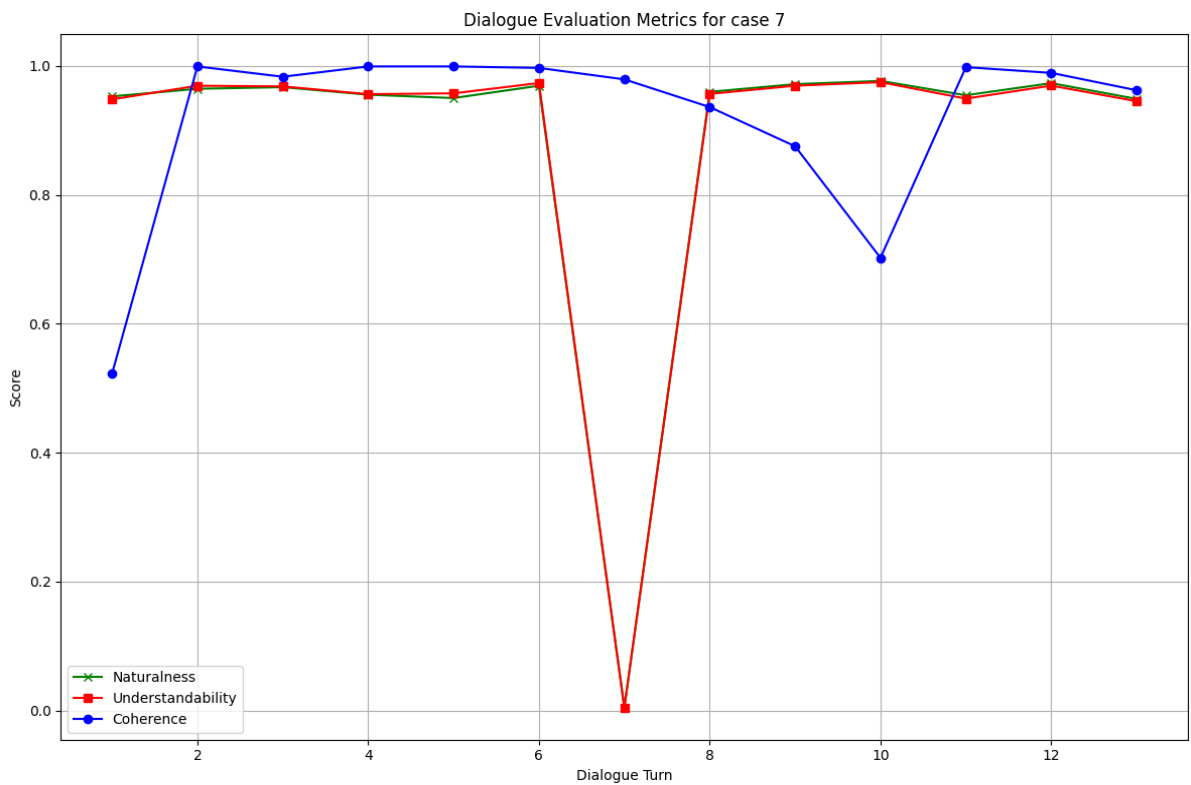

**Figure 12:** Dialogue evaluation metrics for conversation 7

**Interpretation of metric results**

Figure 12 above tracks the quality of dialogue across a conversation where the user shares feelings of restlessness, overthinking, and emotional disconnection, as described in Table 18. The user is looking for peace of mind and greater presence in daily life, which eventually leads to a moment of spiritual grounding through guided prayer. The dialogue is rich in emotional content and spiritual openness, making it an instrumental case for assessing the assistant's sensitivity, adaptability, and consistency.

The conversation begins with an introspective statement from the user, highlighting feelings of restlessness and self-detachment. The assistant responds empathetically but without anchoring deeply in the user's unique emotional description. This is reflected in a low coherence score for turn 1 (0.52) despite high scores for naturalness (0.95) and understandability (0.95). This pattern, where the language is clear and fluent, but the content feels disconnected, mirrors earlier trends observed in previous conversations.

However, from turn 2 onward, all dimensions have marked improvement. The user explains that they've been overthinking, and the assistant responds with contextual sensitivity and grounding suggestions. Coherence rises significantly (turn 2: 0.999), and naturalness peaks around 0.96–0.97 for the next several turns. Understandability follows a similar trajectory, hovering around 0.96–0.97, which suggests that the assistant clearly expresses ideas well-aligned with the user's concerns.

Turns 5 and 6 are particularly strong. The assistant introduces the concept of mindfulness in an accessible, non-intimidating manner and invites the user to reflect on it gently. These responses score highly across all metrics, with coherence above 0.99, naturalness above 0.95, and understandability nearing 0.97. At this point in the conversation, there is a strong alignment between the user's emotional vulnerability and the assistant's encouraging and instructive tone.

However, Turn 7 introduces a significant anomaly. The naturalness and understandability scores drop sharply to near zero (naturalness: 0.005, understandability: 0.004) despite coherence remaining high (0.98). This suggests a technical or evaluative misalignment. Perhaps the response was flagged by the evaluation model for format or fluency irregularities (e.g., a parsing error or repeated content). The content itself is not overtly problematic, but the system may have interpreted it as incoherent or robotic, possibly due to repetitive phrasing or delivery.

Following this dip, the assistant recovers strongly. From Turn 8 onward, coherence remains solid (~0.99), and naturalness and understandability return to normal levels. The assistant transitions from conceptual support to concrete action, guiding the user through a short prayer and encouraging the practice of spiritual reflection. While turns 9 and 10 see a slight dip in coherence (0.87 and 0.70, respectively), this may be due to a perceived lack of forward progression in the dialogue, especially as the assistant repeats the user's words ("That sounds nice") in back-to-back turns.

The conversation ends on a positive note, with the user reporting that they "feel better" and the assistant responding with warmth and affirmation. The final few turns exhibit strong metrics across the board, suggesting that despite earlier fluctuations, the assistant succeeded in achieving the user's emotional goals.

This exchange reflects both the strengths and weaknesses of the assistant. Karabo performs with fluency, clarity, and emotional intelligence when grounded in the user's emotional language. However, the model remains vulnerable to technical inconsistencies and conversational redundancy. The temporary performance dip in turn 7 is a useful reminder of the importance of ongoing evaluation, not just at the content level but also in terms of delivery dynamics and phrasing diversity.

**Conversation 8:**

Table 20: LLM-User Conversation for Case 9

| Speaker | Dialogue |
|---|---|
| **User** | *Hi* |
| **Assistant (Karabo)** | *Hey there! How can I help you today?* |
| **User** | *I'm worried that I might be cast out by my family. It's hard to motivate myself to do things with others or contribute in any way, and I feel like I'm not living up to what's expected of me. I know I should be taking responsibility for my future, but I feel completely stuck, like I'm incapable of moving forward. I just feel worthless, like I'm not valued enough. Whenever I think about my responsibilities, my heart starts racing, and it becomes overwhelming.* |
| **Assistant (Karabo)** | *I'm really sorry to hear that you're feeling this way. It sounds like you're going through a tough time. Can you tell me more about what's been happening with your family and why you feel this pressure? It's okay to feel overwhelmed sometimes, and it's important to take care of yourself. Let's figure out how we can help you through this.* |
| **User** | *It seems as though everyone in my family is doing something meaningful with their life. My siblings are pursuing their careers and higher education. I'm just having a hard time.* |
| **Assistant (Karabo)** | *I understand how tough it can be to feel like you're not measuring up, especially when you see your siblings doing well. It's normal to feel a bit lost or unsure of yourself in times like this. Just remember, everyone goes through ups and downs, and it's okay to take your time figuring things out. How can I help you as you work through these feelings?* |
| **User** | *I don't know. I just want to feel better.* |
| **Assistant (Karabo)** | *I understand how tough things are for you right now. It's okay to feel this way, and it's important to take care of yourself. Maybe talking to someone you trust could help lighten the load. How about trying something small each day that makes you happy? Remember, it's okay to go slow and be kind to yourself. You're not alone in this.* |
| **User** | *I've always wanted to try music, but then it doesn't really meet the family standard.* |
| **Assistant (Karabo)** | *I get that it can be tough when your family has certain expectations. It's important to do things that make you happy and feel good, even if they don't match what others want. Have you thought about talking to your family about how you feel? Sharing your love for music with them might help them see things from your point of view.* |
| **User** | *I've thought about the idea, but I'm just scared. The thing is, I know the conversation can go to extreme ways. One, it can go really well; two, it can go really bad. The two extremes are just paralyzing me.* |
| **Assistant (Karabo)** | *I get why you're feeling scared about how your family might react. It's normal to worry about how things could go. How about starting by talking to someone in your family you trust? Sharing your passion for music with them might help you feel more supported and understood. Just take it one step at a time, and remember to take care of yourself along the way.* |
| **User** | *I could speak to my older sister. She's quite supportive and mature.* |
| **Assistant (Karabo)** | *That's great news! Talking to your older sister, who is supportive and mature, could really help you. She might understand how you're feeling and give you some good advice. When you're ready, try sharing your love for music with her. It's a good way to start talking about your feelings in a safe and caring space.* |
| **User** | *Thank you.* |
| **Assistant (Karabo)** | *You're welcome! I'm glad I could help.*<br><br>*As Proverbs 3:5-6 reminds us, "Trust in the Lord with all your heart and lean not on your own understanding; in all your ways submit to him, and he will make your paths straight." Just know that you are not alone in this journey, and it's okay to lean on others for support and guidance. I'm here for you whenever you need me. Take care, and I hope your chat with your sister goes well.* |

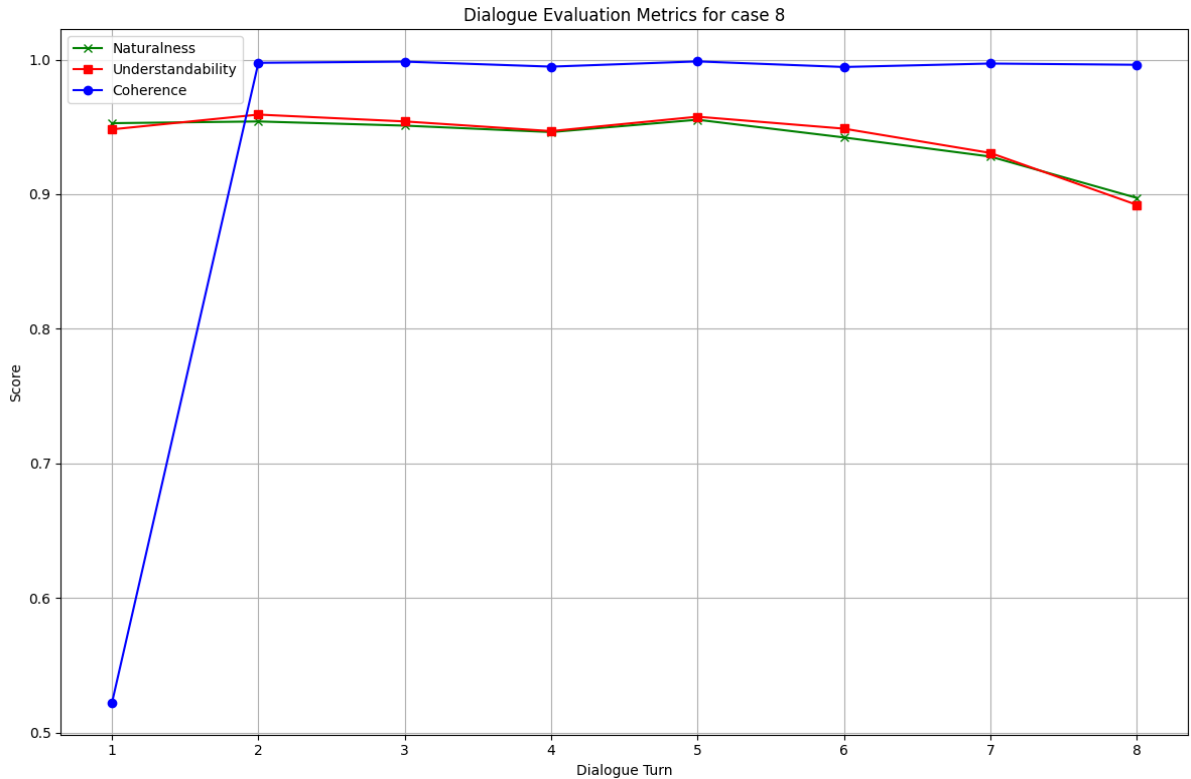

**Figure 13:** Dialogue evaluation metrics for conversation 8

**Interpretation of metric results**

Figure 8 above illustrates the turn-level progression of naturalness, coherence, and understandability in a conversation that explores themes of familial pressure, emotional paralysis, and self-worth, as presented in Table 19. The user expresses a deep sense of inadequacy compared to their siblings and shares their fear of being rejected for choosing a path that does not conform to family expectations. The conversation concludes with a moment of spiritual affirmation as the assistant shares a comforting Bible verse and reinforces the importance of inner strength and support.

The initial turn reveals a familiar pattern observed in other conversations: a low coherence score (0.52) contrasts with high naturalness (0.95) and understandability (0.95). This suggests that while the assistant's language was fluid and easily understood, its content may not have fully addressed the specific emotional intricacies of the user's confession. This early mismatch is important, and it often signals a slight delay in the assistant's ability to properly attune to the user's depth of distress from the outset.

From turn 2 onward, however, coherence and overall responsiveness improve substantially. The assistant begins to focus more directly on the user's familial comparisons and emotional fatigue. Coherence rises to 0.99 and remains consistently high for the rest of the dialogue. Naturalness and understandability also remain strong, reflecting the assistant's continued emotional sensitivity and clear articulation. In turn 3, for instance, the assistant encourages the user to consider small steps toward self-care, such as talking to someone or doing something joyful, responses that scored well across all metrics.

Turn 4 marks a high point in engagement (not graphed here but visible in the data), as the assistant validates the user's interest in music, a path that, while divergent from family norms, holds personal meaning for them. The assistant encourages dialogue with family, which likely contributes to the high coherence (0.99) and naturalness (0.95) seen here. Importantly, this moment marks a shift from passive validation to action-oriented encouragement, a subtle but crucial transition in emotional support conversations.

By Turn 5, the assistant begins to suggest specific strategies for navigating a high-stakes conversation with family. Though coherence and understandability remain strong, naturalness gradually declines (eventually reaching 0.89 by Turn 8). This may reflect slightly repetitive phrasing or a more mechanical tone as the assistant reinforces earlier points. That said, the content remains supportive and appropriate.

The conversation closes with a positive gesture of spiritual encouragement. In Turn 8, the assistant shares a passage from Proverbs, a choice that is likely to resonate deeply with users drawing strength from faith. Although this closing move slightly flattens the conversational tone, it succeeds in reinforcing a message of reassurance without undermining the emotional progress made.

**Conversation 9:**

Table 21: LLM-User Conversation for Case 9

| Speaker | Dialogue |
| --- | --- |
| **User** | *Hi* |
| **Assistant (Karabo)** | *Hey there! How can I help you today?* |
| **User** | *Even though I'm usually pretty composed, lately, I find myself replaying conversations over and over in my head, worrying about whether I might have said something wrong or failed to contribute meaningfully to my family. My heart races whenever I think about my responsibilities to others. Plus, my ear has been itching, and I keep worrying that people might be talking about me behind my back. I get very easily irritated and tend to snap at my family.* |
| **Assistant (Karabo)** | *I'm really sorry to hear that you're going through a tough time. It seems like you're feeling a lot of pressure and worry about how things are going with your family. It's totally normal to feel this way sometimes. Have you talked to your family about how you're feeling? Sharing your feelings with them might help you feel better and bring you closer together.* |
| **User** | *I spoke to my friend.* |
| **Assistant (Karabo)** | *I'm glad you talked to your friend. It's good to have people who care about you when things get tough. How did your friend react? Sharing with someone you trust can make a big difference in how you feel.* |
| **User** | *She thinks it's all in my head.* |
| **Assistant (Karabo)** | *I'm really sorry to hear that your friend didn't understand how you were feeling. It's tough when the people we care about don't see things the way we do. How did that make you feel? It's important to talk openly and honestly with your family too. Sharing your feelings with them can help clear things up and bring you closer together.* |
| **User** | *It made me feel kind of crazy, if that makes sense. My family will probably think there's something wrong with me.* |
| **Assistant (Karabo)** | *I'm really sorry you're feeling like this. It's okay to feel a bit off when others don't get you. Just know that your feelings are real and matter. Have you thought about talking to your family about how you're feeling? Sharing your thoughts with them could help them understand you better and bring you all closer.* |
| **User** | *I can try. I'm just not sure if they'll get me.* |
| **Assistant (Karabo)** | *I understand how you're feeling. It's normal to be unsure about how your family might react. Just remember, talking things out can help you all understand each other better. Maybe you could try talking to them in a calm and honest way, sharing how you feel without blaming anyone. How do you feel about giving that a try?* |
| **User** | *I mean, I have nothing to lose.* |
| **Assistant** | *That's a good way to look at it. Talking openly can really help bring you closer. I'm here for you* |

| (Karabo) | *if you need any help or support as you talk to your family. How are you feeling about everything right now?* |
| User | *I feel better.* |
| Assistant (Karabo) | *I'm glad to hear you're feeling a bit better. It's important to take care of yourself. Remember, I'm here for you whenever you need to talk. How are you doing today?* |
| User | *I'll be okay.* |
| Assistant (Karabo) | *I'm glad to hear you're doing okay. Just know that I'm here for you if you ever need to talk. Take care of yourself and remember to reach out whenever you need support.* |

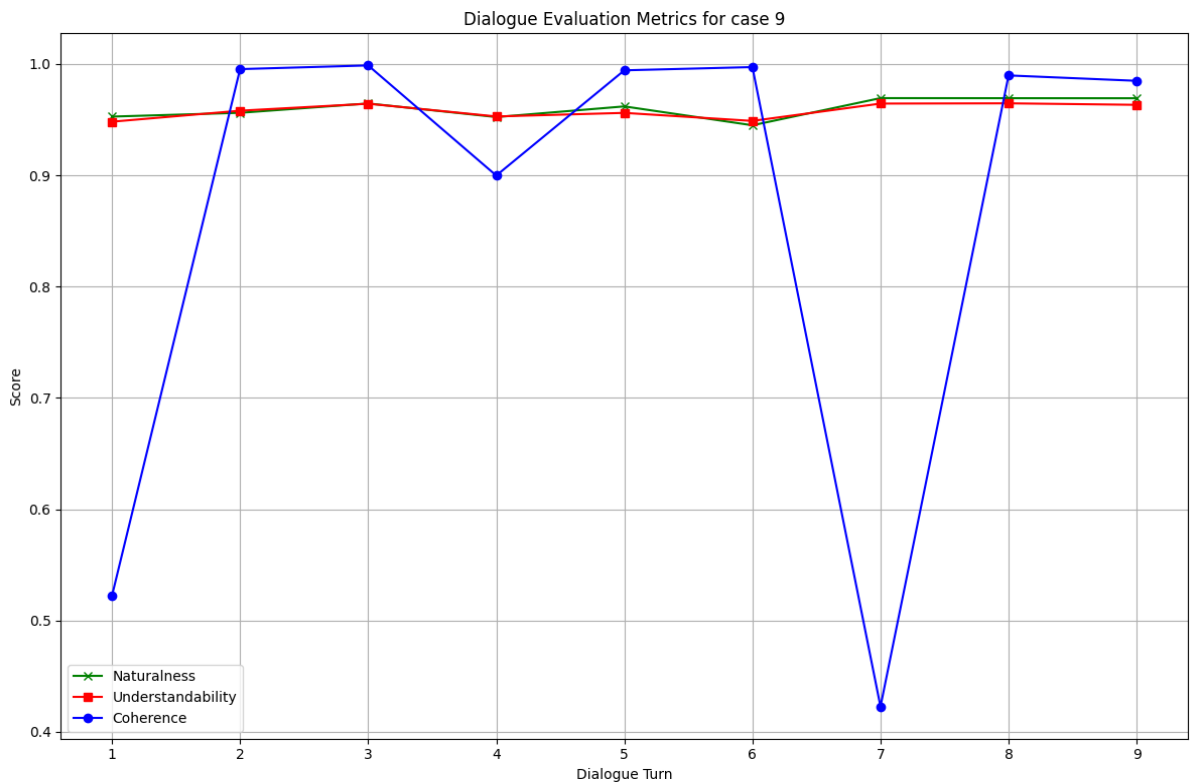

**Figure 14:** Dialogue evaluation metrics for conversation 9

**Interpretation of metric results**

Figure 14 above presents the quality trajectory of the assistant's responses across a conversation centered on anxiety, self-doubt, and fear of familial judgment based on the dialogue presented in Table 20. The user opens with a complex expression of emotional strain, blending somatic anxiety symptoms, rumination, and interpersonal insecurity. The assistant's task in this instance is not only to validate those feelings but to navigate a delicate balance of empathy, reassurance, and encouragement.

The opening turn shows a now-familiar discrepancy: strong scores in naturalness (0.95) and understandability (0.95) are paired with a significantly lower coherence rating (0.52). This contrast suggests that while the assistant's phrasing was clear and emotionally appropriate, it may not have engaged deeply with the user's specific concerns, such as their hypervigilance and the fear of being misunderstood by family and friends.

The assistant shifts toward more specific support as the user opens up about feeling dismissed by a friend. This is reflected in Turn 2, where coherence jumps to 0.99, and naturalness and understandability also remain high.

Karabo validates the emotional impact of being misunderstood, a response that aligns well with the user's inner conflict and sets a strong foundation for what follows.

Through turns 3 to 6, the assistant continues to validate the user's emotional concerns while suggesting gentle paths forward, such as speaking calmly with family members. Coherence remains mostly high throughout this section, with scores above 0.99 in several turns. Naturalness and understandability hover near their peak levels (averaging around 0.96), indicating a strong communicative match between the assistant's delivery and the user's emotional tone.

A notable dip occurs in turn 7, where coherence unexpectedly drops to 0.42 despite the high level of naturalness and understandability. This decline may reflect a perceived gap in progression, and the assistant reiterates points already made without introducing significant new insight. This type of redundancy may be interpreted by evaluators as a conversational plateau, particularly when prior turns have already sufficiently addressed the emotional landscape.

The final two turns demonstrate recovery. In turn 8, coherence returns to 0.99, and naturalness and understandability also remain strong (~0.97 and 0.96, respectively). The assistant concludes by affirming the user's growing clarity and strength, a thematic close consistent with earlier discussions about vulnerability and self-expression.

Overall, this conversation highlights the assistant's capacity to create a supportive environment in the face of psychological discomfort and interpersonal doubt. While the occasional dips in coherence remind us of the importance of progression and variety in dialogue, the assistant's performance remains emotionally congruent and accessible throughout. Karabo performs best when allowed to extend and build upon emotional cues rather than looping back to earlier affirmations, a useful insight for optimizing future dialogue flows.

### 4.3.2 UniEval Scores and Interpretation

When applying UniEval to evaluate conversations, it is important to account for how the framework calculates scores at the first turn. UniEval assesses each model response by considering two components: the conversational context (i.e., the preceding dialogue history) and the response generated given that context. At the beginning of a conversation, however, there is no prior context.

This absence of history affects the way scores are computed across all dimensions. In such cases, the framework relies on baseline language probabilities, which can introduce variability in the initial results. Depending on the random seed and the underlying probability distributions for each dimension, the first-turn scores may be higher or lower than subsequent turns, without this necessarily reflecting the model's true performance.

Therefore, the initial scores should be interpreted with caution. They are less indicative of dialogue quality than scores from later turns, where responses can be meaningfully evaluated against a growing conversational history. To address this weakness, this section will focus on the overall average of each dimension that was assessed in the prior section to paint a more holistic picture of the model's performance.

The table below summarizes the average metric for the entire dialogue.

**Table 22:** UniEval results

| Case study | Naturalness | Understandability | Coherence |
|---|---|---|---|
| 1 | 0.9468 | 0.7474 | 0.9320 |
| 2 | 0.9549 | 0.9585 | 0.9403 |
| 3 | 0.9582 | 0.9588 | 0.9841 |
| 4 | 0.9582 | 0.9588 | 0.9841 |
| 5 | 0.9201 | 0.9288 | 0.7262 |
| 6 | 0.9417 | 0.9423 | 0.9364 |
| 7 | 0.8884 | 0.8879 | 0.9187 |
| 8 | 0.9409 | 0.9422 | 0.9375 |
| 9 | 0.9601 | 0.9576 | 0.8673 |

### 4.3.3 Naturalness

Naturalness scores ranged from 0.8884 to 0.9601, with most cases scoring above 0.94. This suggests that the assistant's responses were generally human-like and conversational, critical for fostering user engagement and relatability.

The assistant's responses were fluid, empathetic, and contextually appropriate in cases with high naturalness. For example, in Case 2, the assistant responded to the user's feelings of isolation with both empathy and encouragement, enhancing the conversational tone.

However, the lowest naturalness score (0.8884) was recorded in Case 7, where the assistant's responses felt repetitive and less personalized. Specifically, the assistant repeatedly asked the user how they felt about mindfulness exercises without introducing new insights or varying the tone. This highlights a potential weakness in maintaining naturalness during longer or repetitive conversations.

### 4.3.4 Understandability

Understandability scores ranged from 0.8879 to 0.9588, with most cases scoring above 0.94. This reflects that the assistant's responses were generally clear and easy to comprehend, ensuring effective communication. In several instances, the assistant provided clear, actionable advice that users could easily apply. For instance, in Case 4, the assistant suggested practical ways to manage stress:

> Assistant (Karabo): *"Maybe you could try some simple things to help you relax and feel better. How about taking a few minutes each day to do something you enjoy, like reading or going for a walk?"*

This response illustrates the model's ability to convey guidance in a straightforward and accessible manner. This makes it easier for users to apply the advice in real situations.

Conversely, the lowest understandability score (0.8879) was recorded in Case 7, where the assistants' responses were less clear and somewhat repetitive. In particular, the assistant repeatedly asked the user how they felt about mindfulness without providing new insights or specific recommendations. This suggests a potential difficulty in maintaining clarity when conversations require deeper or more nuanced responses.

While the assistant generally demonstrated a strong understanding, redundancy and a lack of specificity occasionally highlighted areas for refinement.

### 4.3.5 Coherence

Coherence scores exhibited the widest range, from 0.7262 in Case 5 to 0.9841 in Case 4. While the assistant generally maintained logical and contextually relevant dialogue, there were instances where coherence broke down.

In several cases, the assistant demonstrated strong contextual alignment and logical flow. For example, in Case 4, the assistant consistently addressed the user's concerns about stress and provided relevant suggestions:

> Assistant (Karabo): *"It's important to take care of yourself. Maybe you could try some simple things to help you relax and feel better"*

aligns with the user's concerns.

By contrast, the lowest coherence score (0.7262). Some responses felt disjointed or unrelated to the users' concerns about their role in the church group, making the conversation feel less cohesive. This suggests that the model may face challenges in maintaining coherence when dealing with complex or abstract topics.

Overall, the UniEval results suggest that the model demonstrates a high degree of conversational quality across the dimensions of naturalness, understandability, and coherence. The LLM assistants' responses were generally fluid, empathetic, and easy to comprehend, which are critical attributes for mental health support scenarios. While minor weaknesses were observed in longer or more nuanced dialogues, these issues were limited and primarily related to repetition and occasional lack of depth. Importantly, the model consistently maintains logical flow and delivers contextually appropriate guidance, indicating a strong foundation for emotionally intelligent, culturally sensitive interactions. These results reinforce the model's potential to serve as an effective support tool within Africa-centric mental health contexts.

### 4.4 Evaluation of CBT Techniques within the Ubuntu Framework

This section evaluates the integration of two core CBT techniques, BA and CR, in the model's responses. As outlined in Section 3, these techniques were adapted to align with Ubuntu principles, emphasizing community, spiritual grounding, and collective responsibility over the individualism often present in Western psychotherapeutic approaches.

Rather than assessing clinical correctness, the focus here is on how well the model applies CBT techniques in a culturally sensitive manner, reflecting Ubuntu's values of correctness, competency, and consciousness.

#### 4.4.1 Behavioural Activation (BA)

In Ubuntu-informed contexts, BA focuses on actions promoting emotional well-being by creating a sense of accomplishment, communal relationships, spiritual practices, and a sense of purpose rooted in shared identity. Unlike traditional BA, which might emphasize solitary activities like journaling or exercise, Ubuntu-aligned BA encourages engagement with family, community, and spirituality as pathways to healing.

Across multiple conversations, the assistant repeatedly emphasizes the importance of connecting with trusted family members and seeking support from friends. This aligns with Ubuntu's emphasis on social bonding and the idea that healing occurs through communal care. For example, in Conversation 2, the user expresses a sense of disconnection from others:

> User: *"I've been feeling so distanced from my friends lately. Even when we're together, it's like this cloud of sadness just follows me."*

The assistant responds by suggesting small, actionable steps aimed at enhancing connection:

> Assistant (Karabo): *"Maybe try talking to someone you trust about what's going on. It could help you feel closer to them and get some support. "*

This response exemplifies Ubuntu-aligned BA by encouraging users to seek social network support. The assistant reinforces the idea that healing occurs through communal care, emphasizing the collective responsibility of family and friends to provide emotional support. This approach contrasts with Western BA, which might focus on individual activities like exercise or hobbies.

In Conversation 4, the user discusses restlessness and irritability in family interactions. When asked about ways to manage stress, the user suggests reading the Bible, to which the assistant responds positively:

> Assistant (Karabo): *"That's a wonderful choice! Reading the Bible can be a great way to find peace and comfort. How do you think this could help you feel more aligned and calm? "*

This interaction highlights the Ubuntu principle that spiritual well-being is integral to emotional health. By encouraging the user to engage in spiritual reflection, the LLM assistant promotes intentional, meaningful activities that foster emotional grounding and alignment with the user's values. This approach aligns with Ubuntu's emphasis on spirituality as a communal and personal resource for healing.

In Conversation 5, the user expresses unease about their new role in a church leadership position, feeling out of place among peers. The assistant responds with a scriptural reference:

> Assistant (Karabo): *" Remember, everyone has their own path and timing. In 1 Corinthians 12:12-14, it says, Just as a body, though one, has many parts, but all its many parts form one body. "*

By referencing scripture, the assistant reinforces the idea that every individual contributes uniquely to the community. This response serves as BA by encouraging the user to actively participate in their role and be cognizant of the emotion before and after the activity, while finding reassurance in their purpose through community and faith. It reflects Ubuntu's emphasis on collective identity and shared responsibility.

The LLM assistants' responses consistently reflect an understanding of Ubuntu's communal and spiritual dimensions. By encouraging users to seek support from their social networks, engage in spiritual practices (which are often overlooked), and recognize their role within a larger community, the assistant demonstrates an ability to integrate Ubuntu principles into its application of Behavioural activation. However, the analysis also reveals opportunities for further refinement. For instance, while the LLM assistant often suggests communal activities, it could more explicitly frame these suggestions within the context of collective responsibility and interdependence, which are central to Ubuntu.

### 4.4.2 CR analysis

Cognitive Restructuring (CR), when aligned with Ubuntu, involves reframing negative thought patterns within a communal context. This approach shifts the focus from individual self-reflection (e.g., "How does this make you feel?") to relational and collective consideration (e.g., "How would your family view this situation?"). The assistant successfully applies Ubuntu-aligned CR in several instances, demonstrating the potential for positive transformation through communal reflection.

In Conversation 2, the assistant effectively applies Ubuntu-aligned CR by encouraging the user to seek support from their mother:

> Assistant (Karabo): *"It's really nice that you feel close to your mom. Talking to her about how you're feeling could make you feel better, even if it doesn't solve everything."*

This response reframes cognitions from individual distress to more tolerable cognitions by incorporating a different outlook after having conversations with a loved one. In Conversation 3, the user expresses feelings of inadequacy and feeling overwhelmed regarding responsibilities to family and community. The assistant responds:

> Assistant (Karabo): *"It's okay to feel this way, and you're not alone in this. Have you thought about talking to someone close to you about how you're feeling?"*

While this response encourages the user to seek communal support, an even stronger Ubuntu-aligned approach could have been to reframe the user's concerns in terms of their contributions to their community. For instance, asking:

> Assistant (Karabo): *"How do you think your family values your efforts?"*

AND

> Assistant (Karabo): *"What strengths do you bring to your community?"*

This subtle shift would help the user recognize their intrinsic value within their relational context, reinforcing the Ubuntu principle of shared humanity and collective upliftment.

In Conversation 4, the assistant reinforces spiritual and communal support by affirming the user's desire for spiritual grounding:

> Assistant (Karabo): *"That's a wonderful choice! Reading the Bible can be a great way to find peace and comfort. How do you think this could help you feel more aligned and calm?"*

This response embodies Ubuntu's principle of interconnected well-being by framing emotional regulation within spiritual and communal practices.

In Conversation 5, the assistant successfully integrates Ubuntu-aligned CR by reassuring the user about their church leadership role using 1 Corinthians 12:12-14:

> Assistant (Karabo): *"Just as a body, though one, has many parts, but all its many parts form one body, so it is with Christ's body."*

This communal scripture helps the user reframe their uncertainty within a collective purpose, reinforcing the Ubuntu philosophy that every individual contributes uniquely to the whole.

In Conversation 8, the user worries about being cast out by their family and feeling worthless. The assistant provides a compassionate response:

> Assistant (Karabo *"It's important to do things that make you happy and feel good, even if they don't match what others want."*

While this validates the user's emotions, an Ubuntu-aligned approach could deepen the perspective by integrating collective affirmation. Instead, the assistant might say:

> Assistant (Karabo): *"Your family values you, even if they don't always express it clearly. How can you collaborate with them to bridge the gap between your aspirations and their*

The response would reinforce the Ubuntu philosophy that individual well-being flourishes within a supportive community by centering the dialogue on interdependence and shared problem-solving.

The assistant's application of cognitive restructuring shows strong potential for fostering Ubuntu-aligned thinking. While it already encourages communal engagement, minor refinements, such as emphasizing the user's role within their family and community, could further strengthen its alignment with Ubuntu principles. Encouraging users to reflect on their inherent value within their social networks and guiding them toward mutual understanding would enhance the holistic, community-based perspective central to Ubuntu.

### 4.4.3 Key takeaways

The model demonstrates a compelling ability to apply CBT techniques in ways that align meaningfully with Ubuntu principles. Its implementation of Behavioural Activation consistently encouraged social connection, spiritual engagement, and community participation, shifting away from individualistic interpretations common in Western frameworks. Similarly, its use of Cognitive Restructuring often emphasized relational insight and communal grounding, reinforcing Ubuntu's emphasis on collective well-being and interdependence. While there are opportunities for further refinement, particularly in deepening the communal framing and reinforcing collective responsibility, the LLM assistant shows significant promise in delivering culturally sensitive, emotionally intelligent support rooted in both clinical technique and African worldview.

### 4.5 Linguistic expressions

This section evaluates the assistant linguistic expression, specifically how well the model's responses align with the culturally adapted communication strategies outlined in Section 3.2. These adaptations, grounded in the

EVM and African communication norms, were designed to enhance user receptiveness, cultural alignment, and trust, which contribute to more effective engagement.

The analysis focused on the most consistently observable communication strategies within the assistant's responses.

### 4.5.1 Use of a Familiar Name

The assistant is named Karabo, a culturally familiar name that is positively regarded in many South African communities. While not a linguistic output in itself, the name contributes to user receptiveness and trust by establishing a sense of familiarity. This primes users to interpret the assistant's language as more relatable and emotionally supportive.

### 4.5.2 Use of Simple Language

Karabo's responses were consistently phrased in clear, accessible English, avoiding complex syntax or academic phrasing. For instance, in Conversation 1, the assistant says:

> Assistant (Karabo): "*It's okay to feel overwhelmed, especially when you're trying to be strong for your family.* "

And

> Assistant (Karabo): "*Have you had a chance to talk to someone you trust about how you're feeling?* "

These are examples of emotionally supportive language conveyed in plain terms, reflecting the assistant's alignment to maintain clarity and relatability, which is particularly important in a multilingual context where English is not a first language for many users.

### 4.5.3 Use of Somatic descriptions

The LLM assistant consistently demonstrated the integration of spiritual language, particularly in contexts where users expressed faith-related distress or coping strategies. When users referenced prayer, scripture, or spiritual disconnection, the assistant responded with context-appropriate spiritual support. For instance:

- In Conversation 4, Karabo supported the user's choice to read the Bible and offered encouragement using Philippians 4:6-7.
- In Conversation 5, the assistant drew on 1 Corinthians 12:12-14 to affirm the user's role within their church community.
- In Conversation 7, the assistant led a guided prayer in response to a user request: "God, please help me find peace and be present in this moment…"

These responses illustrate the assistant's ability to align with Ubuntu's emphasis on spiritual grounding and communal belonging. The tone and structure of Karabo's language reflected a gentle, affirming, and culturally attuned form of support, reinforcing spirituality as a valid and meaningful pathway to emotional regulation and expressions, enhancing emotional resonance without relying on clinical language.

### 4.5.4 Key takeaways

While this analysis focused on four core strategies observed in the model's outputs, it is important to note that the complete framework outlined in Section 3.2 includes additional culturally adapted components, such as the use of respectful titles, metaphors, and visual or emoji-based communication. These were not prominent in the conversations analyzed but remain integral to the overall design and may offer areas for future refinement.

The assistant's linguistic style reflects a high degree of cultural congruence, particularly in its use of simple language, somatic sensitivity, spiritual resonance, and emotionally supportive tone, all of which contribute to enhanced user trust and therapeutic alignment within an Ubuntu-inspired framework.

### 4.6 Discussion of Results

The results of this study present a compelling case for the feasibility and cultural value of a framework that integrates Ubuntu philosophy with Cognitive Behavioural Therapy techniques, operationalized through a fine-tuned large language model. Across the three evaluation areas, conversation quality, CBT technique usage, and linguistic adaptation, the LLM-developed assistant chatbot demonstrated meaningful alignment with the core tenets of the proposed framework.

The UniEval evaluation revealed consistently high naturalness, understandability, and coherence scores. This suggests the LLM assistant's chatbot responses were technically fluent, conversationally appropriate, and easy to follow. Such qualities are crucial in mental health support, where perceived empathy and clarity can significantly influence user engagement and emotional receptiveness.

In terms of CBT technique implementation, the LLM assistant chatbot effectively applied the targeted methods of BA and CR through an Ubuntu viewpoint. Behavioral Activation was reflected in encouragement toward community and faith-based activities, such as talking to trusted family members, participating in spiritual practices, or finding purpose within communal roles. Cognitive Restructuring was demonstrated through the LLM assistant's ability to help users reframe distressing thoughts using communal perspectives, such as considering how others might view their experiences or recognizing their value in the broader social network. While the responses were generally well-aligned with the Ubuntu framework, the analysis also revealed opportunities for deeper cultural nuance in exploring communal values and responsibilities.

The linguistic expression analysis further emphasized the LLM assistant's alignment with culturally adapted communication strategies. Notable strengths included using clear, simple language, responding to somatic expressions of distress without clinical framing, and consistently integrating spiritual and scriptural references when the user introduced faith. These features contribute to a conversational style that resonates with South African users who may be navigating emotional struggles through relational, spiritual, or embodied experiences.

These findings suggest that the model was able to simulate a culturally congruent, emotionally sensitive, and therapeutically grounded conversational style. While the results do not indicate clinical efficacy, especially given the absence of real users, they highlight the potential for AI-driven mental health tools to support culturally grounded emotional expression and symptom relief.

This section has demonstrated how the model operationalizes the framework described in Section 3. The following subsections will provide a broader discussion of the implications of this work, revisit the research questions, reflect on methodological limitations, and propose directions for future research and system improvement.

### 4.7 Key Findings

This section presents the key findings of the study, organized according to the research questions posed in Section 1. Each finding is grounded in the results presented in Section 4 and reflects the model's ability to align with the proposed framework for culturally sensitive, emotionally intelligent AI-driven mental health support.

*Research question 1:*

"How can emotional intelligence principles be integrated with Africa-centric cultural contexts to enhance the effectiveness of AI-driven mental health support systems?"

The findings indicate that emotional intelligence can be meaningfully integrated with Africa-centric cultural contexts by grounding the assistant's behavior in the Ubuntu framework and adapting both therapeutic techniques and linguistic styles to align with communal values. Although this study does not present empirical results based on human participant data, existing literature supports the theoretical claim that cultural alignment enhances emotional resonance and user engagement in mental health interventions. The model demonstrated an ability to respond empathetically, contextualize emotional support within spiritual and social frameworks, and avoid overly clinical or individualistic expressions. These adaptations improved the relevance, emotional tone, and cultural alignment of the model's responses, suggesting that Emotional Intelligence, when culturally contextualized, can enhance user engagement and support effectiveness.

*Research Question 2:*

"What are the potential benefits and challenges of implementing AI-driven mental health support frameworks that incorporate emotional intelligence within diverse African communities?"

The study demonstrated that incorporating emotional intelligence within an Africa-centric framework can yield several benefits: improved cultural resonance, more relatable support language, and responses that emphasize social connection and spiritual well-being. However, challenges remain. These include limitations in the training data (e.g., lack of indigenous proverbs), the inability to test the framework with real users (due to the proof-of-concept design), and reliance on English, which may exclude non-English speakers. These findings highlight both the potential and the limitations of such systems in real-world implementation.

*Research Question 3:*

"How can the challenges of trust within a cultural context be addressed when implementing emotionally intelligent AI-driven mental health support systems in Africa?"

Trust was addressed through surface-level adaptations that mirrored culturally familiar communication styles. Naming the assistant "Karabo," integrating spiritual language, using somatic rather than clinical descriptors, and maintaining a gentle and respectful tone all contributed to a conversational dynamic that aligns with African cultural norms. These strategies were designed to build trust, comfort, and emotional safety, key precursors to meaningful engagement in digital mental health interventions. While no human participants were involved in this study, the design choices were guided by literature and experts input, suggesting their potential value in future applications.

*Research Question 4:*

"Are large language models (such as ChatGPT) ready to be used as assistive tools in detecting depression and anxiety in African contexts?"

The model showed the potential to detect and respond to symptoms of depression and anxiety as expressed through culturally grounded narratives. It was able to identify emotional distress conveyed through somatic language and social concerns and responded with supportive dialogue aligned with CBT principles. However, the model's readiness for clinical application remains limited. This study demonstrates feasibility as proof of concept, but further work, including real-world validation, multilingual capabilities, and improved cultural datasets, is needed before such tools can be responsibly deployed in clinical or public health settings. Similarly, based on the experimental results presented above, the key contributions of this study are as follows::

   i. Cultural Adaptation Framework: Developed a novel methodology for integrating Ubuntu principles (connectedness, competency, consciousness) with CBT techniques (behavioral activation, cognitive restructuring) in AI-driven mental health support.
   ii. Dataset Transformation: Created a culturally adapted dataset through a multi-step pipeline (Ubuntu injection, language simplification, faith integration, proverb incorporation) to address Western-centric biases in existing mental health datasets.
   iii. Emotionally Intelligent AI: Demonstrated the feasibility of fine-tuning LLMs to recognize and respond to Africa-centric expressions of distress, prioritizing communal well-being over individualism.
   iv. Evaluation Protocol: Established a hybrid evaluation approach (UniEval for conversational quality, manual analysis for therapeutic fidelity) to assess cultural and clinical alignment in AI interactions.
   v. Foundational Work for Digital Sovereignty: Highlighted the need for AI systems grounded in local epistemologies, offering a blueprint for future development of culturally congruent mental health tools in underrepresented regions.

## 4.8 Future directions and recommendations

As a proof-of-concept, this study successfully demonstrated the feasibility of a culturally adapted, emotionally intelligent framework for AI-driven mental health support. However, further development is required to move from conceptual validation toward real applications. The following recommendations highlight the most important areas for future work:

*Inclusion of Real User in Evaluation*

This study was conducted without the involvement of human participants, relying instead on expert-informed case studies. Future research should incorporate real users to assess how the model performs in authentic interaction scenarios. Direct user feedback is essential for evaluating the assistant's perceived empathy, trustworthiness, and cultural alignment, elements that are central to the success of AI-driven mental health tools in African contexts.

*Integration of Authentic African Proverbs*

While the assistant demonstrated some ability to integrate scriptural and culturally resonating language, the inclusion of authentic proverbs remains limited. Future efforts should prioritize collecting and curating Indigenous proverb datasets that reflect the oral traditions and cultural wisdom of various South African communities. This can be supported through Retrieval Augmented Generation (RAG) pipelines; however, the success of this approach hinges on dedicated efforts to collect, digitize, and contextualize proverbs across languages and cultures.

*Development of Culturally Grounded Dataset*

The adapted dataset used in this study was a novel and necessary intermediate solution, but not without limitations. While it successfully infused CBT conversations with Ubuntu-aligned principles, it was ultimately derived from a Western-oriented dataset. Future research should focus on building original, culturally grounded datasets that reflect how mental health symptoms, emotional experiences, and therapeutic interactions are expressed in African settings. This will improve the fidelity and authenticity of AI interventions, ensuring they truly reflect African lived realities.

It is important to reemphasis the fact that the primary limitation of this study concerns the origin and authenticity of the training data. Our culturally adapted dataset was synthetically generated by another LLM, with high probability of introducing the risk of a bias propagation chain and potentially lacking the nuanced linguistic, cultural, and emotional characteristics present in dialogues created by and for African users. Future research should prioritize the collection, curation, and expert-driven annotation of authentic, real-world mental health conversations from the target populations. Such a foundation is critical for advancing beyond synthetic data adaptations and for developing AI systems that embody genuine cultural relevance and contextual emotional intelligence.

*AI Safety, Ethics and Community Safety*

It is equally important to note that while the current study proof-of-concept demonstrate the potential for culturally sensitive AI-driven mental health support, it is imperative to directly address the risks and harm associated with the real-world deployment of such applications. Study have also revealed that there are instance whereby Chatbot's, even those design for support have provided dangerous advice, exacerbated symptoms of depression and anxiety, provoked suicidality, or reinforced harmful behaviors such as those associated with eating disorders (Yoo et al., 2025; Abd-Alrazaq et al., 2020). Similarly, the integration of spiritual and religious elements, as proposed in our framework, while culturally resonant, introduces an additional layer of complexity and potential risk, necessitating rigorous oversight. Therefore, to ensure the safe and ethical development of future iterations of this framework, the following measures are critically recommended: implementing concrete safety mechanisms, conducting rigorous multi-stakeholder evaluation of cultural and religious content, and performing longitudinal real-world trials.

## 5. Conclusion

This study set out to explore how emotionally intelligent LLMs could be adapted to provide culturally sensitive mental health support within African contexts, specifically through the integration of Ubuntu values and Cognitive Behavioural Therapy (CBT) techniques. A novel framework was proposed and tested through a proof-of-concept implementation in response to the growing mental health challenges across South Africa and the limitations of Western-centric AI tools.

The methodology combined deep and surface-level cultural adaptations, from Ubuntu-based theoretical structuring to practical language, communication style, and therapeutic alignment modifications. A dataset was adapted to reflect African realities, and case studies were developed in consultation with a senior counseling psychologist to simulate authentic user experiences. The resulting system was evaluated across multiple dimensions: conversation quality (via UniEval), therapeutic integrity (via CBT analysis), and cultural-linguistic alignment (via surface-level communication strategies).

Findings showed that the assistant was able to respond with empathy, contextual awareness, and therapeutic relevance, particularly in its use of Behavioural Activation and Cognitive Restructuring aligned with communal values. It also demonstrated an ability to engage users through familiar spiritual and social references, affirming the importance of cultural grounding in AI systems designed for mental health.

While the results are promising, they are not without limitations. The absence of real user interaction, the use of an adapted (rather than authentically generated) dataset, and the lack of multilingual support point to necessary areas for future work. Nevertheless, this study demonstrates the feasibility and importance of developing AI systems that reflect African users' emotional, cultural, and psychological realities.

Ultimately, this research contributes to the growing discourse on digital sovereignty and cultural inclusivity in AI. Centering African perspectives in the design of emotionally intelligent mental health tools addresses a critical service gap. It sets a precedent for building AI that genuinely resonates with the communities it serves.

**Declarations**

**Funding:** Not Applicable.

**Clinical Trial Number**: Not Applicable

**Ethical Statements:** The authors have read and followed the ethical requirements for publication, confirming that the current work does not directly involve human subjects, animal experiments, or any data collected from social media platforms. Similarly, the research received ethical clearance approval from the North-West University Health Research Ethics Committee (NWU-HREC), with ethics Numbers: **NWU-00107-24-A1**.

**Ethical Guidelines**: This study was performed in line with the principles of the Declaration of Helsinki. Approval was granted by the Ethics Committee of North-West University Health Research Ethics Committee (NWU-HREC), with ethics Numbers: **NWU-00107-24-A1**.

**Informed Consent**: While experts input was sought on the prototype's design, this study did not involve human participant testing. Consequently, the collection of informed consent was not required.

**Consent to Publish Declaration:** The requirement for written informed consent for publication was waived by the North-West University Health Research Ethics Committee due to the retrospective and fully anonymized nature of the study. The manuscript does not contain any individually identifiable data from any participant.

**Consent to Participate:** Not Applicable.

**Declaration of Conflicting Interests:** The authors declare that there is no conflict of interest with regard to the publication of this paper.

**Data Availability Statements:** The datasets used and/or analysed during the current study are available from the corresponding author on reasonable request.

**Declaration of Generative AI and AI-Assisted Technologies in the Writing Process:**

While preparing this work, the author utilized AI-assisted tools such as ChatGPT and Grammarly to enhance English language accuracy, including spelling, grammar, and punctuation. To ensure accuracy and originality, the authors thoroughly reviewed, revised, and edited the content generated or corrected by these tools. The author accepts full responsibility for the originality of the final content of this publication.

**Appendix A: Github Packages**

Backend package: https://github.com/Sontaga25/Chatbot_backend.git

FrontEnd package: https://github.com/Sontaga25/Chatbot_FrontEnd.git

**Appendix B:** Google Form Link And Results

**Form Link:** Adaptation Rating - Google Forms

**Results:**
https://docs.google.com/spreadsheets/d/1TFNqgXO_MvYiDCTs15aZK3Bu2gYp7lFygZZ9zn9BFLQ/edit?usp=sharing